\documentclass[runningheads]{llncs}

\usepackage{eccv}

\usepackage{eccvabbrv}
\usepackage{graphicx}
\usepackage{array}
\usepackage{booktabs}
\usepackage{multirow}
\usepackage{adjustbox}
\usepackage{float}
\usepackage{tikz}
\usetikzlibrary{arrows.meta,positioning}

\usepackage[accsupp]{axessibility}  

\usepackage{hyperref}

\usepackage{orcidlink}

\newcommand\ourmethod{PromptEvolver}

\newcommand{\promptexample}[7]{%
\begin{figure}[H]%
\centering%
\begin{minipage}[t]{0.30\textwidth}\centering
\includegraphics[width=\textwidth]{#1}\\[2pt]
{\small (a) Reference image}
\end{minipage}\hfill
\begin{minipage}[t]{0.30\textwidth}\centering
\includegraphics[width=\textwidth]{#2}\\[2pt]
{\small (b) Initial generation}\\[2pt]
{\scriptsize\raggedright\textbf{Initial prompt:} #4\par}
\end{minipage}\hfill
\begin{minipage}[t]{0.30\textwidth}\centering
\includegraphics[width=\textwidth]{#3}\\[2pt]
{\small (c) Evolved reconstruction}\\[2pt]
{\scriptsize\raggedright\textbf{Evolved prompt:} #5\par}
\end{minipage}
\caption{#6}\label{#7}%
\end{figure}%
}

\newcommand{\failurecase}[5]{%
\begin{figure}[H]%
\centering%
\setlength{\tabcolsep}{4pt}%
\begin{tabular}{cc}%
\includegraphics[width=0.35\textwidth]{#1} &
\includegraphics[width=0.35\textwidth]{#2} \\
{\small (a) Original} & {\small (b) Reconstructed}%
\end{tabular}\\[6pt]%
{\small%
\begin{tabular}{@{} p{0.95\textwidth} @{}}%
\textbf{Evolved Prompt:} #3%
\end{tabular}}%
\caption{#4}\label{#5}%
\end{figure}%
}

\begin{document}

\title{PromptEvolver: Prompt Inversion through Evolutionary Optimization in Natural-Language Space}

\titlerunning{PromptEvolver}

\author{Asaf Buchnick\inst{1}\orcidlink{0009-0007-0682-2602} \and
Aviv Shamsian\inst{1}\orcidlink{0009-0008-7418-5024} \and
Aviv Navon\inst{1}\orcidlink{0000-0002-5788-5597} \and
Ethan Fetaya\inst{1}\orcidlink{0000-0003-3125-1665}}

\authorrunning{A. Buchnick et al.}

\institute{Bar-Ilan University, Ramat Gan,
Israel
\email{\{buchnia5,aviv.shamsian,aviv.navon,ethan.fetaya\}@biu.ac.il}}

\maketitle

\begin{abstract}
Text-to-image generation has progressed rapidly, but faithfully generating complex scenes requires extensive trial-and-error to find the exact prompt. In the prompt inversion task, the goal is to recover a textual prompt that can faithfully reconstruct a given target image. Currently, existing methods frequently yield suboptimal reconstructions and produce unnatural, hard-to-interpret prompts that hinder transparency and controllability. In this work, we present \ourmethod{}, a prompt inversion approach that generates natural-language prompts while achieving high-fidelity reconstructions of the target image. Our method uses a genetic algorithm to optimize the prompt, leveraging a strong vision-language model to guide the evolution process. Importantly, it works on black-box generation models by requiring only image outputs.  Finally, we evaluate \ourmethod{} across multiple prompt inversion benchmarks and show that it consistently outperforms competing methods.
\end{abstract}

\section{Introduction}
\label{sec:intro}


Text-to-image (T2I) diffusion models~\cite{rombach2022high,labs2025flux1kontextflowmatching,xie2024sana} have transformed visual content creation, enabling users to generate photorealistic images from natural-language prompts. Yet the quality of the generated image depends critically on the prompt, when even small changes in wording can produce dramatically different outputs. In practice, users engage in extensive trial-and-error, manually crafting, adjusting, and iterating on prompts to achieve a desired visual result. This process is time-consuming, and best practices can often change as T2I models evolve.\\

Prompt inversion addresses the reverse process: given a target image, recover a textual prompt that, when fed to a T2I model, faithfully reconstructs that image. One application is for image editing \cite{mokady2023null}, where a modified prompt is used to generate targeted changes in the image. This can also help T2I practitioners by generating a prompt to a reference image we wish to generate a variation of. Beyond creative workflows, prompt inversion can enable understanding and auditing of generative models by revealing what textual descriptions best explain a given output. In
 \cite{croitoru2024reverse}, the authors also showed that prompt-inversion can be used to help faithful generation.\\

Formally, given a target image $I$ and a T2I model $\mathcal{M}$, the goal is to find a prompt ${p}$ such that if $\hat{I}$ is sampled from $\mathcal{M}({p})$ then $\hat{I} \approx I$ (in some semantically meaningful metric).
This is fundamentally challenging for multiple reasons. First, we might not have direct access to the model $\mathcal{M}$, and are just able to call an external service to generate images. Moreover, even when the model is available, computing gradients through the whole generation process is not trivial. Additionally, the optimization is over the complex and discrete space of natural language. Finally, the model is highly sensitive to the prompt, and small changes in wording can lead to drastically different generated images, while semantically equivalent prompts may yield very different visual outputs~\cite{mo2024dynamic, dehouche2023s}.

Recently, a substantial body of works has tackled prompt inversion through optimization in continuous embedding or latent spaces~\cite{gal2022image,ruiz2023dreambooth,mokady2023null,wei2023elite}.
While these methods can achieve high-fidelity reconstruction, they suffer from several fundamental limitations.
First, they assume white-box access to the model (e.g., gradients, activations, or parameter updates), which makes them unsuitable for closed-source or API-based models.
Second, the recovered representations are not human-readable: they exist as embedding vectors or modified model weights rather than natural-language text, preventing users from understanding, editing, or transferring the result.
Third, these methods are inherently tied to the specific model architecture on which they were optimized, and do not generalize across different T2I models.
Alternative approaches that operate in discrete token space~\cite{wen2023hard,mahajan2024prompting,qiu2025steps,li2025editor} have also been proposed recently, yet they still rely on gradient-based optimization through the text encoder, with the resulting prompts often lacking grammatical coherence and naturalness.
Gradient-free methods~\cite{pharmapsychotic2022clipinterrogator,kim2025visually,he2024automated} avoid the need for model internals, but typically employ single-trajectory search strategies, such as beam search or in-context refinement, that are prone to local optima, limiting their reconstruction quality. Moreover, we found that a simple baseline that queries a powerful vision-language model (VLM) with a straightforward instruction prompt can be on-par or outperform these more complex prompt-optimization approaches.\\


In this work, we present \ourmethod{}, a prompt inversion method that overcomes these limitations through \textit{evolutionary optimization} in natural-language space.
Our key insight is that a VLM can serve as both the generator and the refiner of candidate prompts. It produces diverse initial descriptions of the target image, and then acts as the crossover and mutation operator in a genetic algorithm, combining and modifying prompts while directly observing the target image. Unlike embedding-level methods, \ourmethod{} operates entirely in the space of human-readable text, producing natural prompts that users can inspect, understand, and edit. Differently from gradient-based discrete methods~\cite{qiu2025steps}, it requires no access to model internals-only the ability to query the T2I model and observe its output, making it applicable to both open source and black-box models. Furthermore, in contrast to single-trajectory search, our genetic algorithm performs population-based optimization that preserves diversity across candidates. This promotes a broader exploration of the prompt space and reduces the susceptibility to finding a bad local optima.\\

\begin{figure*}[h]
\centering
\setlength{\tabcolsep}{1.5pt}
\renewcommand{\arraystretch}{0.8}
\newcolumntype{C}{>{\centering\arraybackslash}m{0.15\textwidth}}
\begin{tabular}{m{1.2em}CCCCCCCC}
\rotatebox{90}{\small\textbf{Original}} &
\includegraphics[width=0.15\textwidth]{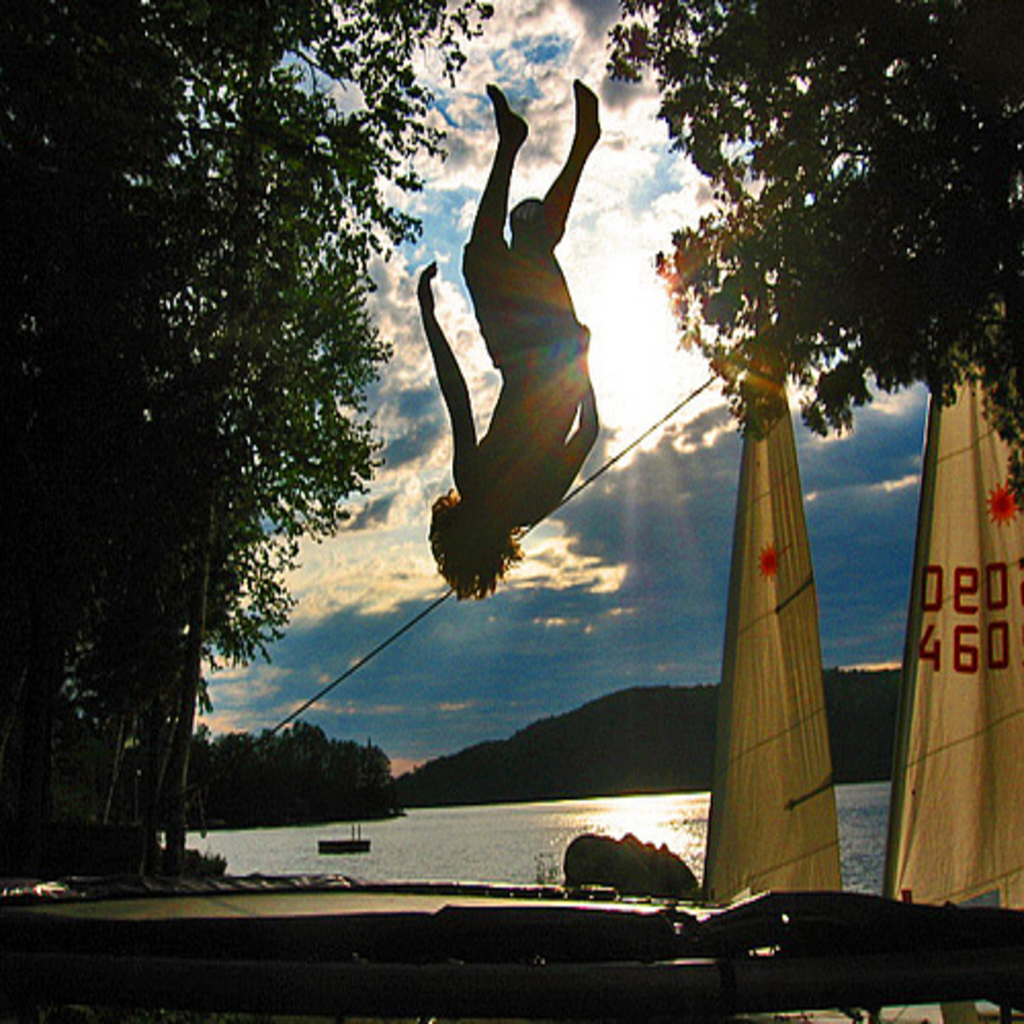} &
\includegraphics[width=0.15\textwidth]{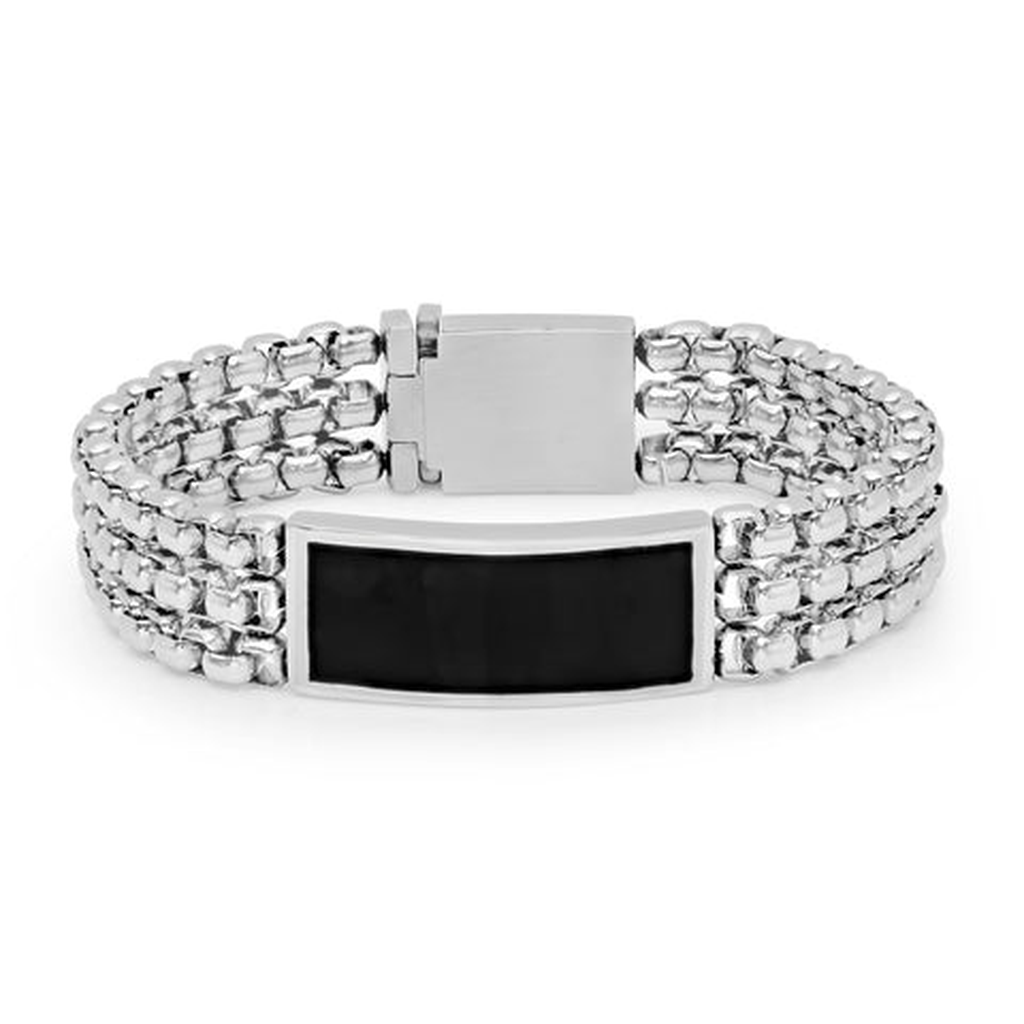} &
\includegraphics[width=0.15\textwidth]{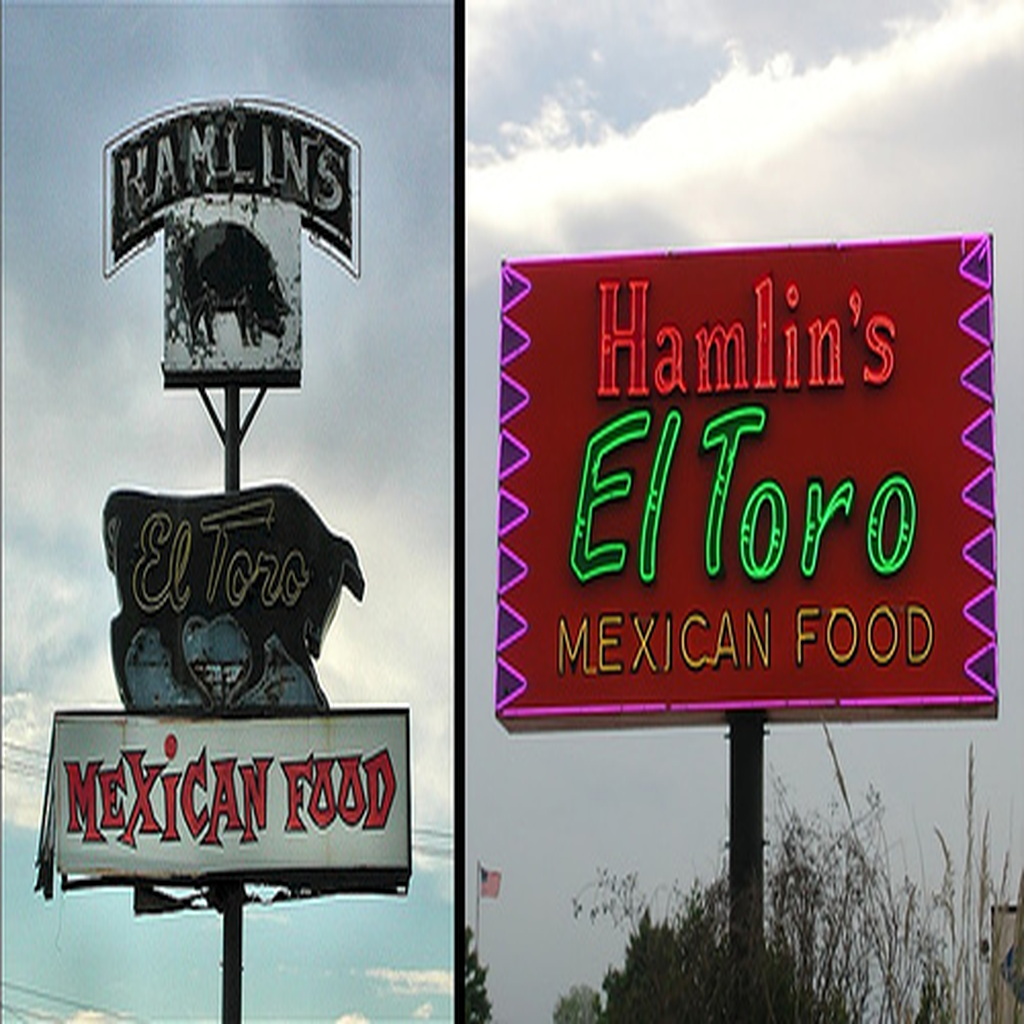} &
\includegraphics[width=0.15\textwidth]{figs/main/1/original/image_408_original.png} &
\includegraphics[width=0.15\textwidth]{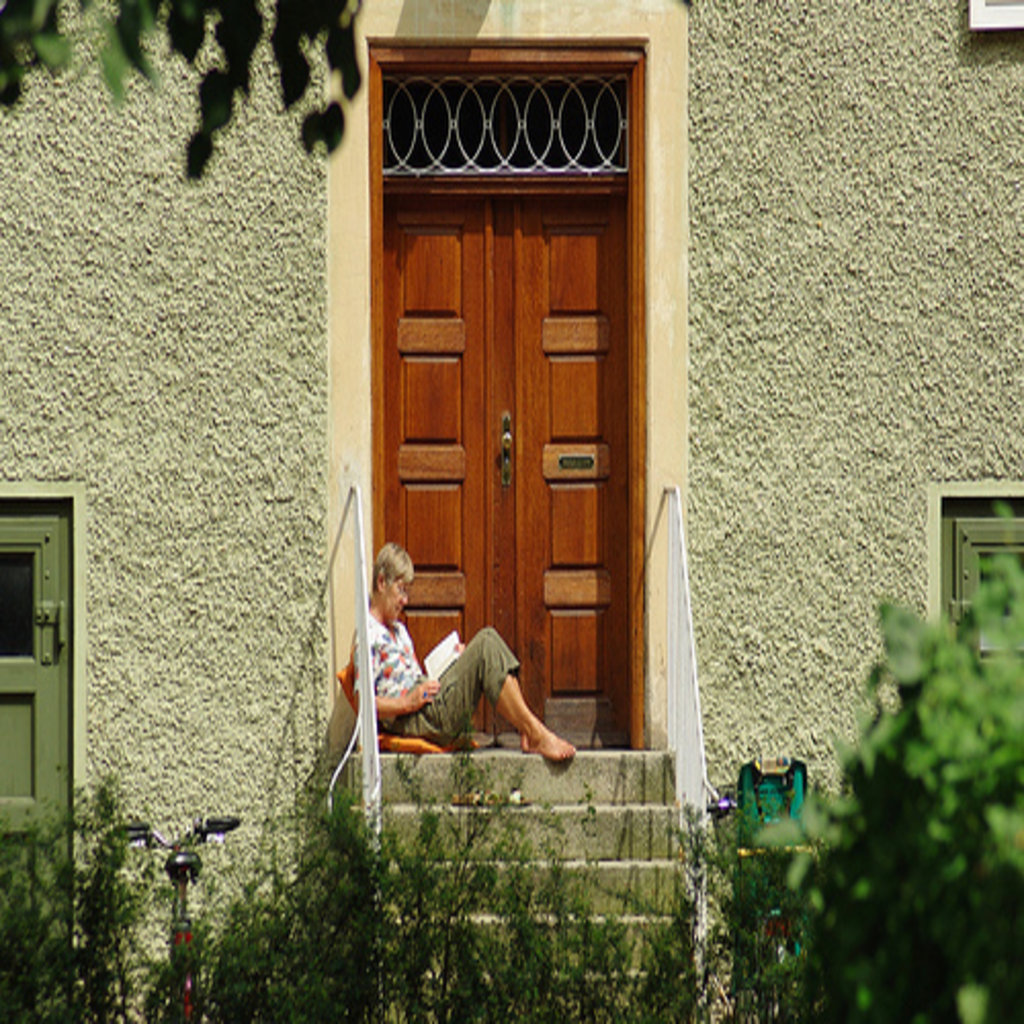} &
\includegraphics[width=0.15\textwidth]{figs/main/1/original/image_428_original.png} \\[1pt]
\rotatebox{90}{\small\textbf{Ours}} &
\includegraphics[width=0.15\textwidth]{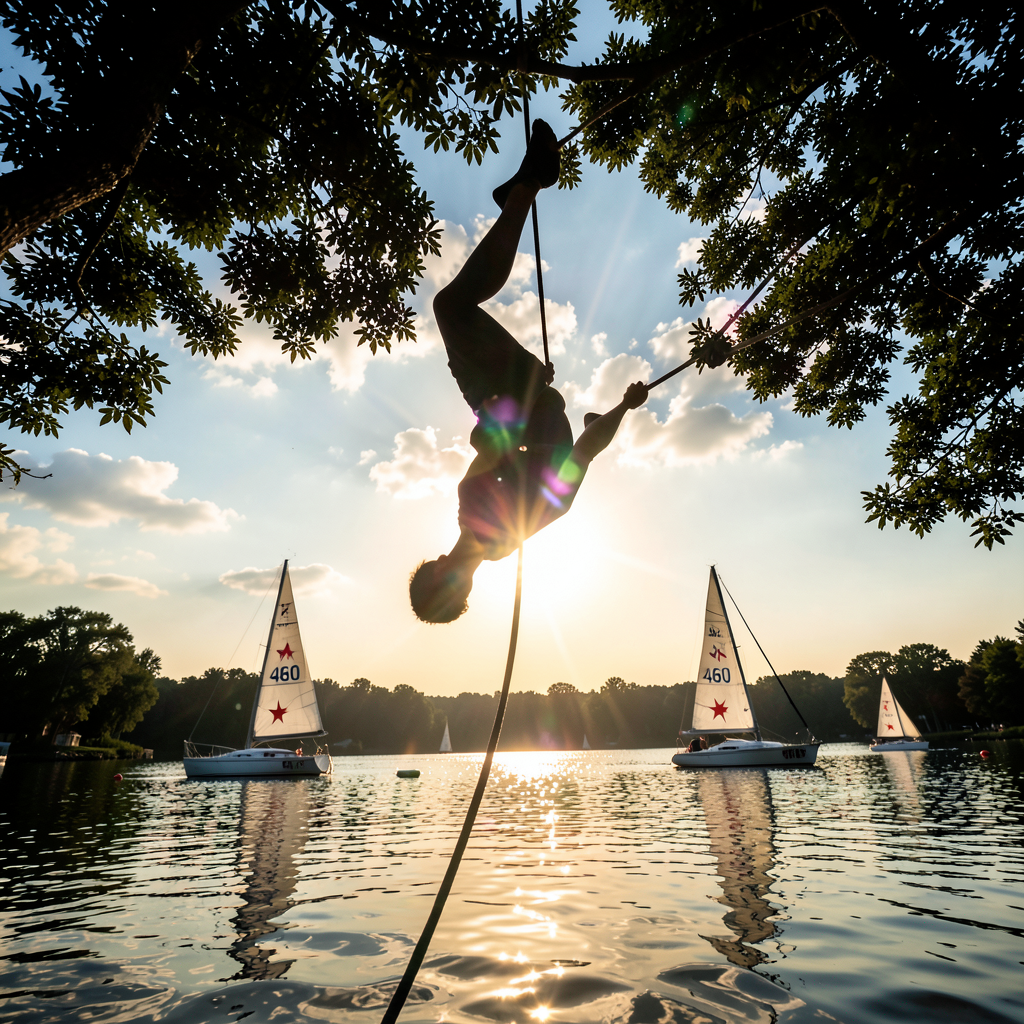} &
\includegraphics[width=0.15\textwidth]{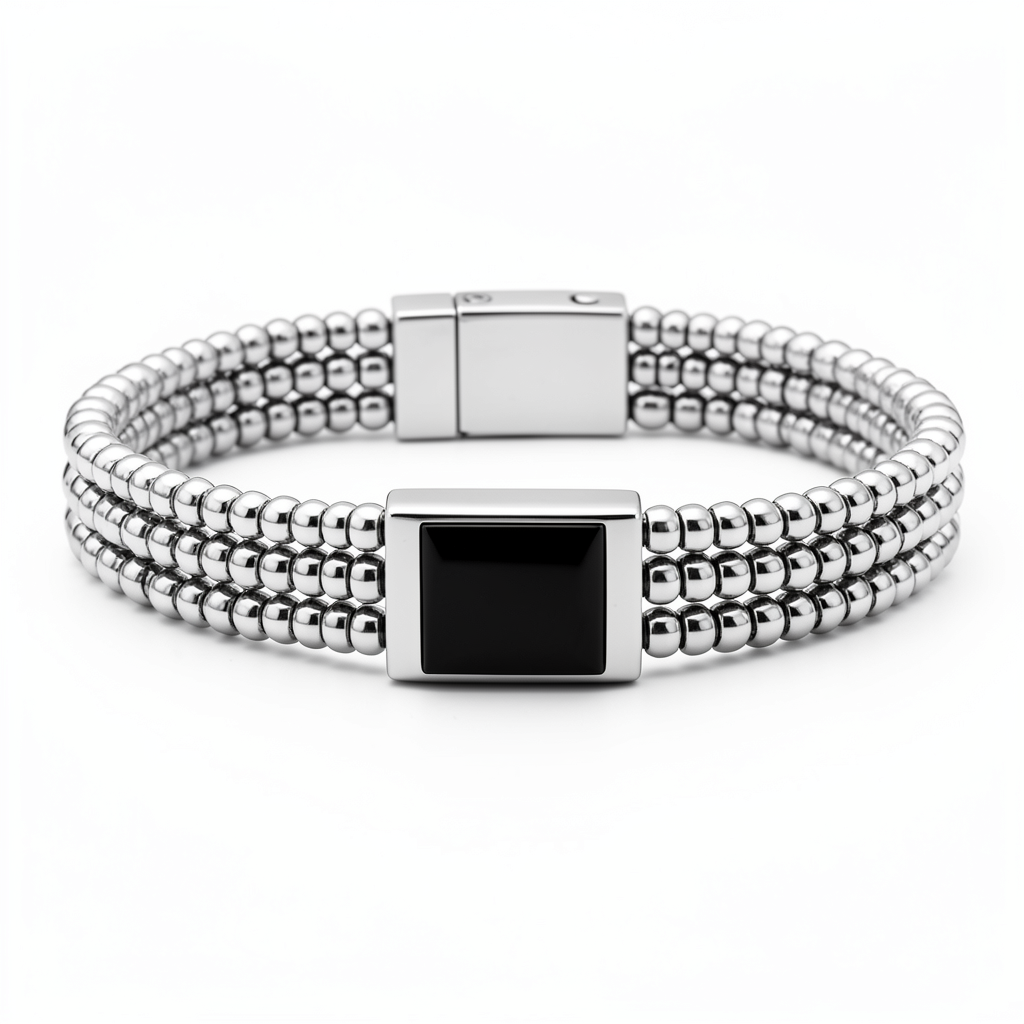} &
\includegraphics[width=0.15\textwidth]{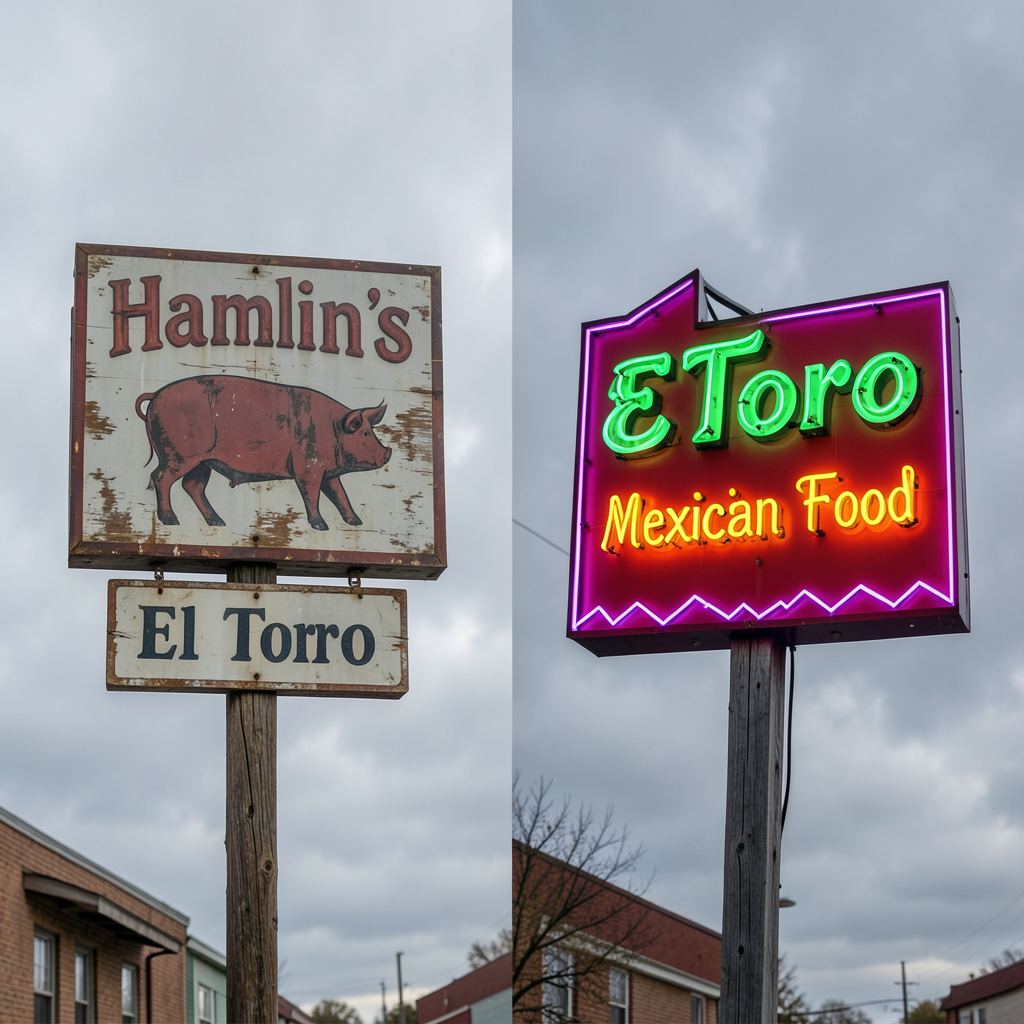} &
\includegraphics[width=0.15\textwidth]{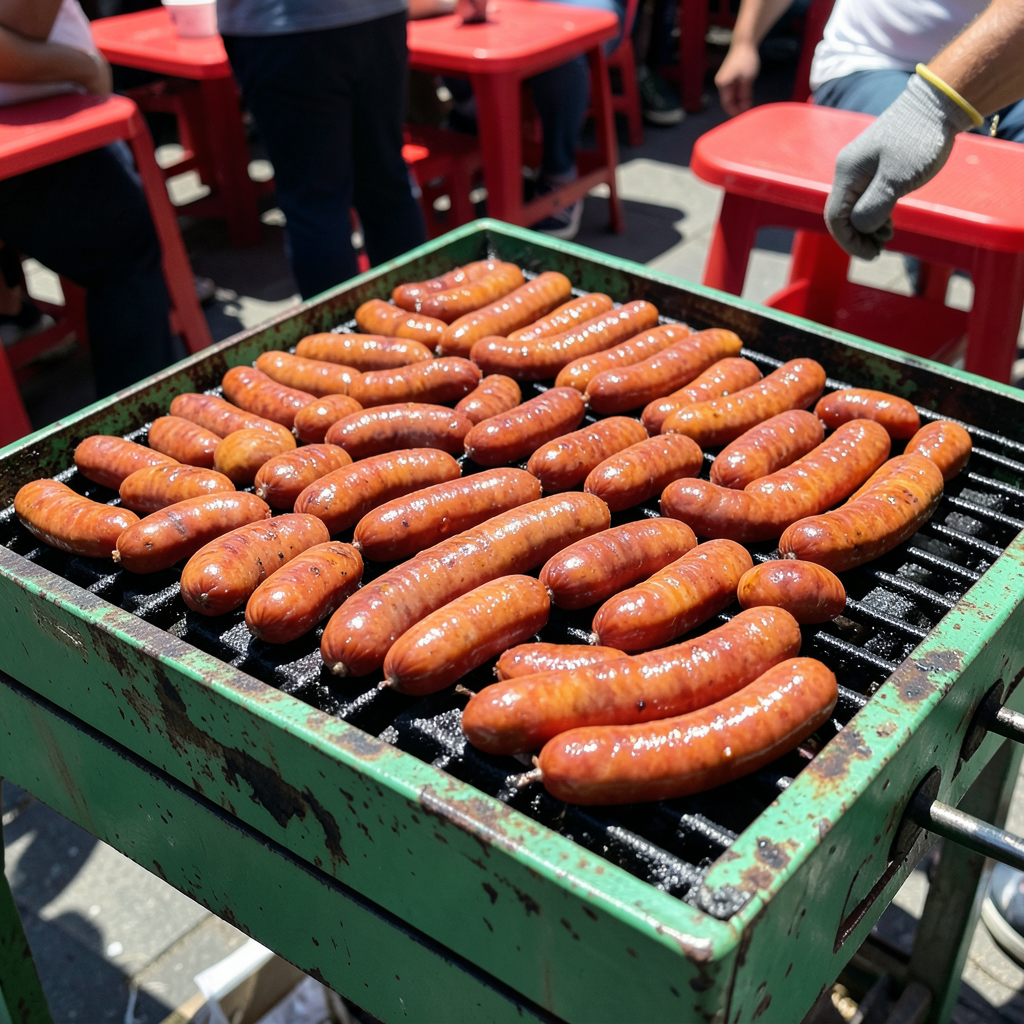} &
\includegraphics[width=0.15\textwidth]{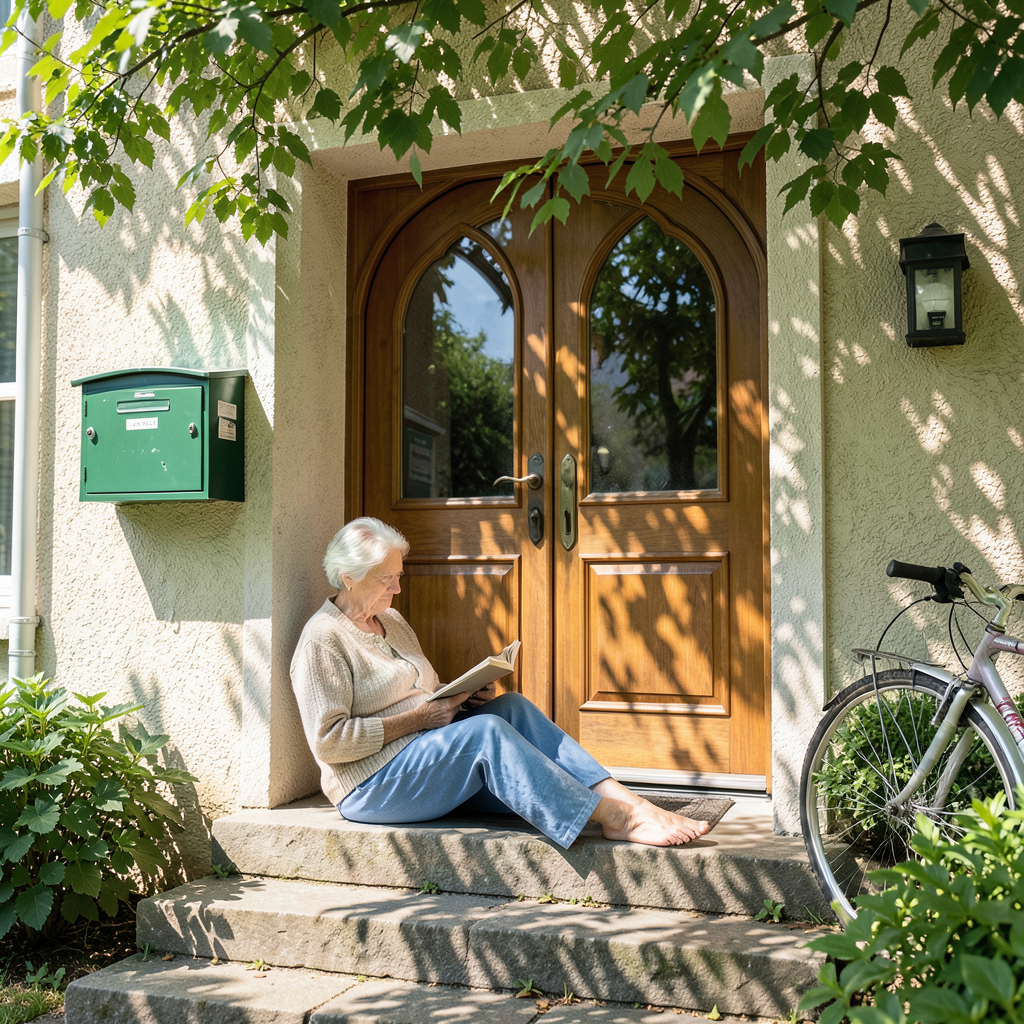} &
\includegraphics[width=0.15\textwidth]{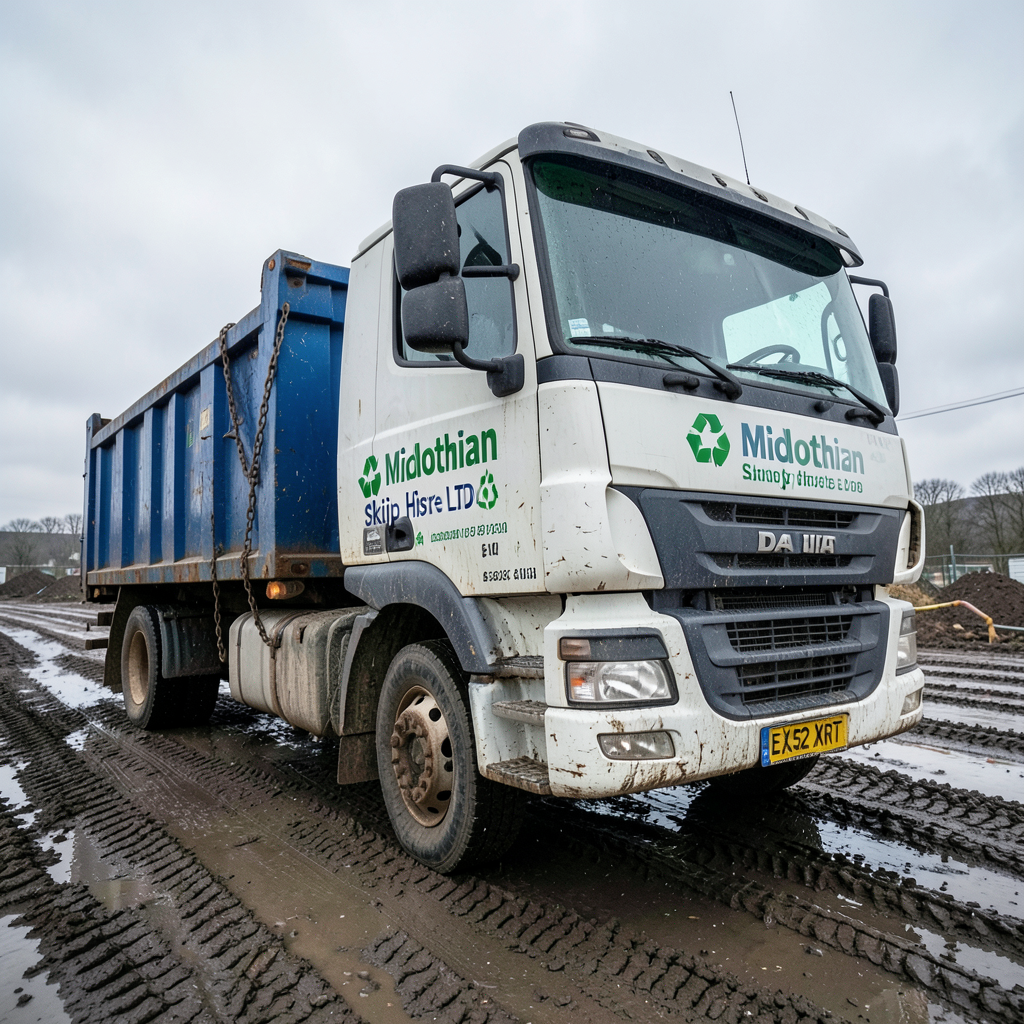} \\[1pt]
\rotatebox{90}{\small\textbf{VLM-Base.}} &
\includegraphics[width=0.15\textwidth]{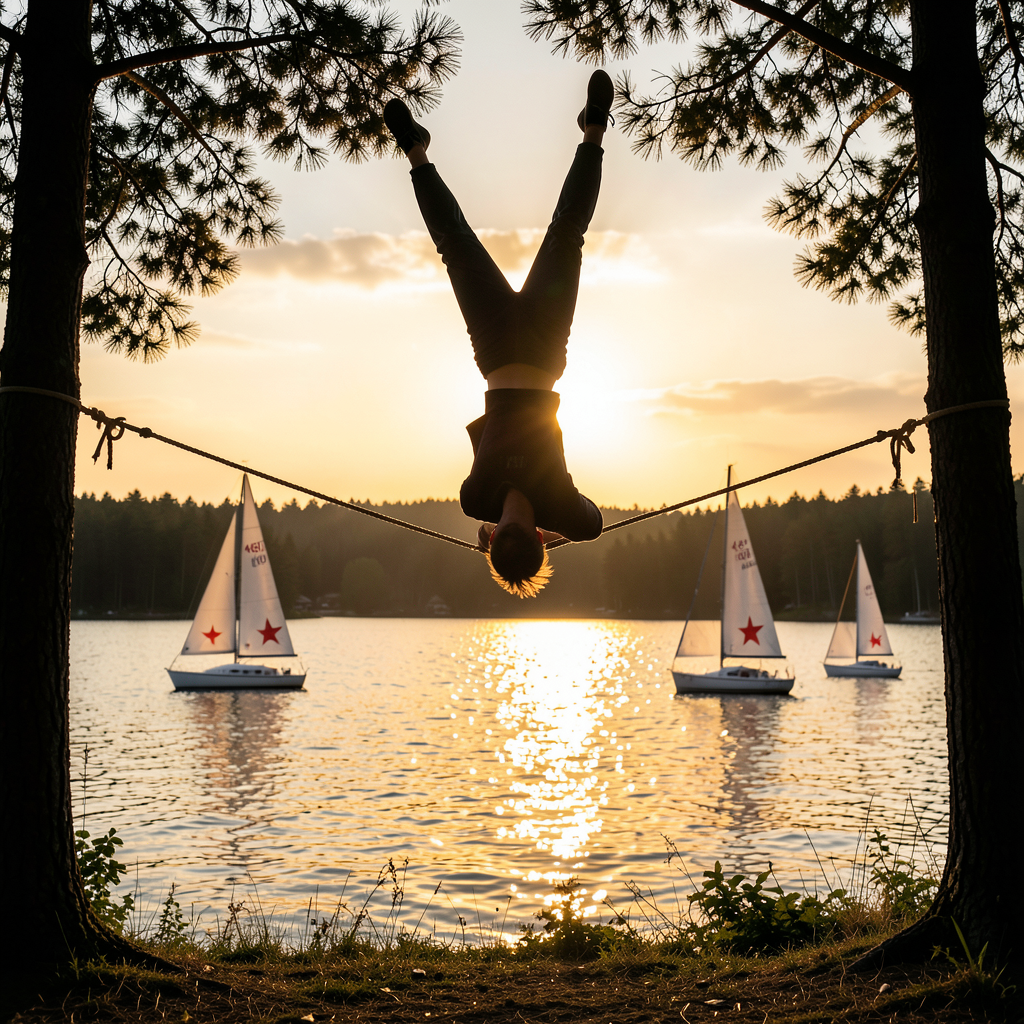} &
\includegraphics[width=0.15\textwidth]{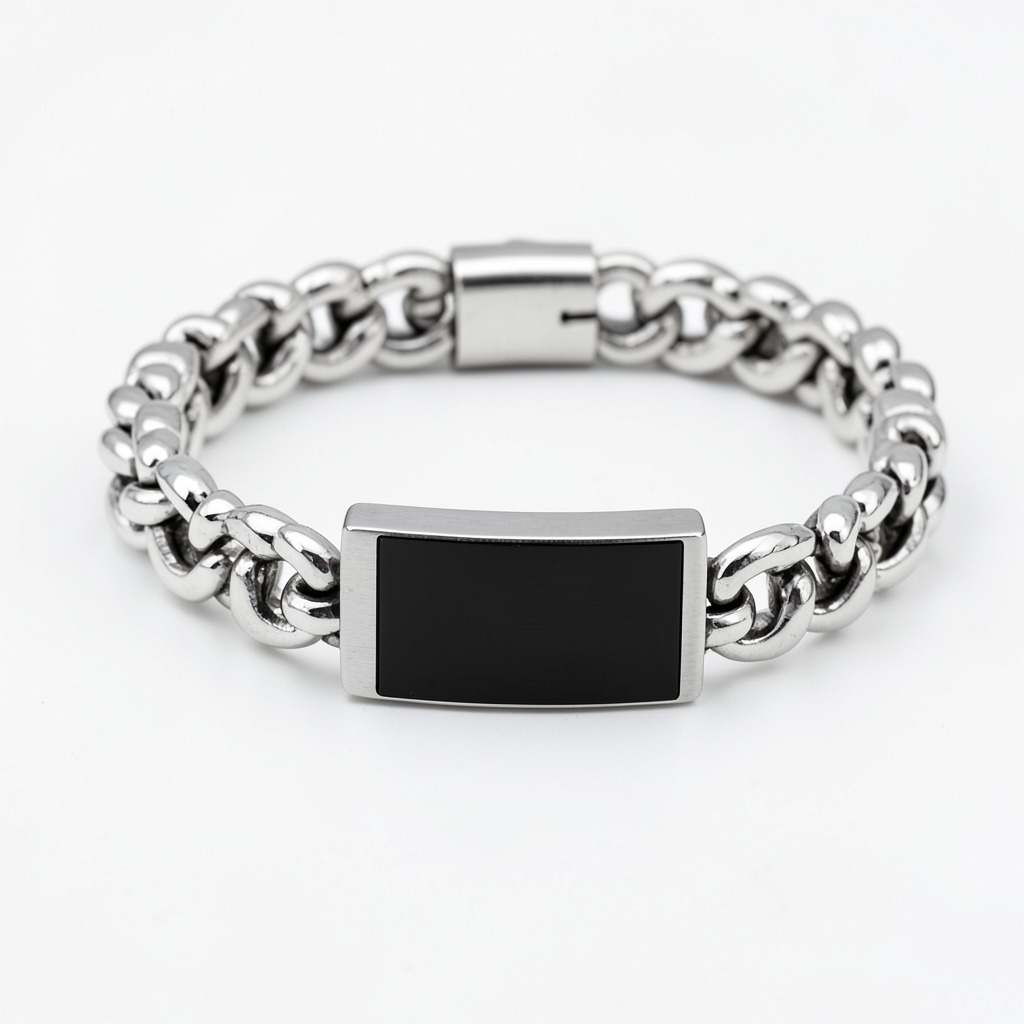} &
\includegraphics[width=0.15\textwidth]{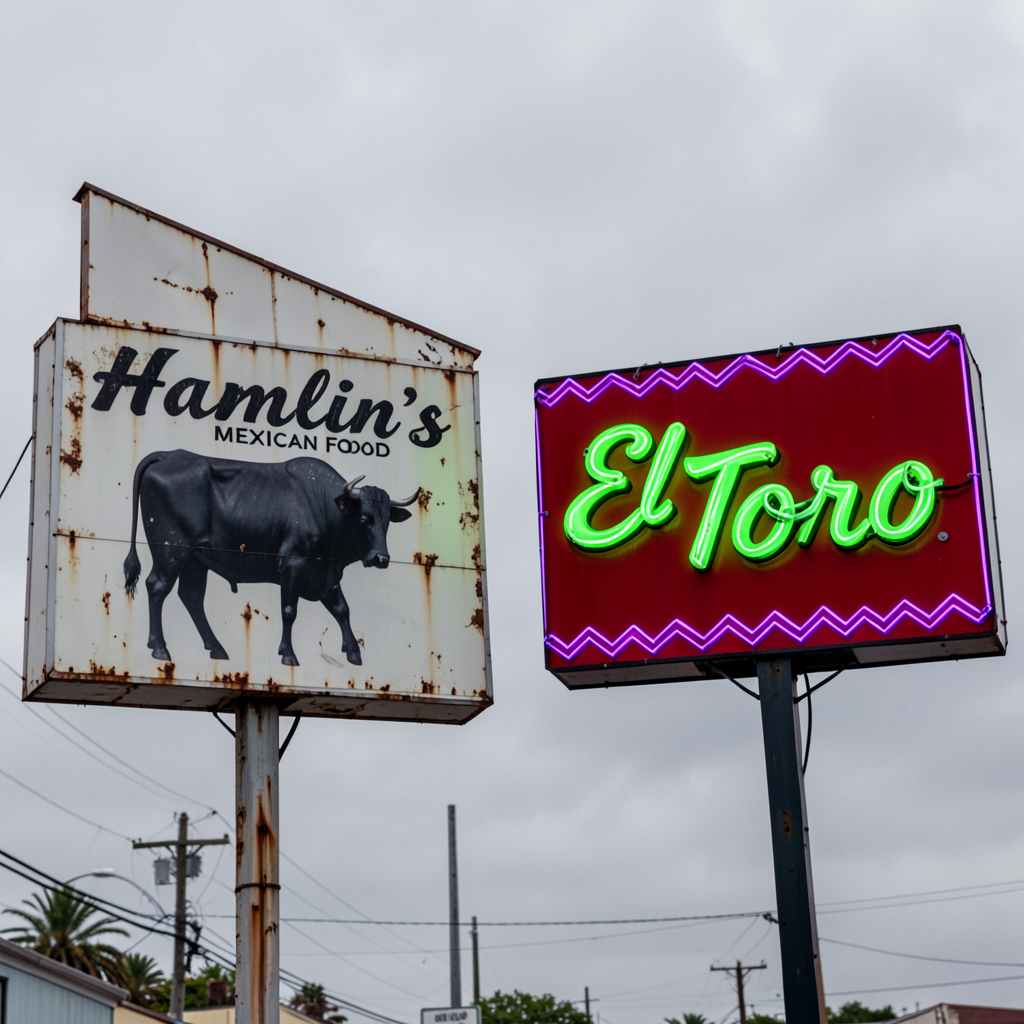} &
\includegraphics[width=0.15\textwidth]{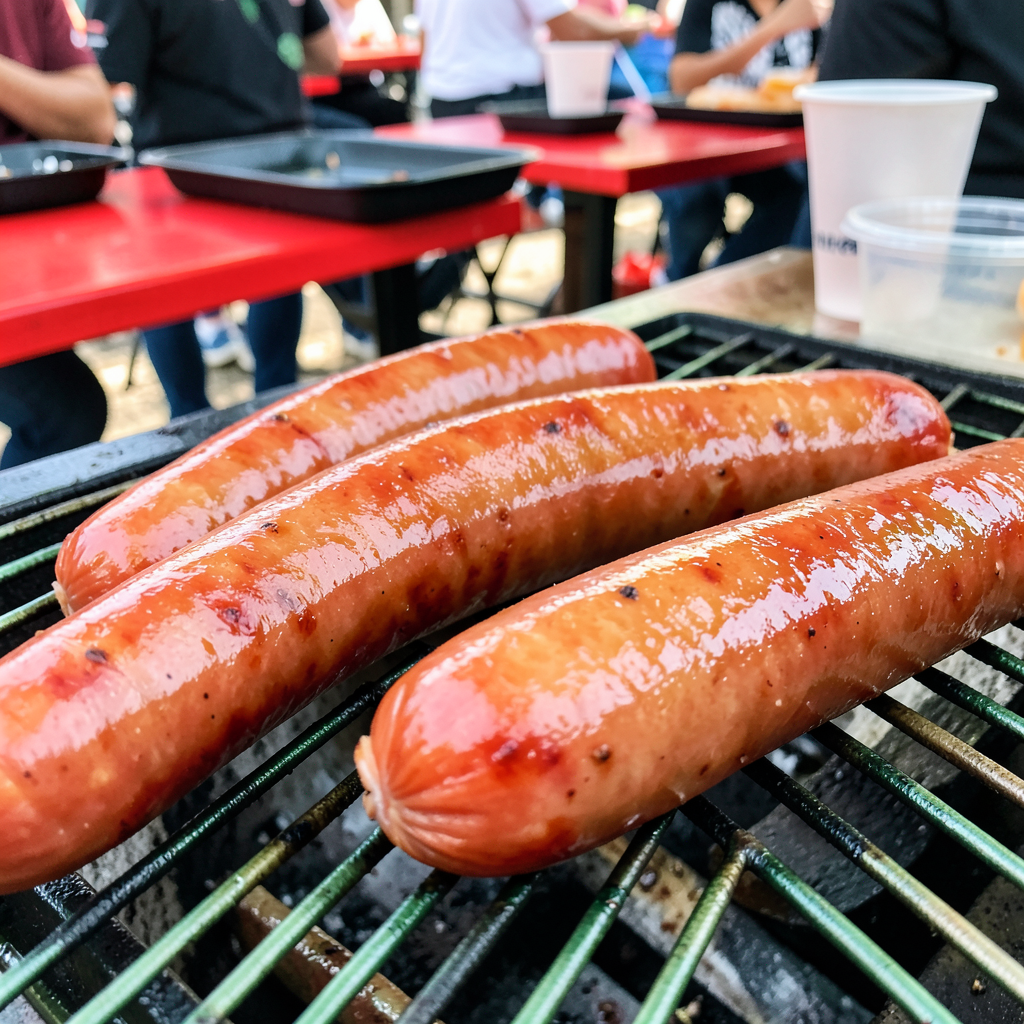} &
\includegraphics[width=0.15\textwidth]{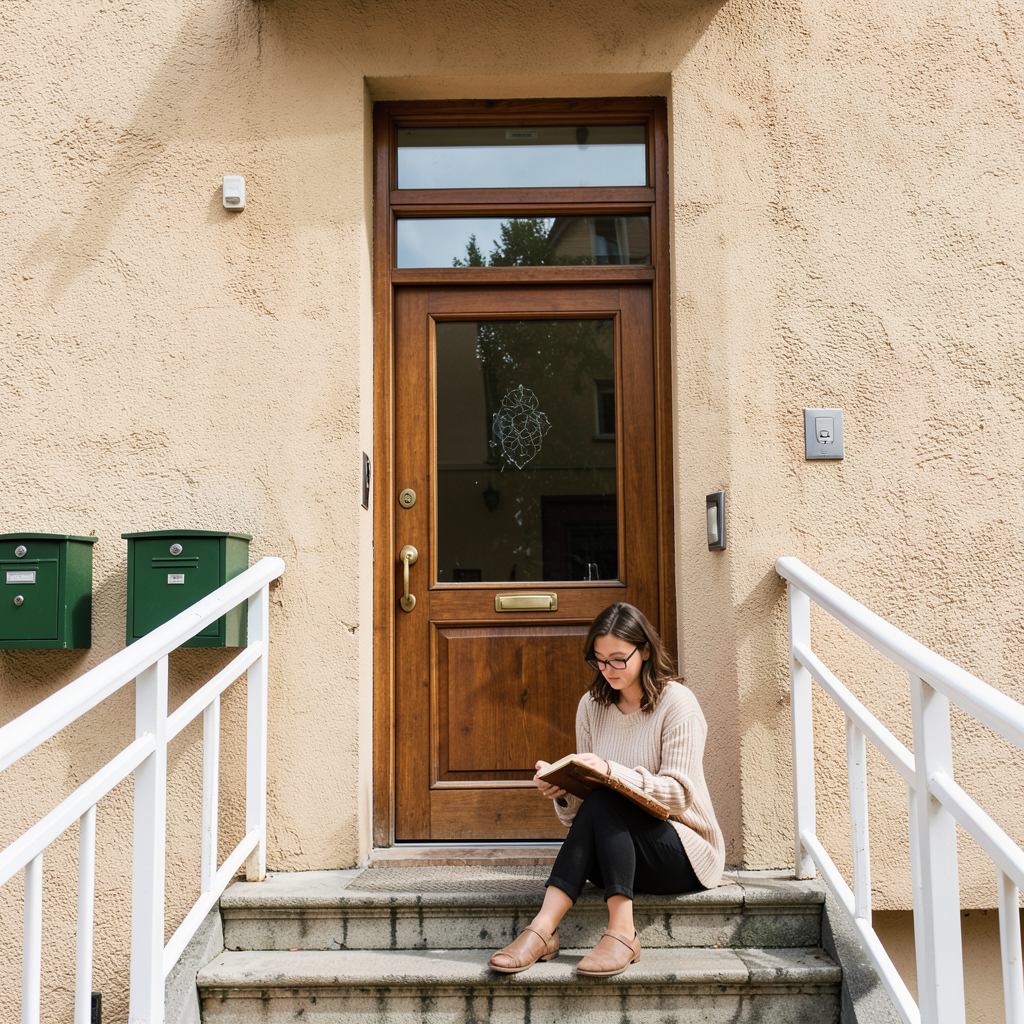} &
\includegraphics[width=0.15\textwidth]{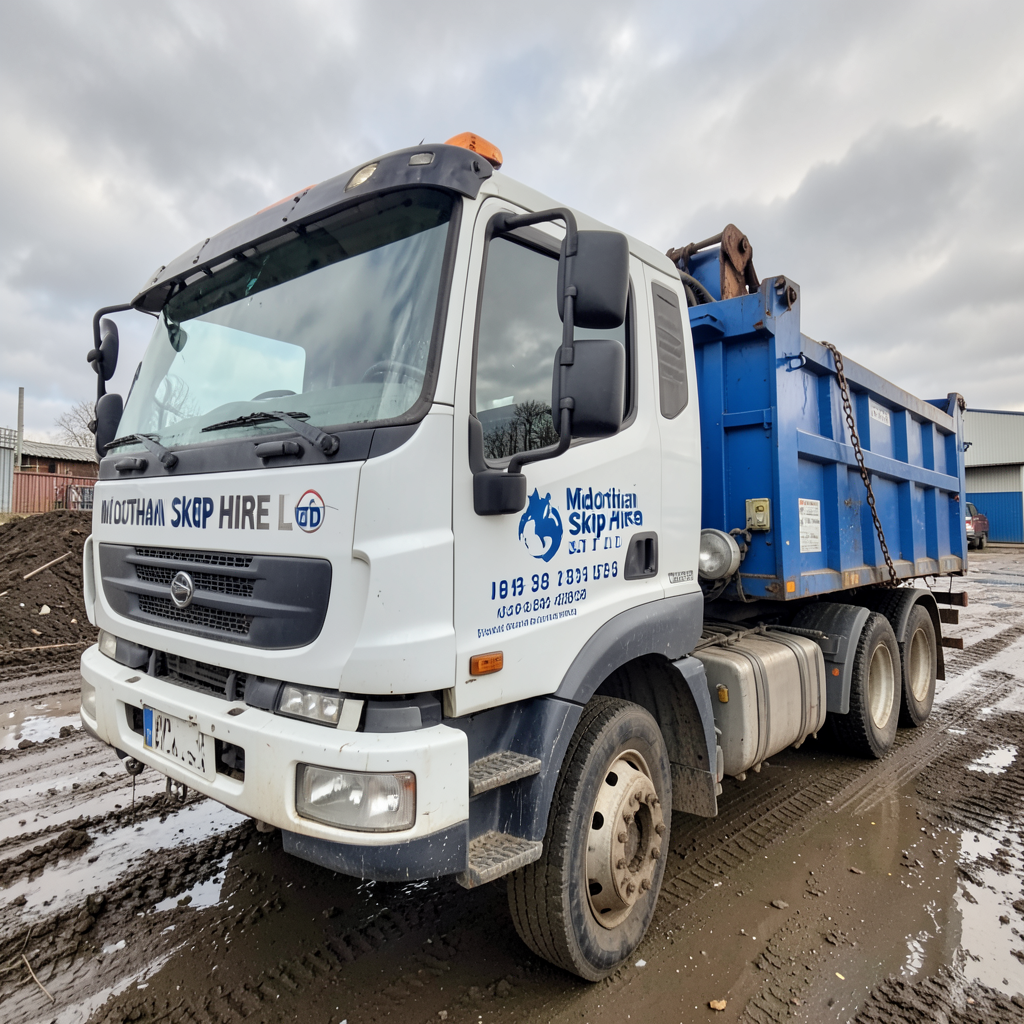} \\
\end{tabular}
\caption{\textit{Reconstruction comparison}: \ourmethod{} versus a VLM baseline. We demonstrate the robustness of \ourmethod{} to challenging T2I concepts such as colors, counting, object positioning, and camera orientation.}
\label{fig:intro_examples}
\end{figure*}

We provide an example of images generated by our recovered prompt vs the VLM-baseline in Fig. \ref{fig:intro_examples}. As one can see from these examples, both models nicely capture the essence of the original image, but \ourmethod{} better captures the fine details, like text in the images, the women age, the number of layers in the bracelet, etc.\\

We summarize our contributions as follows: (i) We propose \ourmethod{}, the first text-level evolutionary framework for prompt inversion that uses image-aware VLM operators-crossover and mutation conditioned on the target image to evolve natural-language prompts through a genetic algorithm; (ii) We demonstrate that \ourmethod{} achieves SoTA results in prompt inversion compared to recent baselines. Our method consistently outperforms the baselines across multiple benchmarks, achieving up to 7.8\%  improvement in image reconstruction score.


\section{Related Work}
\label{sec:related}
\subsection{Prompt Inversion for Text-to-Image Models}
Prompt inversion seeks to recover a textual representation that, when fed to a T2I model, faithfully reconstructs a given target image.
A large body of work addresses this problem through gradient-based optimization in continuous embedding or latent spaces.
Textual Inversion~\cite{gal2022image} learns a pseudo-word embedding for a particular object by minimizing the denoising loss, with extensions such as P+~\cite{voynov2023p+} and NeTI~\cite{alaluf2023neural} enriching the representation via per-layer or timestep-dependent conditioning.
Although Textual Inversion was introduced for concept personalization, its embedding-optimization objective can be directly applied to single-image reconstruction by optimizing a pseudo-token embedding to reproduce a target image.
Beyond embedding-only inversion, other continuous approaches adapt the generator itself.
DreamBooth~\cite{ruiz2023dreambooth} and Custom Diffusion~\cite{kumari2023multi} fine-tune model weights to bind subjects to unique identifiers, while feed-forward encoders such as ELITE~\cite{wei2023elite}, E4T~\cite{arar2023domain}, and BLIP-Diffusion~\cite{li2023blip} amortize the cost by mapping images to embeddings in a single forward pass.
A parallel line of work inverts images into the noise latent space: DDIM inversion~\cite{song2020denoising} reverses the deterministic sampling ODE, and subsequent methods improve its accuracy~\cite{wallace2023edict,mokady2023null,pan2023effective,garibi2024renoise}, enabling high-fidelity reconstruction and downstream editing~\cite{hertz2022prompt,tumanyan2023plug,brack2024ledits++}.\\

Critically, both embedding-level and latent-space methods are designed primarily for object-centric scenarios with a single dominant subject~\cite{gal2022image,ruiz2023dreambooth,kumari2023multi}, and struggle with cluttered scenes containing multiple subjects, fine-grained spatial relationships, and rich stylistic attributes.
More recent methods target the discrete token space directly, aiming to recover natural-language prompts via gradient-based optimization.
PEZ~\cite{wen2023hard} optimizes one-hot token distributions via projected gradient descent to maximize CLIP similarity between the prompt and target image embeddings.
PH2P~\cite{mahajan2024prompting} improves upon PEZ with a delayed projection scheme and timestep-aware optimization, yielding more interpretable prompts.
STEPS~\cite{qiu2025steps} reformulates the search as sequential probability tensor estimation for more efficient exploration of the combinatorial token space.
EDITOR~\cite{li2025editor} initializes embeddings with a captioning model, then refines them in latent space and decodes them to text, achieving strong image similarity and prompt interpretability.
Reverse Stable Diffusion~\cite{croitoru2024reverse} takes a supervised approach, training a model on large-scale prompt-image pairs to directly predict the generating prompt.
All the above methods require gradient access to the T2I model, model-specific training, or optimization in non-interpretable embedding spaces.
In contrast, \ourmethod{} operates purely in natural language, needs no gradients or model changes, and produces human-readable prompts.

\subsection{Gradient-free prompt inversion.}
Most relevant to our work are gradient-free methods that produce readable prompts without requiring access to model internals.
CLIP Interrogator~\cite{pharmapsychotic2022clipinterrogator} combines an image captioning backbone with nearest-neighbor retrieval in the CLIP embedding space.
VGD~\cite{kim2025visually} leverages a large language model for autoregressive prompt generation, incorporating CLIP-score guidance during beam-search decoding to steer each token toward better visual alignment. PRISM~\cite{he2024automated} uses the in-context learning ability of an LLM to iteratively refine a population of candidate prompts. 

\subsection{LLM-guided Evolutionary Prompt Optimization}
Recent work has explored combining LLMs with evolutionary search, including genetic algorithms to automatically optimize discrete, human-readable prompts for downstream tasks.
EvoPrompt~\cite{guo2023connecting} connects LLMs with evolutionary operators (e.g., mutation and crossover) to iteratively improve a population of candidate prompts while maintaining linguistic coherence. Complementary directions refine the evolutionary operators or search objective, such as SPRIG~\cite{zhang2024sprig}, which applies an edit-based genetic algorithm to optimize system prompts from structured components.
Related approaches also formulate prompt search explicitly as a genetic process guided by model feedback, e.g., Genetic Prompt Search~\cite{zhao2023genetic}, which uses language model probabilities to steer token-level mutations in discrete prompt tuning.
These methods demonstrate that evolutionary optimization can effectively explore the combinatorial space of prompts.

\subsection{Vision-Language Models for Image Understanding}
VLMs bridge visual and textual modalities and have become the dominant paradigm for image captioning and visual question answering~\cite{li2025survey}.
BLIP-2~\cite{Li2023BLIP2BL} introduces a lightweight Q-Former module that connects a frozen image encoder to a frozen large language model.
InstructBLIP~\cite{dai2023instructblip} extends this framework with instruction-aware visual feature extraction, improving zero-shot generalization.
The Qwen-VL family~\cite{bai2023qwen} integrates visual perception directly into large language models, supporting multi-image understanding and grounded generation.
VLMs are natural candidates for prompt inversion - given an image, a VLM can generate a rich textual description in a single forward pass. However, a single-shot prompt, no matter how detailed, rarely captures the precise combination of content, style, and composition needed for faithful T2I reconstruction.
\ourmethod{} addresses this limitation by using the VLM to generate an initial population of diverse T2I prompts, which are then iteratively refined through a genetic algorithm.

\begin{figure*}[t]
\centering
\includegraphics[width=\textwidth]{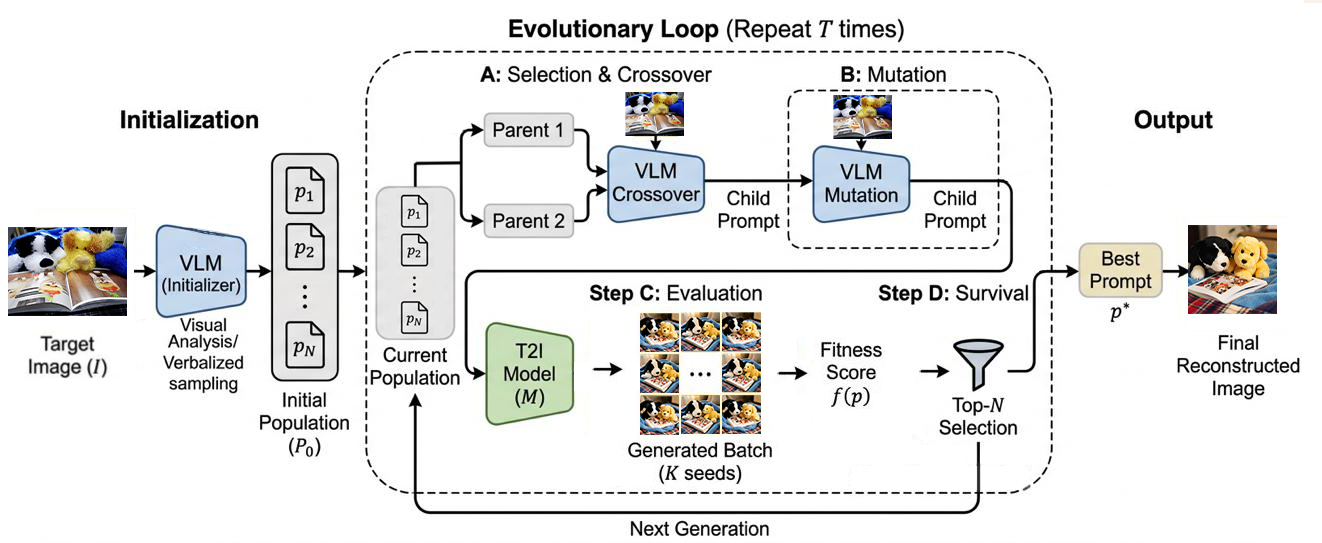}
\caption{Overview of \ourmethod{}. Given a target image, a VLM generates an initial population of $N$ diverse candidate prompts. The evolutionary loop then repeats for $T$ generations: (\textbf{A})~two parents are selected via tournament selection and combined by the VLM into a child prompt through \emph{crossover}, conditioned on the target image; (\textbf{B})~the child optionally undergoes \emph{mutation}, where the VLM applies targeted edits; (\textbf{C})~the child prompt is evaluated by generating $K$ images via the T2I model and computing the mean image-to-image similarity to the target; (\textbf{D})~the top-$N$ individuals from parents and offspring are retained. After $T$ generations, the highest-fitness prompt and its best reconstruction are returned.}
\label{fig:method_overview}
\end{figure*}

\section{Method}\label{sec:method}

Given a text-to-image generation model $\mathcal{M}$, a target image $I$, and an image-to-image similarity score $s$, our goal is to find a natural-language prompt $p^{*}$ that maximize the similarity $s$:
\begin{equation}\label{eq:objective}
    p^{*} = \arg\max_{p} \; \mathbb{E}_{\hat{I}\sim \mathcal{M}(p)}\bigl[s(I,\hat{I})\bigr],
\end{equation}
where the expectation is over the stochastic sampling of $\mathcal{M}$.
Importantly, unlike image captioning\cite{stefanini2022show}, our objective evaluates prompts \emph{indirectly}, through the images they produce, rather than measuring textual accuracy.
While it is desirable for $p^{*}$ to remain human-readable, the prompts our method generates are typically more verbose than standard captions, as they must encode the precise visual details needed for faithful reconstruction (see examples in the supplementary material). Importantly, we note that \ourmethod{} is agnostic to the choice of $\mathcal{M}$ and $s$: it treats the generator as a black box and requires only rendered image output.\\

We optimize the prompt using a genetic algorithm (GA)\cite{srinivas2002genetic}.
An overview of the full pipeline is shown in Fig.~\ref{fig:method_overview}.
The algorithm proceeds as follows:
\begin{enumerate}
    \item \textbf{Initialize} a diverse population of $N$ candidate prompts by querying a vision-language model (VLM) with the target image.
    \item \textbf{For each generation} $t = 1, \dots, T$:
    \begin{enumerate}
        \item \textbf{Crossover}: select two parents and use the VLM to combine them into a child prompt while viewing the target image.
        \item \textbf{Mutation}: with probability $p_m$, use the VLM to apply targeted edits to the child.
        \item \textbf{Evaluation}: generate $K$ random images from the child prompt and compute the mean image-to-image similarity to the target.
        \item \textbf{Selection}: Retain the top-$N$ individuals by fitness from the set of parents and offsprings.
    \end{enumerate}
\end{enumerate}
We now describe each step in detail.

\paragraph{Initialization.}
\label{par:init}
To start the evolutionary search with a diverse yet relevant initial population, we employ \emph{verbalized sampling}\cite{zhang2025verbalized}: a single VLM call that receives the target image and a structured analysis template.
The template instructs the VLM to first perform a coarse analysis pass: identifying subjects, actions, settings, spatial layout, camera angle, lighting, and style, followed by a fine pass that captures small objects, materials, textures, specific colors, patterns, and atmospheric effects. Based on this two-pass analysis, the VLM is asked to produce $N$ distinct prompts in a single generation call, each approximately $50$-$60$ words.
To encourage diversity, the VLM is asked that each prompt must open with a different visual element, and vary sentence order and descriptive style (ranging from terse and factual to atmospheric).

\paragraph{Crossover.}
\label{par:crossover}
Crossover produces a child prompt by combining two parent prompts under the visual guidance of the target image. Parents are selected via tournament selection \cite{goldberg1991comparative} with tournament size $2$: Each parent is chosen by first drawing two individuals uniformly at random from the population, and the one with higher fitness is selected. The VLM then receives both parents alongside the target image and follows a structured five-step reasoning process:
(1)~identify elements \emph{shared} between the two prompts, which are likely accurate and should be preserved;
(2)~identify elements that \emph{differ};
(3)~inspect the target image to determine which variant of each differing element is more faithful, paying special attention to spatial descriptions;
(4)~combine the shared elements with the most accurate differing elements into a single coherent prompt;
(5)~ensure the result flows naturally. See full prompts in the supplementary material.
By reconciling differences using the target image, crossover acts as a form of \emph{visual arbitration}: it retains the consensus of the population while resolving disagreements through direct visual verification. We note that we tried adding the image generated by each prompt to the crossover process, but we did not find it to have any positive effect.

\paragraph{Mutation.}
\label{par:mutation}
With probability $p_m$, the child produced by crossover undergoes mutation.
The VLM receives the child prompt and the target image, and follows a six-step protocol:
(1)~compare each claim in the prompt against the image and note inaccuracies;
(2)~identify important visual elements that are missing;
(3)~verify the spatial accuracy of object placements;
(4)~assess whether any descriptions are too vague;
(5)~consider more precise alternatives for key terms;
(6)~apply $1$--$3$ targeted changes that correct errors, fill gaps, or improve specificity, while leaving already accurate elements untouched.
The mutation operator uses a higher sampling temperature ($0.9$) than the crossover step ($0.7$), to encourage broader exploration of the prompt space.

\paragraph{Fitness evaluation.}
\label{par:fitness}
We evaluate each candidate prompt $p$ by generating $K$ images $\{\hat{I}_k\}_{k=1}^{K}$ from $\mathcal{M}(p)$ and computing the mean similarity:
\begin{equation}\label{eq:fitness}
    f(p) = \frac{1}{K}\sum_{k=1}^{K} s\bigl(I, \hat{I}_k\bigr).
\end{equation}
To avoid redundant computation, we maintain a cache keyed by prompt string, any prompt that reappears across generations (e.g., a parent that survives selection) is looked up directly, skipping redundant text-to-image inference.

\paragraph{Selection.}
\label{par:selection}
After generating $N$ offspring in each generation, we form a combined pool of $2N$ individuals (parents and offspring), sort by fitness, and retain the top-$N$.
This strategy guarantees that the best prompt found so far is never lost, while still allowing superior offspring to displace weaker parents.\\

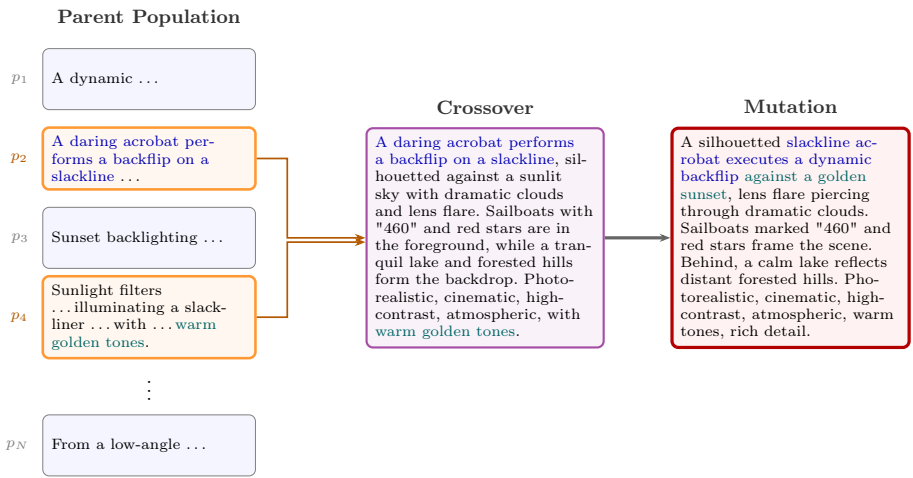
\begin{figure*}[h]
\centering
\resizebox{\textwidth}{!}{%
\begin{tikzpicture}[
    promptbox/.style={
        rectangle, rounded corners=3pt, draw=black!50, fill=blue!4,
        text width=3.2cm, minimum height=1cm, align=left,
        font=\scriptsize, inner sep=4pt
    },
    selectedbox/.style={
        promptbox, draw=orange!80, fill=orange!6, line width=1.2pt
    },
    childbox/.style={
        promptbox, draw=violet!70, fill=violet!5, line width=1pt,
        text width=3.6cm
    },
    starbox/.style={
        rectangle, rounded corners=3pt, draw=red!70!black, fill=red!5,
        text width=3.6cm, minimum height=1cm, align=left,
        font=\scriptsize, inner sep=4pt, line width=1.5pt
    },
    opbox/.style={
        rectangle, rounded corners=5pt, draw=black!60, fill=gray!10,
        minimum width=1.8cm, minimum height=0.7cm, align=center,
        font=\small\bfseries
    },
    arr/.style={-{Stealth[length=5pt]}, thick, black!60},
    selarr/.style={-{Stealth[length=5pt]}, thick, orange!70!black},
    sublabel/.style={font=\scriptsize\itshape, text=black!50},
    plabel/.style={font=\scriptsize, text=black!50},
]

\node[font=\footnotesize\bfseries, text=black!80] (poptitle) at (0, 3.0) {Parent Population};

\node[promptbox]   (p1) at (0,  2.0) {A dynamic \ldots};
\node[selectedbox] (p2) at (0,  0.7) {\textcolor{blue!70!black}{A daring acrobat performs a backflip on a slackline} \ldots};
\node[promptbox]   (p3) at (0, -0.6) {Sunset backlighting \ldots};
\node[selectedbox] (p4) at (0, -1.9) {Sunlight filters \ldots illuminating a slackliner \ldots with \ldots \textcolor{teal!80!black}{warm golden tones}.};
\node at (0, -3.0) {$\vdots$};
\node[promptbox]   (p5) at (0, -4.0) {From a low-angle \ldots};

\node[plabel, anchor=east] at ([xshift=-3pt]p1.west) {$p_1$};
\node[plabel, text=orange!70!black, font=\scriptsize\bfseries, anchor=east] at ([xshift=-3pt]p2.west) {$p_2$};
\node[plabel, anchor=east] at ([xshift=-3pt]p3.west) {$p_3$};
\node[plabel, text=orange!70!black, font=\scriptsize\bfseries, anchor=east] at ([xshift=-3pt]p4.west) {$p_4$};
\node[plabel, anchor=east] at ([xshift=-3pt]p5.west) {$p_N$};

\node[childbox] (child) at (5.5, -0.6) {\textcolor{blue!70!black}{A daring acrobat performs a backflip on a slackline}, silhouetted against a sunlit sky with dramatic clouds and lens flare. Sailboats with "460" and red stars are in the foreground, while a tranquil lake and forested hills form the backdrop. Photorealistic, cinematic, high-contrast, atmospheric, with \textcolor{teal!80!black}{warm golden tones}.};

\node[font=\footnotesize\bfseries, text=black!80, anchor=south] at (child.north) [yshift=3pt] {Crossover};
\draw[selarr] (p2.east) -- ++(0.5,0) |- (child.west);
\draw[selarr] (p4.east) -- ++(0.5,0) |- ([yshift=-2pt]child.west);

\node[starbox] (pstar) at (10.5, -0.6) {A silhouetted \textcolor{blue!70!black}{slackline acrobat executes a dynamic backflip} \textcolor{teal!80!black}{against a golden sunset}, lens flare piercing through dramatic clouds. Sailboats marked "460" and red stars frame the scene. Behind, a calm lake reflects distant forested hills. Photorealistic, cinematic, high-contrast, atmospheric, warm tones, rich detail.};
\node[font=\footnotesize\bfseries, text=black!80, anchor=south] at (pstar.north) [yshift=3pt] {Mutation};

\draw[arr, line width=1.5pt] (child.east) -- (pstar.west);

\end{tikzpicture}
}%
\caption{Prompt evolution flow example in \ourmethod{}. A VLM generates an initial population of $N$ candidate prompts. Each generation, two parents (orange) are selected, combined through \emph{crossover}, and optionally refined via \emph{mutation}, where all VLM operations are guided by the target image.}
\label{fig:prompt_flow}
\end{figure*}

Finally, the prompt that has the best fitness score at the last generation is returned as the algorithms output. We show an example one such iteration in Fig. \ref{fig:prompt_flow}. This prompt is from the leftmost image in Fig. \ref{fig:intro_examples} for reference.

\section{Experiments}
\label{sec: experiments}
In this section, we compare \ourmethod{} with various methods from the prompt inversion domain. We use a variety of datasets and setups to further demonstrate the effectiveness of our approach. Our approach was based on Qwen3-VL-Instruct as the VLM, and FLUX.2-klein-4B as our T2I model. Code, further experimental details,  further examples of output prompts, and generated images, as well as evaluation of generated prompt naturalness, are provided in the supplementary. 

\paragraph{\textbf{Baselines.}} We compare \ourmethod{} with natural baselines from recent prompt inversion works. The compared methods include (1) VLM-Baseline, where we utilize Qwen3-VL~\cite{bai2025qwen3}, prompting it to generate relevant texts for the text-to-image model, then select the prompt that produces the highest-scoring images. (2) VGD~\cite{kim2025visually}, a gradient-free approach that produces human-readable prompts by using LLM and CLIP score guidance. (3) STEPS~\cite{qiu2025steps}, reformulates discrete prompt optimization as sequential probability tensor estimation to enable efficient search in high-dimensional token space. (4) PEZ~\cite{mahajan2024prompting}, which discovers hard prompts by maximizing the cosine similarity between prompt embeddings and target image features in the CLIP latent space. (5) CLIP Interrogator~\cite{pharmapsychotic2022clipinterrogator}, generates descriptive prompts for an input image by combining image captioning models with the CLIP embedding space. We note that previous methods were trained on different T2I models, and therefore, we cannot compare image quality directly. Instead, we compare faithfulness, how much the generated image matches the target image in the desired score. We also note that the VLM-Baseline and \ourmethod{} generate and evaluate the same number of prompts (60 prompts in total, \ourmethod{} generates 10 prompts per generation) with the same T2I model with verbalized sampling \cite{zhang2025verbalized} to ensure diversity. The difference is that the VLM-Baseline generates them all at once (similar to our initialization step), while \ourmethod{} generates them iteratively over generations. \\

\paragraph{\textbf{Datasets.}} To ensure a comprehensive evaluation across datasets with diverse characteristics, we utilize the following datasets: Lexica \cite{SQBZ24}, MS-COCO~\cite{lin2014microsoft},  Celeb-A~\cite{liu2015deep}, Flickr8K \cite{hodosh2013framing}, and LAION-400M~\cite{schuhmann2021laion}. These datasets cover diverse data distributions, both natural images and AI-generated, and span object-centric scenes to portrait imagery, with varying visual and semantic complexity. We follow~\cite{mahajan2024prompting, kim2025visually} and randomly sample $100$ images from the test split of each dataset.

\paragraph{\textbf{Evaluation.}}
We report standard metrics commonly used in the prompt inversion domain. Since our objective is to generate a prompt that reconstructs a given image, we evaluate multiple image-to-image similarity measures. In particular, we report the cosine similarity between image features extracted using several CLIP models. We note that the baselines optimize the CLIP similarity score directly. However, as is common practice, the CLIP model used for evaluation is different from the model used for optimization. Additionally, we report the BLIP~\cite{li2023blip} score, which measures text–image similarity via the normalized L2 distance. Finally, we consider DreamSim~\cite{fu2023dreamsim}, a learned perceptual similarity score that embeds both the target and generated images using a fused representation from pretrained vision models (e.g., CLIP~\cite{radford2021learning}, OpenCLIP~\cite{ilharco_gabriel_2021_5143773}, and DINO~\cite{caron2021emerging}) and compares the resulting embeddings to produce a similarity score aligned with human judgments.

\begin{table*}[h]
\centering
\caption{Quantitative comparison across datasets. We report CLIP, BLIP and DreamSim image-to-image similarities between the original and reconstructed images.}
\label{tab:main_results}
\adjustbox{max width=\textwidth}{%
\begin{tabular}{l ccc ccc ccc ccc ccc}
\toprule
& \multicolumn{3}{c}{\textbf{Lexica}} & \multicolumn{3}{c}{\textbf{COCO}} & \multicolumn{3}{c}{\textbf{CelebA}} & \multicolumn{3}{c}{\textbf{Flickr8K}} & \multicolumn{3}{c}{\textbf{LAION-400M}} \\
\cmidrule(lr){2-4} \cmidrule(lr){5-7} \cmidrule(lr){8-10} \cmidrule(lr){11-13} \cmidrule(lr){14-16}
 & CLIP & BLIP & DreamSim & CLIP & BLIP & DreamSim & CLIP & BLIP & DreamSim & CLIP & BLIP & DreamSim & CLIP & BLIP & DreamSim \\
\midrule
PEZ                   & 0.75 & 0.77 & 0.70 & 0.73 & 0.67 & 0.66 & 0.64 & 0.66 & 0.68 & 0.74 & 0.69 & 0.66 & 0.63 & 0.63 & 0.63 \\
VGD                   & 0.77 & 0.78 & 0.70 & 0.73 & 0.68 & 0.66 & 0.65 & 0.68 & 0.68 & 0.75 & 0.72 & 0.67 & 0.61 & 0.60 & 0.62 \\
CLIP Interrogator     & 0.79 & 0.80 & 0.72 & 0.74 & 0.72 & 0.67 & 0.68 & 0.72 & 0.71 & 0.78 & 0.74 & 0.67 & 0.66 & 0.68 & 0.65 \\
STEPS                 & 0.75 & 0.78 & 0.71 & 0.76 & 0.70 & 0.68 & 0.65 & 0.68 & 0.69 & 0.77 & 0.72 & 0.68 & 0.65 & 0.66 & 0.65 \\
\midrule
VLM-Baseline & 0.77 & 0.85 & 0.78 & 0.79 & 0.85 & 0.76 & 0.64 & 0.79 & 0.76 & 0.77 & 0.85 & 0.76 & 0.79 & 0.84 & 0.78 \\
\midrule

\ourmethod{}-CLIP     & \textbf{0.82} & 0.86 & 0.77 & \textbf{0.83} & 0.86 & 0.75 & \textbf{0.69} & 0.80 & 0.75 & \textbf{0.82} & 0.85 & 0.75 & \textbf{0.84} & 0.85 & 0.77 \\
\ourmethod{}-BLIP     & 0.78 & \textbf{0.89} & 0.77 & 0.79 & \textbf{0.89} & 0.76 & 0.64 & \textbf{0.83} & 0.75 & 0.78 & \textbf{0.88} & 0.76 & 0.80 & \textbf{0.87} & 0.78 \\
\ourmethod{}-DreamSim & 0.78 & 0.87 & \textbf{0.79} & 0.79 & 0.87 & \textbf{0.78} & 0.65 & 0.81 & \textbf{0.77} & 0.78 & 0.86 & \textbf{0.77} & 0.80 & 0.85 & \textbf{0.80} \\
\ourmethod{}-OpenCLIP & 0.79 & 0.86 & 0.77 & 0.80 & 0.86 & 0.75 & 0.65 & 0.80 & 0.75 & 0.79 & 0.85 & 0.75 & 0.81 & 0.85 & 0.77 \\

\bottomrule
\end{tabular}
}%
\end{table*}

\subsection{Generation Results}
\label{sec:exp_generation_results}
We show the experimental results on several benchmarks in Tab. \ref{tab:main_results}. We first observe that the simple VLM-Baseline is on par with or outperforms all previous methods, even on the CLIP score they were directly optimizing. This is probably due to the fact that it has access to a stronger VLM model (Qwen3-VL) that outperforms their VLM/LLM backbone even on a task they were specifically optimized on. We then see that \ourmethod{} outperforms this baseline across all datasets and score functions. We also note that VLM-Baseline and \ourmethod{} outperform previous approaches on scores they were not specifically optimizing. We suspect that this is because it is less reliant on the score function, using it only for selection, and as such, finds solutions that work well on multiple score functions. \\

Despite that, it is still influenced by the target score and, not surprisingly, the best method for each score function is \ourmethod{} that optimizes that specific score. This highlights a main benefit of \ourmethod{}: it is highly versatile and works well with multiple score functions without any adjustments, training, finetuning, or hyperparameter search.

\begin{figure}[!htbp]
\centering
\includegraphics[width=\textwidth]{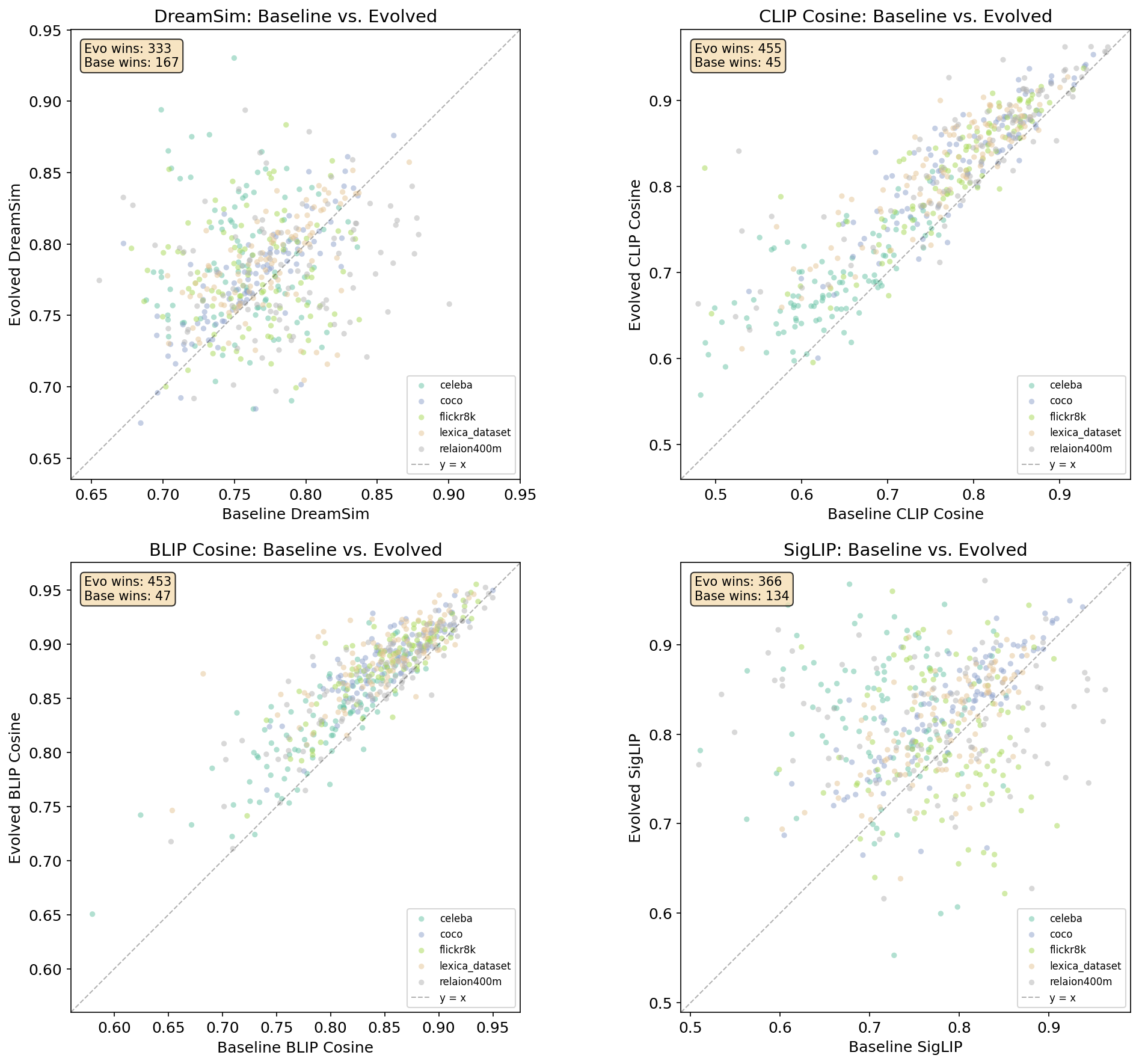}
\caption{Per-image comparison of VLM-Baseline ($x$-axis) versus \ourmethod{} ($y$-axis) across all $500$ evaluation images, for four similarity metrics. Each point is one image, colored by source dataset; the dashed line is $y=x$. Points above the diagonal are images where evolution improved the reconstruction. The legend in the top-left of each panel reports the per-metric win counts (\ourmethod{} vs.\ VLM-Baseline). For each panel we use the \ourmethod{} variant that was guided by that metric.}
\label{fig:scatter_all_metrics}
\end{figure}

\paragraph{\textbf{Per-image analysis.}} The averages reported in Tab.~\ref{tab:main_results} could in principle be driven by a small number of large gains while most images regress. To examine this, we show in Fig.~\ref{fig:scatter_all_metrics} a per-image comparison between VLM-Baseline ($x$-axis) and \ourmethod{} ($y$-axis) across all $500$ evaluation images. Each panel corresponds to one similarity metric, and uses the \ourmethod{} variant that was guided by that metric (e.g., \ourmethod{}-CLIP for the CLIP Cosine panel). Points above the diagonal $y=x$ correspond to images where evolution improved over the baseline. We first observe that \ourmethod{} wins on the substantial majority of images for every metric: $455/500$ for CLIP Cosine, $453/500$ for BLIP Cosine, $366/500$ for OpenCLIP-SigLIP, and $333/500$ for DreamSim. We also see that the wins are distributed across all five datasets, indicating that the improvement is a genuine image-level effect, and not driven by a few outliers. The lower win rate on DreamSim is consistent with the smaller margins observed in Tab.~\ref{tab:main_results} for this metric: the VLM-Baseline caption is often already perceptually close to the target, leaving less headroom to optimize. We note that OpenCLIP-SigLIP is included here as a fourth, optimization-independent metric, and an extended version of Tab.~\ref{tab:main_results} that adds the SigLIP column is provided in Tab.~\ref{tab:extended_openclip}.

\subsection{Human Preference Study}
\label{sec:human_eval}

Since the best competitor was the VLM-Baseline, which can also be applied to the same T2I generation model, we compared reconstruction quality through a human preference study. We recruited 94 unique labelers. For each image, the labeler was presented with both reconstructions and asked to select the one that was semantically closer to the original. Complete experimental details are provided in the supplementary material. In total, we aggregated 956 pairwise comparisons, of which 521 favored \ourmethod{} and 435 favored the VLM-Baseline. A one-sided binomial test confirms that our method is significantly preferred, with a p-value of 0.003.


\subsection{Qualitative Comparison}

\begin{figure}[!htbp]
\centering
\setlength{\tabcolsep}{1.5pt}
\renewcommand{\arraystretch}{0.8}
\newcolumntype{C}{>{\centering\arraybackslash}m{0.175\textwidth}}
\begin{tabular}{m{1.2em}CCCCC}
& \textbf{COCO} & \textbf{Lexica} & \textbf{CelebA} & \textbf{LAION-400M} & \textbf{Flickr8K} \\[2pt]
\rotatebox{90}{\small\textbf{Original}} &
\includegraphics[width=0.175\textwidth]{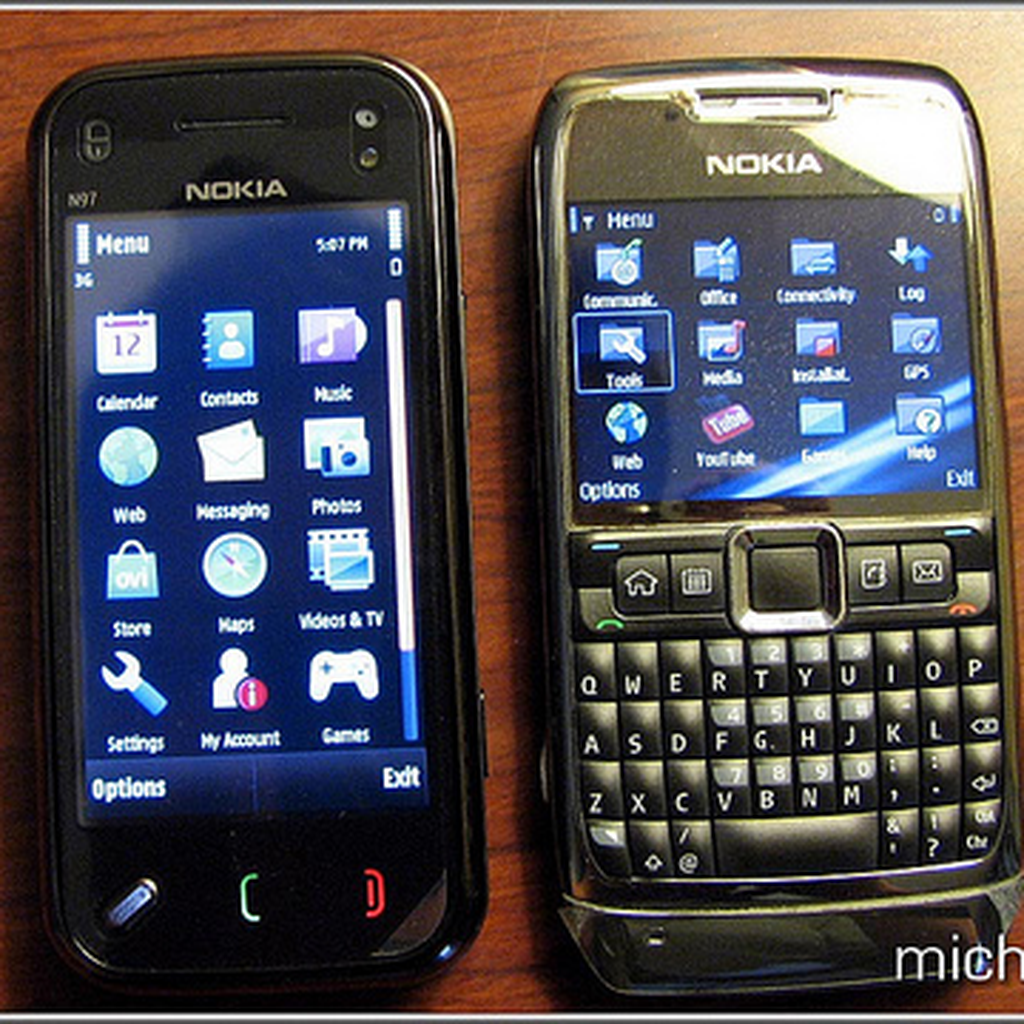} &
\includegraphics[width=0.175\textwidth]{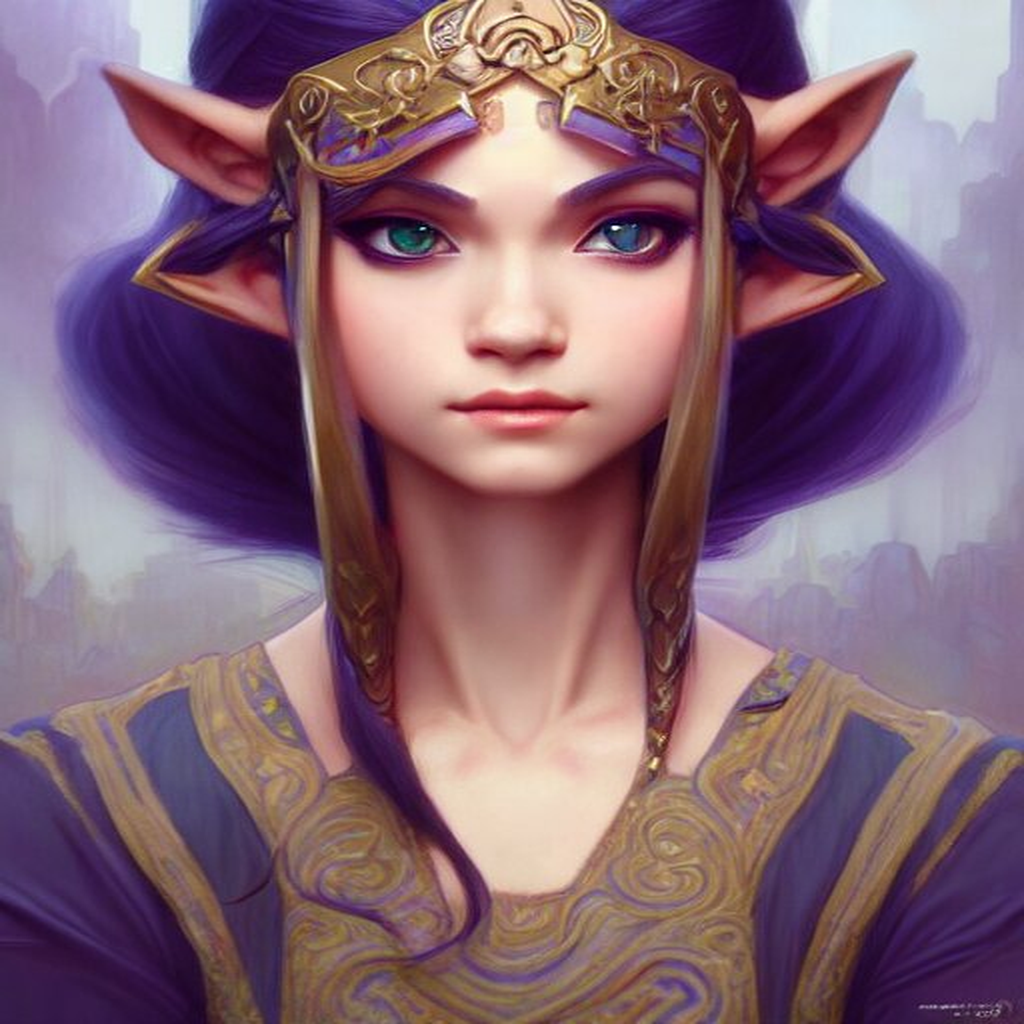} &
\includegraphics[width=0.175\textwidth]{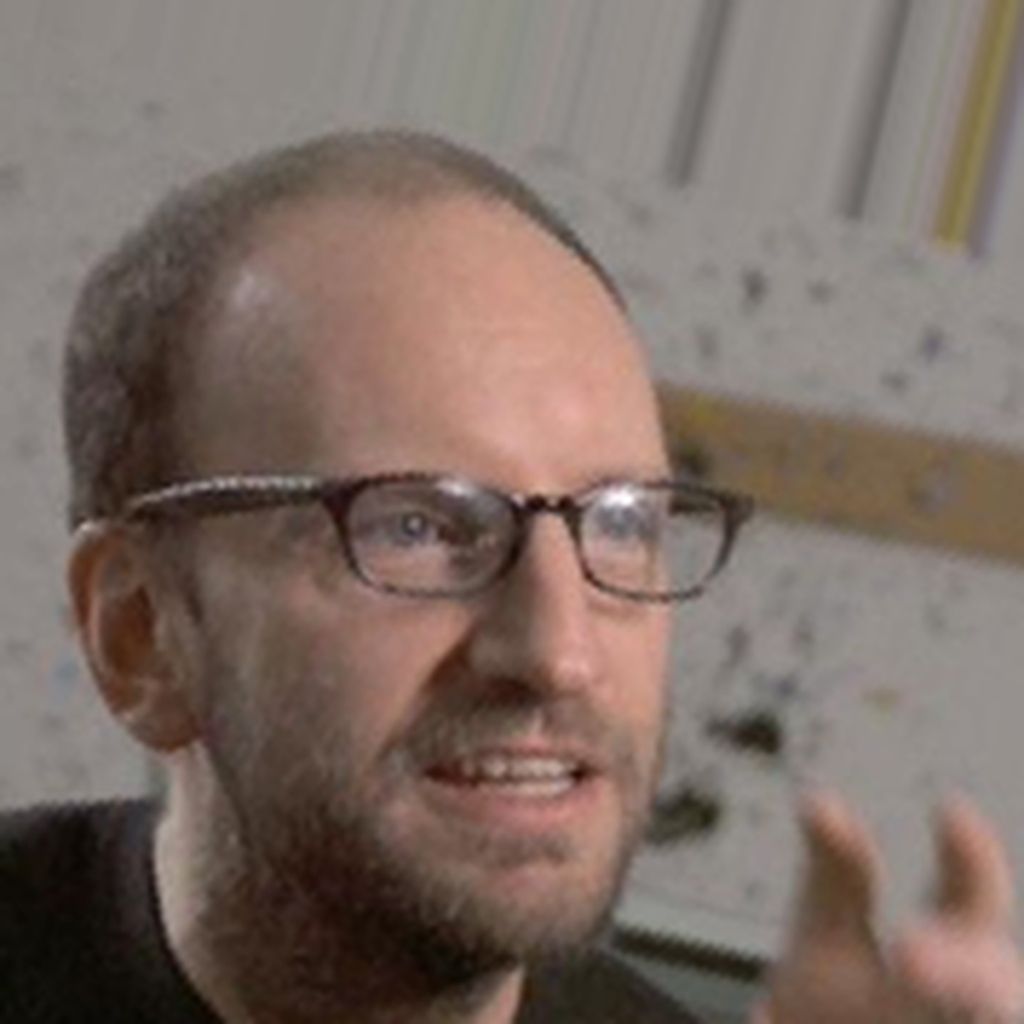} &
\includegraphics[width=0.175\textwidth]{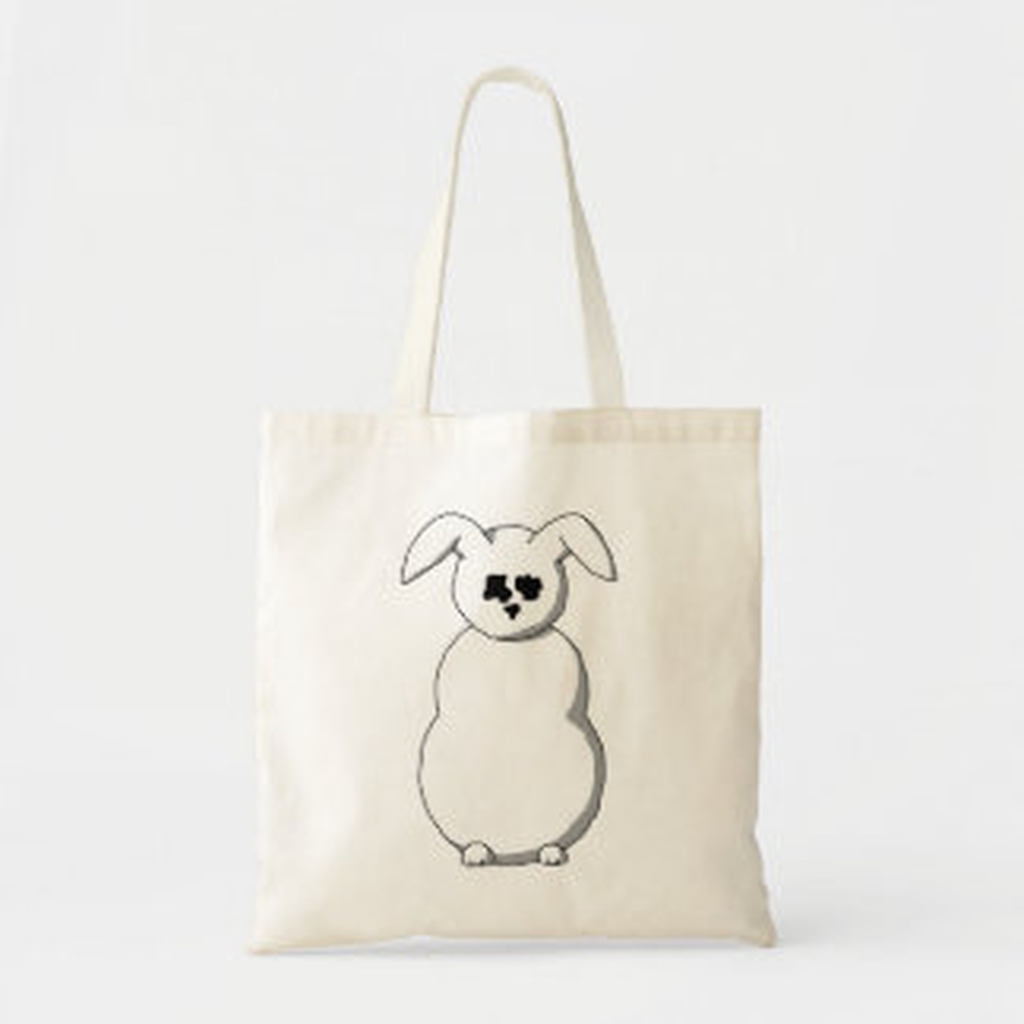} &
\includegraphics[width=0.175\textwidth]{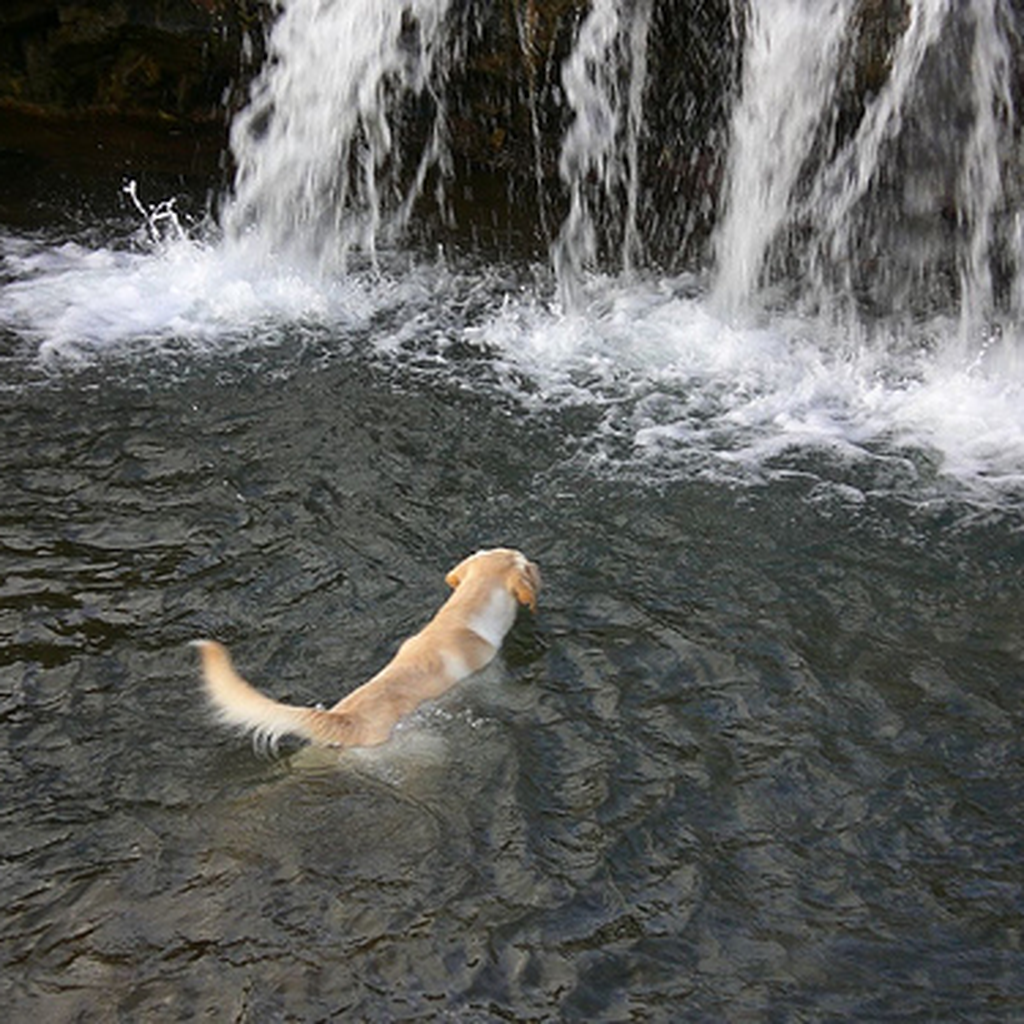} \\[1pt]
\rotatebox{90}{\small\textbf{Ours}} &
\includegraphics[width=0.175\textwidth]{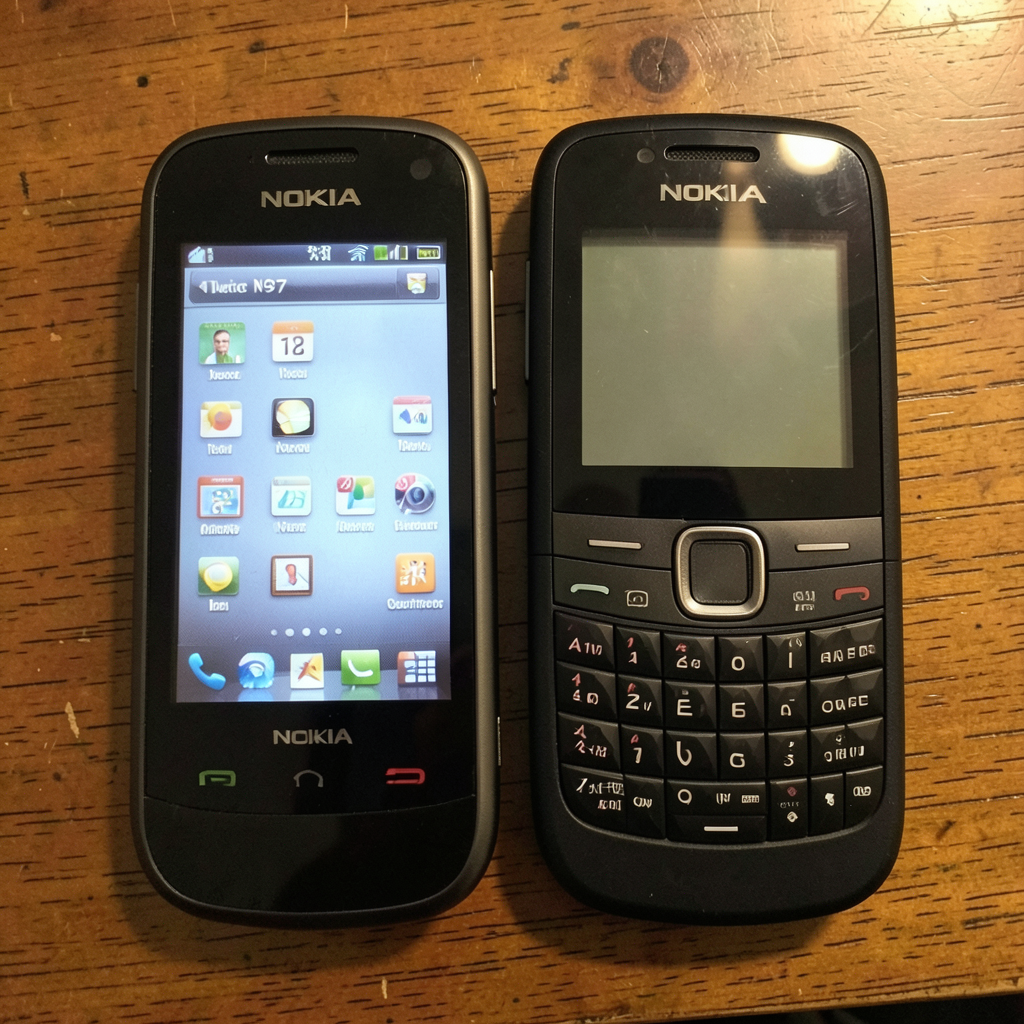} &
\includegraphics[width=0.175\textwidth]{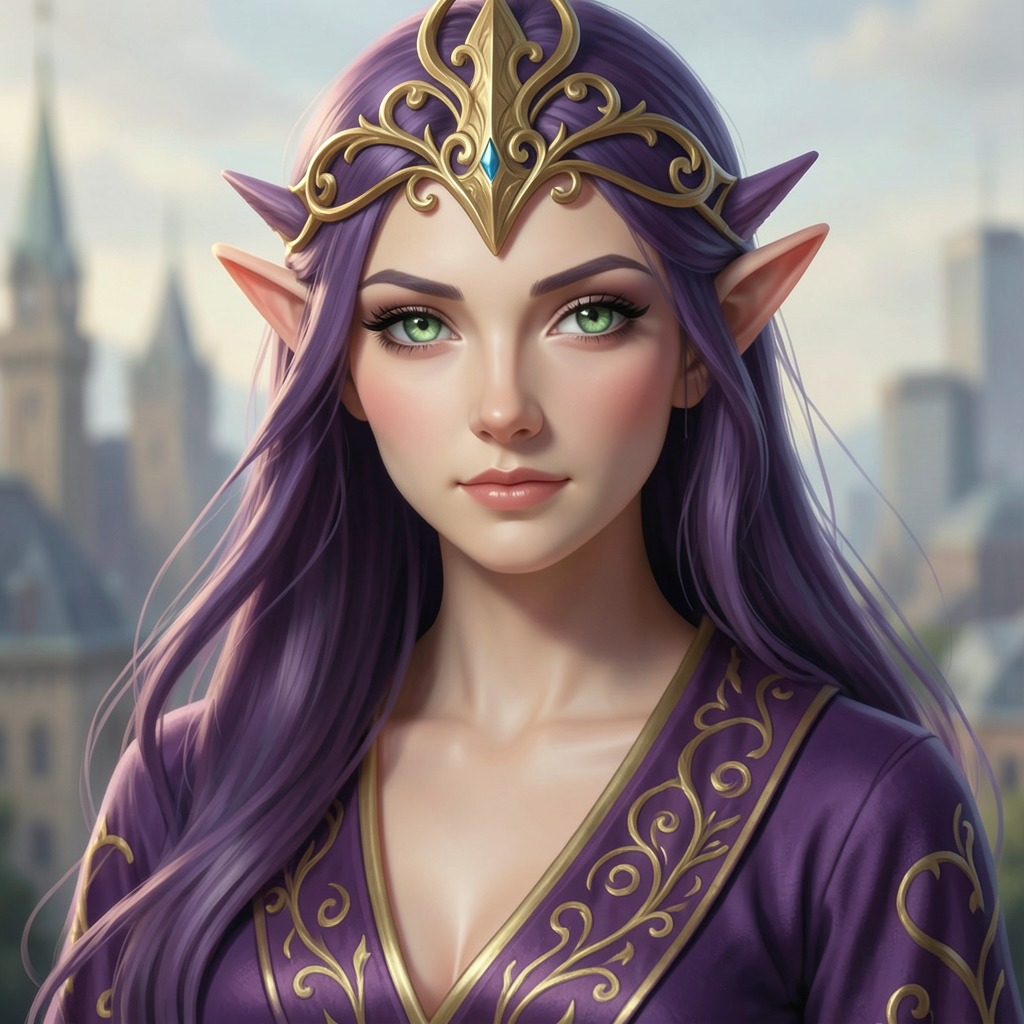} &
\includegraphics[width=0.175\textwidth]{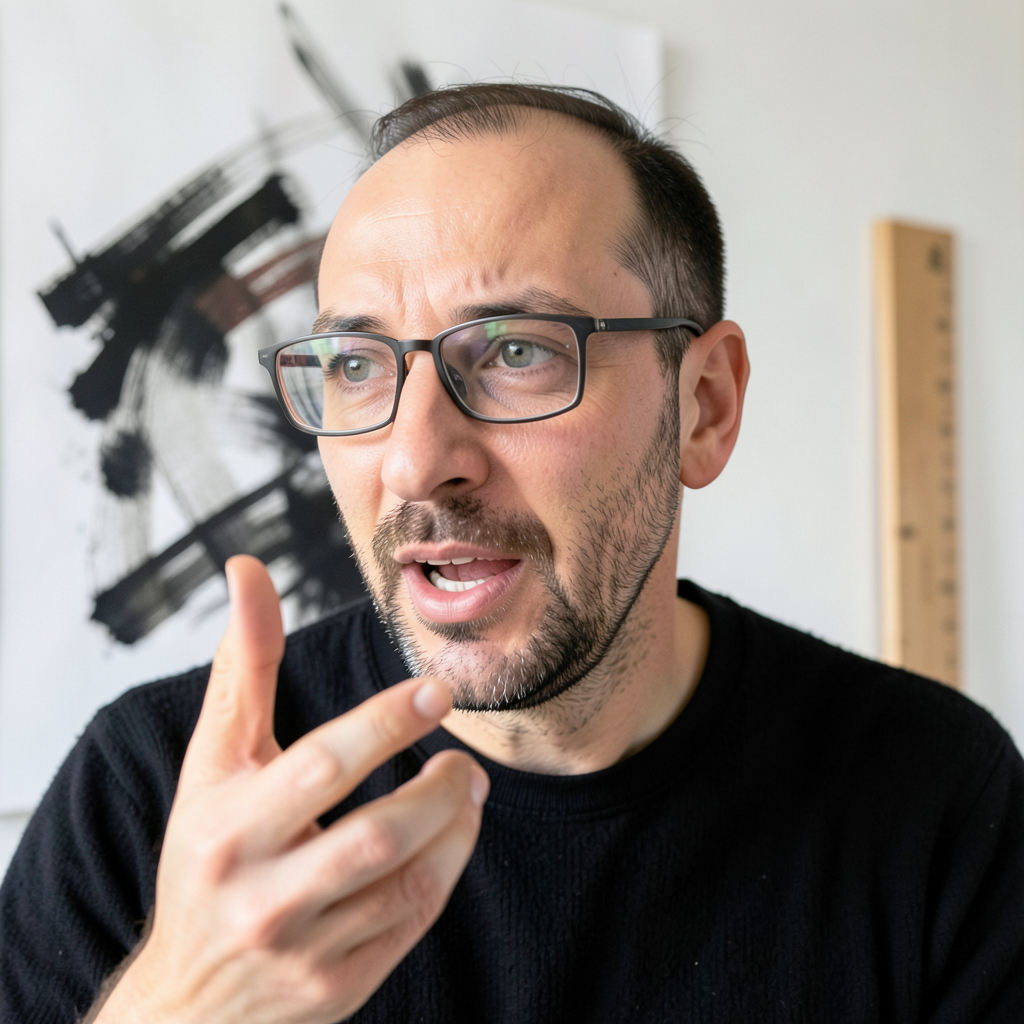} &
\includegraphics[width=0.175\textwidth]{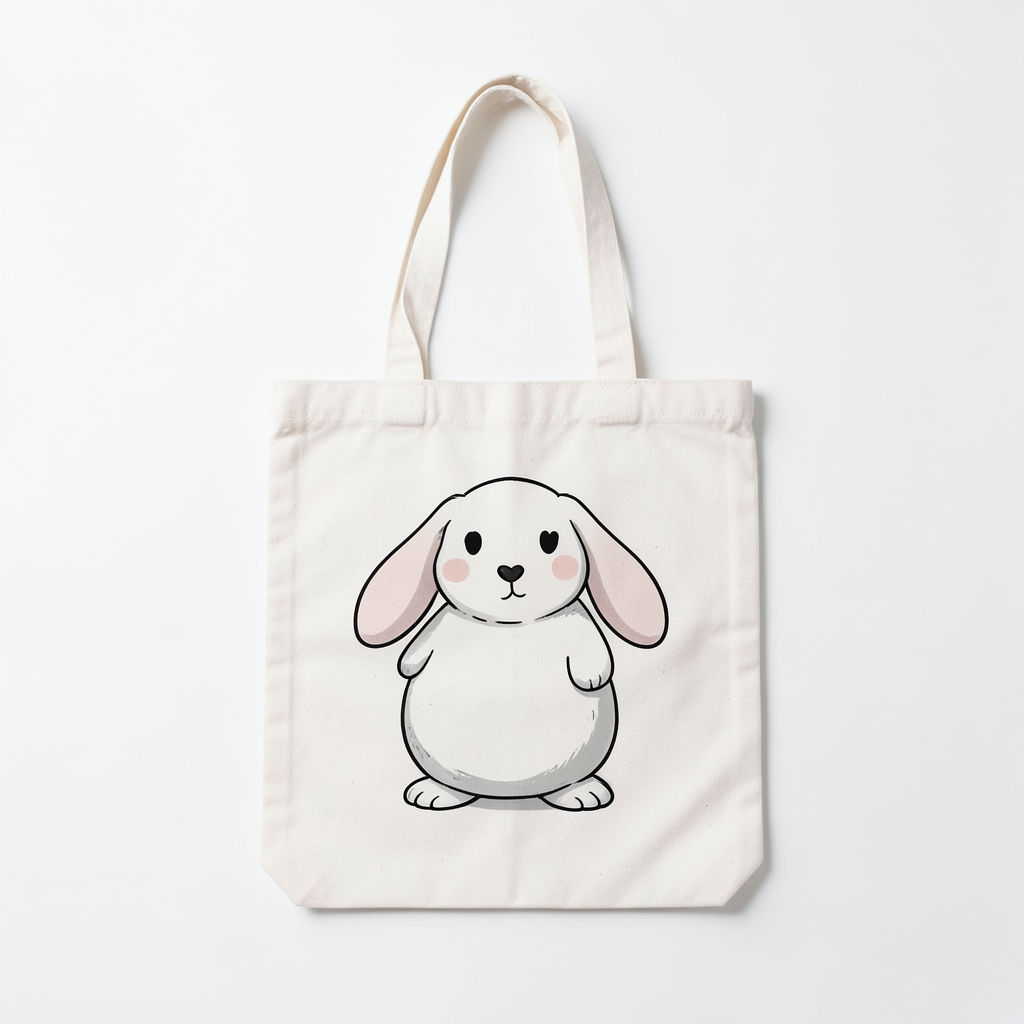} &
\includegraphics[width=0.175\textwidth]{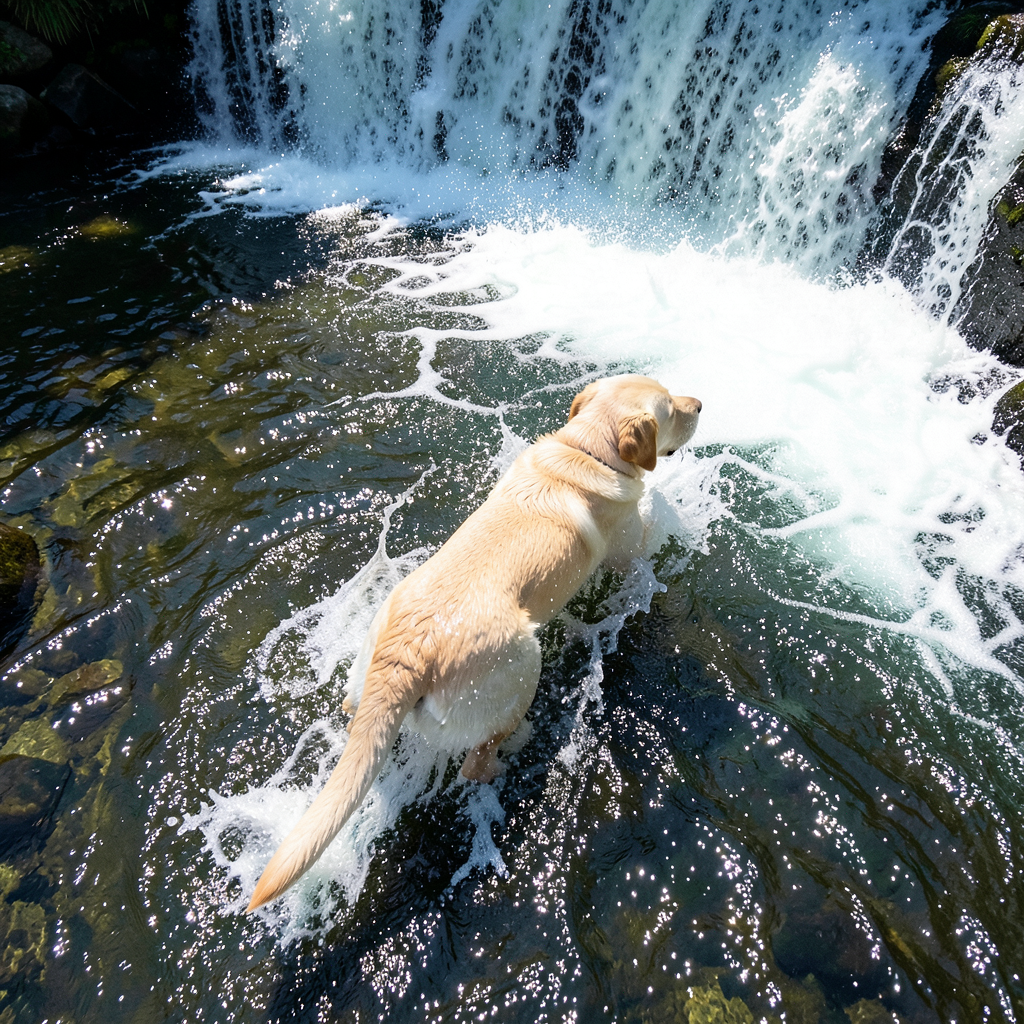} \\[1pt]
\rotatebox{90}{\small\textbf{VLM-Base.}} &
\includegraphics[width=0.175\textwidth]{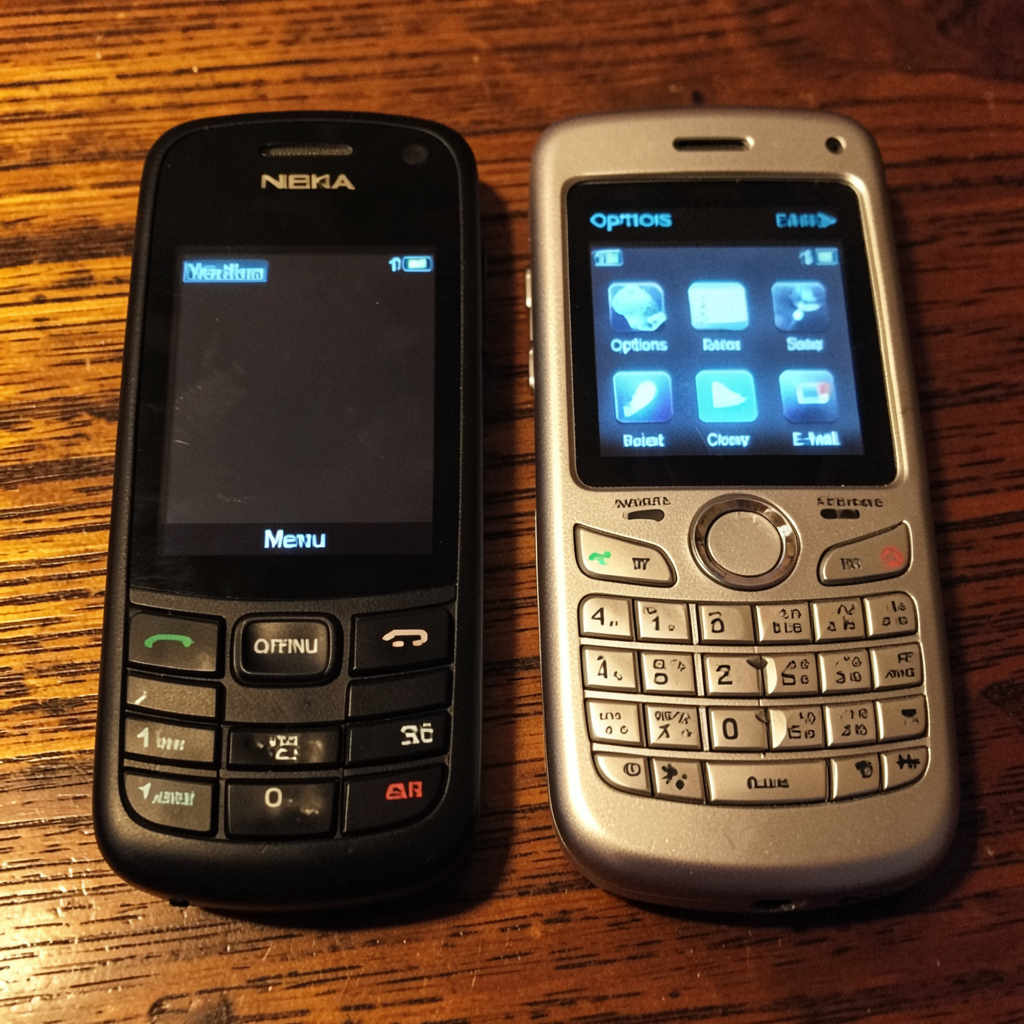} &
\includegraphics[width=0.175\textwidth]{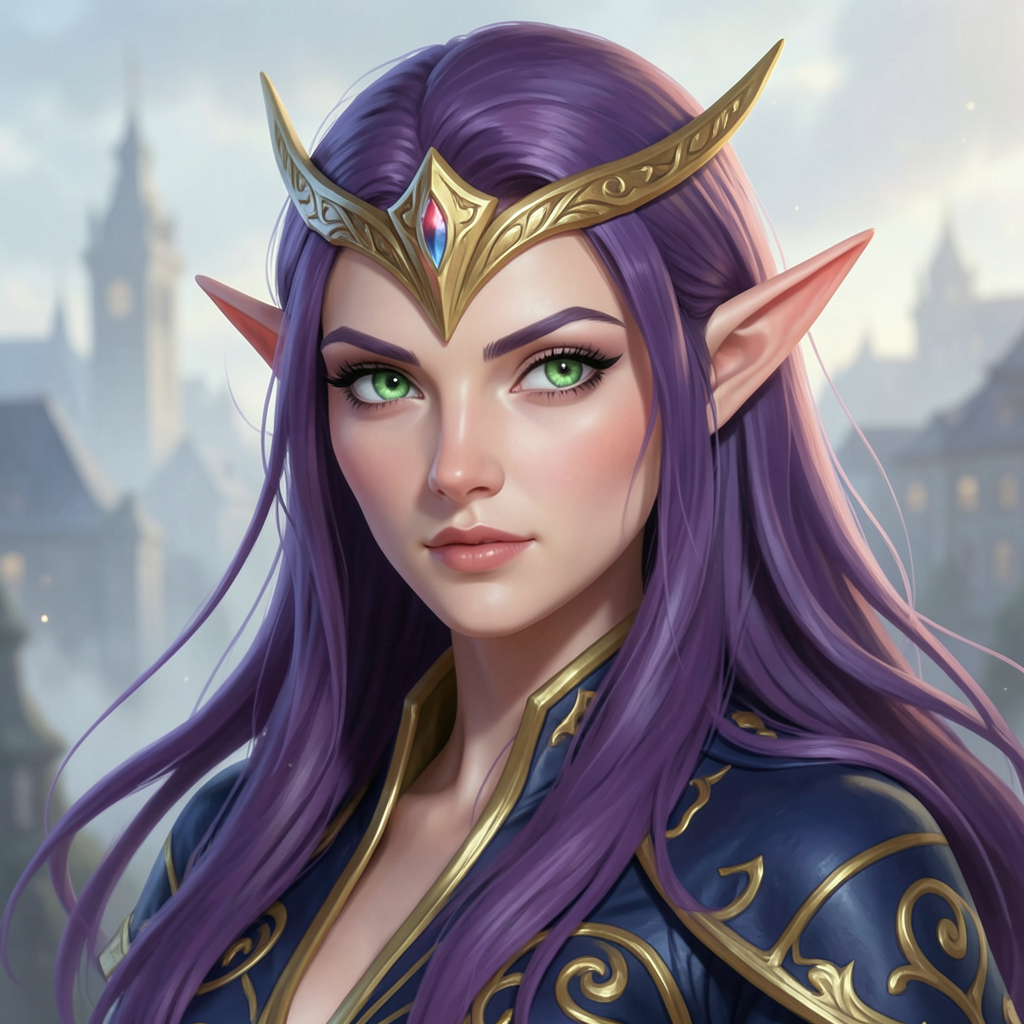} &
\includegraphics[width=0.175\textwidth]{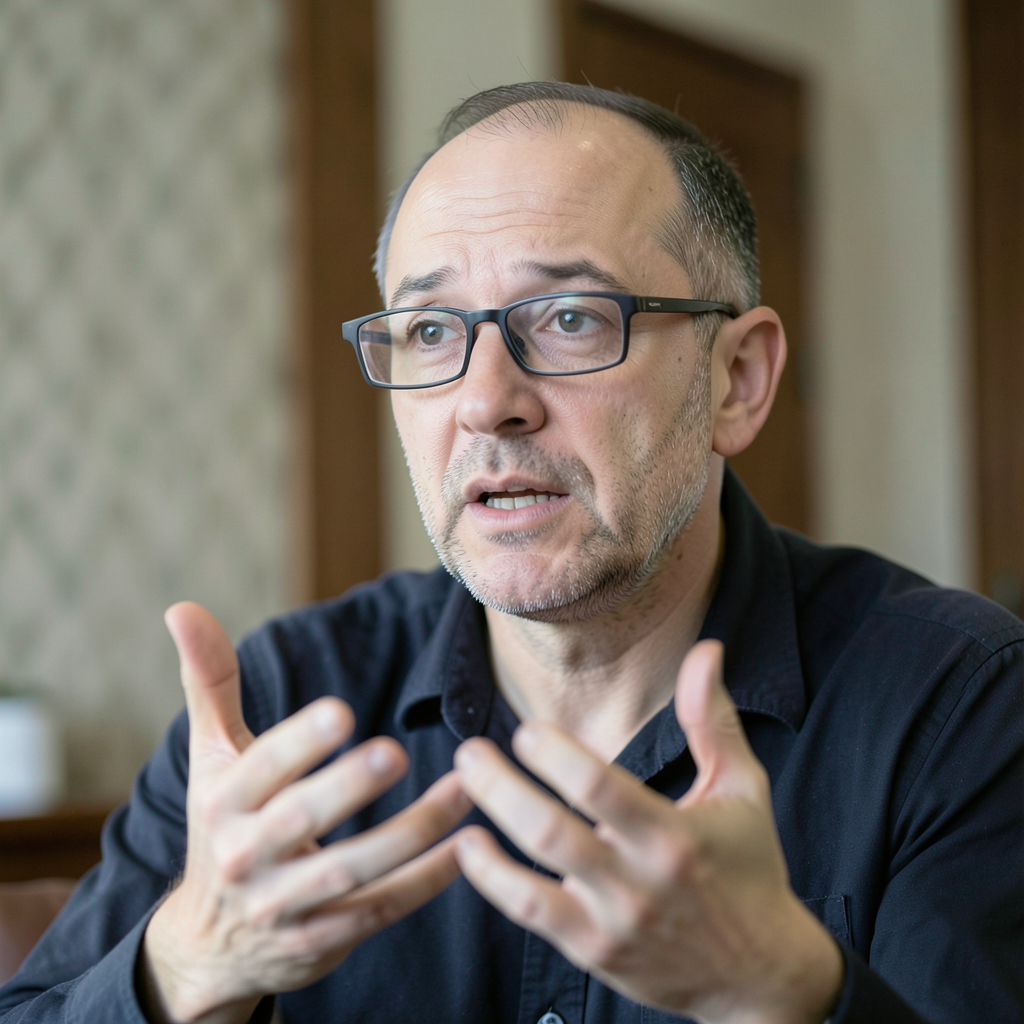} &
\includegraphics[width=0.175\textwidth]{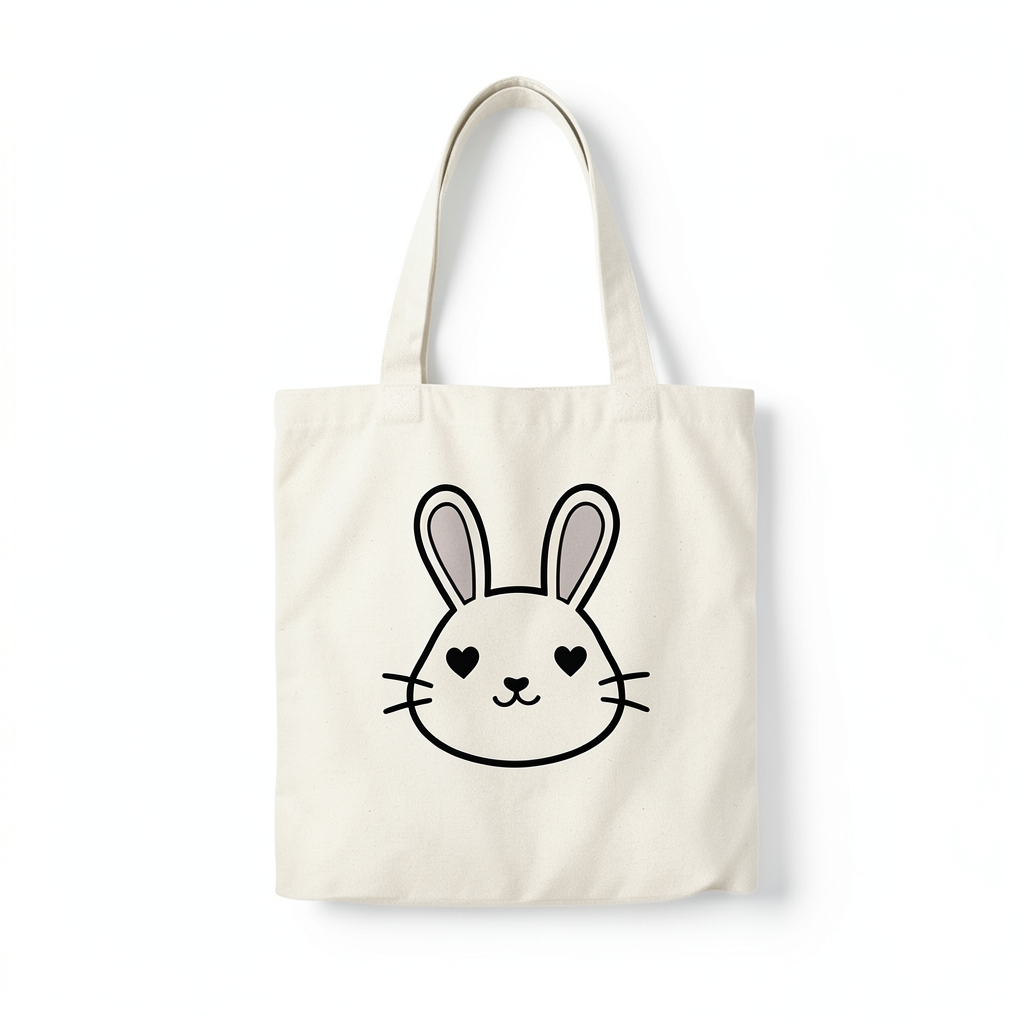} &
\includegraphics[width=0.175\textwidth]{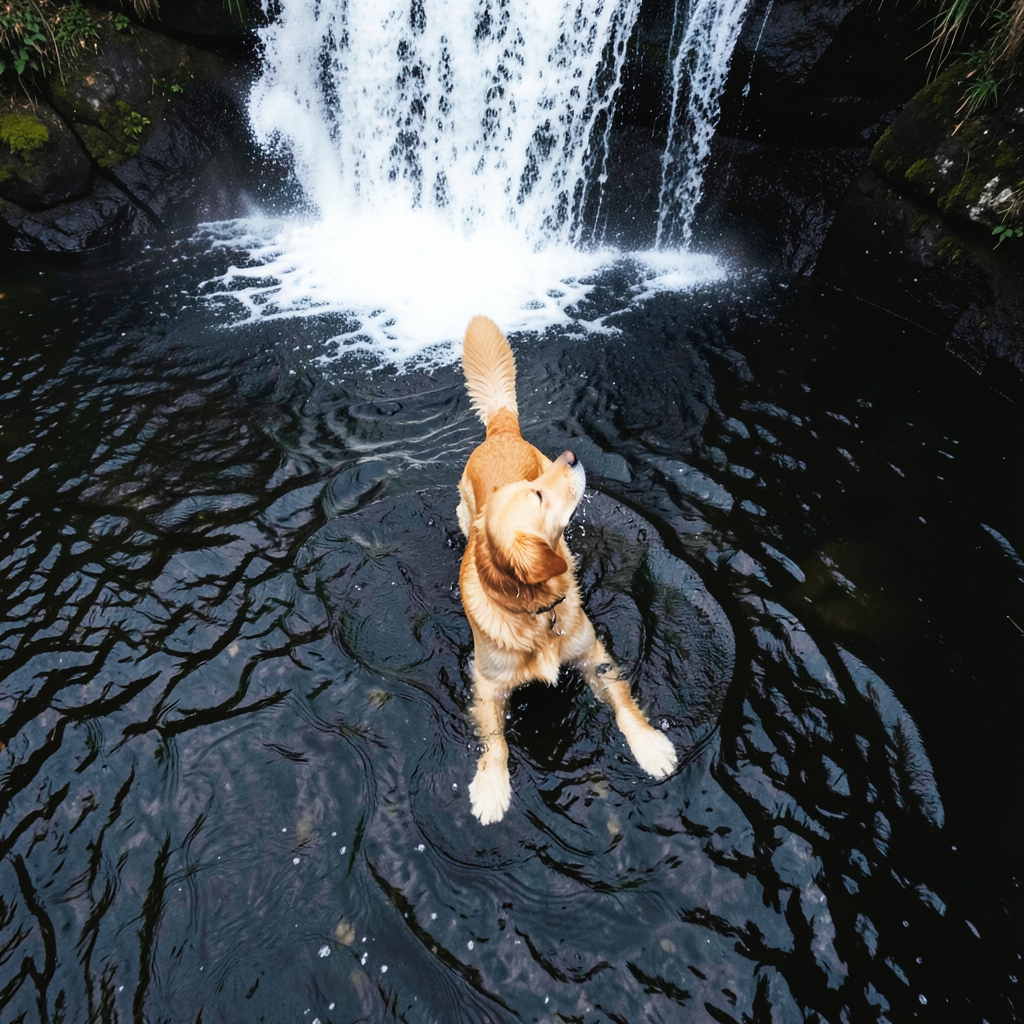} \\
\end{tabular}
\caption{Qualitative comparison across five diverse datasets. Given a target image (top), we show reconstructions from \ourmethod{} (middle) and VLM-Baseline (bottom). Our evolutionary approach produces more faithful reconstructions, better preserving scene composition, style, and fine-grained details.}
\label{fig:intro_comparison}
\end{figure}

We qualitatively compare \ourmethod{} to the strongest baseline, VLM-Baseline. Figure~\ref{fig:intro_comparison} shows one example per dataset. Overall, the baseline often produces a semantically plausible description but drifts from the target in fine details and attribute grounding, while \ourmethod{} better preserves identity cues, object structure, and style, resulting in reconstructions that are more faithful to the original.

We highlight one illustrative case from Flickr8K. The original image shows a dog swimming in a dark green pool near a waterfall, with an oblique viewpoint and a characteristic spread of surface foam. The baseline changes the scene layout and flips the dog to face away from the waterfall. It also pushes the water color toward a uniform black and fails to reproduce the foam spread accurately. In contrast, \ourmethod{} matches the original more closely. It maintains the spatial composition of the dog and the waterfall in the image. Moreover, it recovers the dark green water tone and foam pattern that better resembles the target.

Overall, these examples highlight a consistent gap in inversion fidelity - \ourmethod{} reconstructs images that are closer to the GT in identity, geometry, and fine-grained appearance, whereas the VLM-based baseline often settles for approximate semantics at the expense of precise visual correspondence.

\subsection{Ablation Studies}
We performed ablation studies on the MS-COCO dataset to explore the effect of various parts of our algorithm.  In our first experiment, we explore two variations, using our structured prompts vs using minimalistic, simple prompts for the initialization, crossover, and mutations steps. We also explore two different variants of the VLM, Qwen3-VL-8B-Instruct and Qwen3-VL-8B-Thinking. The results in Tab. \ref{tab:ablation_template} show that while the structured prompt has a positive effect on the CLIP score, the effect is not strong.  We also observe that our structured prompt with the faster Instruct model achieves similar results to the minimal prompt on the thinking variant, which takes, however, more than twice the runtime.

\begin{table}[h]
\centering
\caption{Ablation: VLM template style vs.\ reasoning mode. We compare structured and minimal prompting templates with instruct and thinking VLM variants, reporting CLIP similarity and runtime (seconds per image).}
\label{tab:ablation_template}
\begin{tabular}{lc cc}
\toprule
\textbf{Template} & \textbf{VLM Variant} & \textbf{CLIP}$\uparrow$ & \textbf{Runtime (s)}$\downarrow$ \\
\midrule
Structured & Instruct  & $0.83 \pm 0.007$ & \textbf{381.18} \\
Structured & Thinking  & $0.83 \pm 0.007$ & 885.6 \\
Minimal    & Instruct  & $0.82 \pm 0.007$ & 420.72 \\
Minimal    & Thinking  & $0.83 \pm 0.007$ & 822.72 \\
\bottomrule
\end{tabular}
\end{table}

In another experiment, we tried giving the crossover VLM the generated images for each parent prompt in order to better guide the generation of the child prompt. As we show in Tab. \ref{tab:ablation_crossover_grounding}, the effect is minimal, so we use the simpler version that does not ground the generation on parents generated images.


\begin{table}[h]
\centering
\caption{Ablation: effect of providing generated parent images during crossover. In the default setting, the VLM observes only the target image when combining two parent prompts. We compare this against additionally providing the images generated by each parent prompt. Evaluated with CLIP guidance on 100 random MS-COCO 2017 images.}
\label{tab:ablation_crossover_grounding}
\begin{tabular}{lc}
\toprule
\textbf{Crossover Input} & \textbf{CLIP}$\uparrow$ \\
\midrule
Target image only (\ourmethod{})          & \textbf{0.829} \\
Target image + generated parent images    & 0.827 \\
\bottomrule
\end{tabular}
\end{table}

We furthermore explore the effect of spatial emphasis in the VLM prompts. We experimented with both the BLIP and CLIP as our target score function, and show the results in Tab. \ref{tab:ablation_spatial}. Again, we see only very minimal impact, however, we do observe qualitatively that this helps align the generated images better with the target image. It is well attested that CLIP and BLIP struggle with spatial relations \cite{yuksekgonuland}, and we conjecture that there is an improvement that wasn't captured by these scores. \\

\begin{table}[h]
\centering
\caption{Ablation: effect of spatial emphasis in VLM prompts. We compare the default templates against variants that explicitly instruct the VLM to describe object positions and spatial relationships (e.g., ``person stands to the right of the car''). Evaluated on 100 random MS-COCO 2017 images with CLIP and BLIP guidance.}
\label{tab:ablation_spatial}
\adjustbox{max width=\textwidth}{%
\begin{tabular}{lc cccc}
\toprule
\textbf{Guidance} & \textbf{Spatial Emphasis} & \textbf{CLIP}$\uparrow$ & \textbf{BLIP}$\uparrow$ & \textbf{OpenCLIP}$\uparrow$ & \textbf{DreamSim}$\uparrow$ \\
\midrule
\multirow{2}{*}{CLIP}
 & Without & 0.829 & 0.859 & 0.810 & 0.754 \\
 & With    & \textbf{0.830} & \textbf{0.861} & \textbf{0.811} & \textbf{0.756} \\
\midrule
\multirow{2}{*}{BLIP}
 & Without & \textbf{0.790} & 0.885 & \textbf{0.800} & 0.759 \\
 & With    & 0.785 & \textbf{0.887} & 0.798 & \textbf{0.761} \\
\bottomrule
\end{tabular}
}%
\end{table}

In our last experiment, we tried varying the mutation rate in order to promote better exploration. Results in Tab. \ref{tab:ablation_mutation_rate} show a slight improvement with a higher mutation rate.\\

Concluding all these ablation studies, we conclude that \ourmethod{} is very robust and works well across a wide set of prompts and parameters.

\begin{table}[h]
\centering
\caption{Ablation: effect of mutation rate $p_m$. We vary the probability with which offspring undergo mutation after crossover. The default rate used by \ourmethod{} is $0.1$. Evaluated with CLIP guidance on 100 random MS-COCO 2017 images.}
\label{tab:ablation_mutation_rate}
\begin{tabular}{l cccc}
\toprule
\textbf{Mutation Rate} ($p_m$) & \textbf{CLIP}$\uparrow$ & \textbf{BLIP}$\uparrow$ & \textbf{OpenCLIP}$\uparrow$ & \textbf{DreamSim}$\uparrow$ \\
\midrule
0.1 (default) & 0.829 & 0.859 & 0.810 & 0.754 \\
0.3            & 0.831 & 0.861 & 0.810 & 0.754 \\
0.5            & \textbf{0.834} & \textbf{0.861} & \textbf{0.811} & \textbf{0.756} \\
\bottomrule
\end{tabular}
\end{table}


\section{Conclusion}
In this work, we tackle the task of prompt inversion and introduce \ourmethod{}, a simple approach based on a combination of a genetic algorithm with a strong VLM model.  By combining a vision language model to guide a genetic algorithm, our method optimizes natural-language prompts without requiring access to T2I model internals. It is also very versatile and works without any adjustments or fine-tuning on any combination of T2I models and evaluation score functions. We demonstrate its effectiveness across diverse experimental settings and show it is very robust to implementation details. Finally, \ourmethod{} produces coherent, human-readable prompts, improving interpretability and facilitating downstream editing.

\clearpage
\bibliographystyle{splncs04}
\bibliography{main}

\clearpage
\appendix
\renewcommand{\thesection}{\Alph{section}}
\renewcommand{\theHsection}{\Alph{section}}

\section*{Supplementary Material}
\setcounter{section}{0}
\setcounter{figure}{0}
\setcounter{table}{0}
\renewcommand{\thefigure}{\Alph{section}\arabic{figure}}
\renewcommand{\thetable}{\Alph{section}\arabic{table}}

\section{Implementation Details}
\label{sec:implementation_details}
Across the different experiments of \ourmethod{}, we use the following defaults: N (population size) = 10, T (number of generations) = 5, K (number of images per prompt) = 3, $p_m=0.1$ (mutation rate - ratio of mutations to apply per generation). The mutations occur only after a crossover is applied.
For VLM-Baseline experiments, we use N = 60 to make sure we evaluate the same number of prompts as we do in the evolution process, T = 0, K = 3, $p_m=0.0$ (no mutations).
In both settings, we use Qwen3-VL-Instruct as the VLM by default and FLUX.2-klein-4B as our T2I model.
Since the CLIP text encoder is a common core component in many T2I models, we enforce that the different VLM outputs be shorter than 77 tokens; otherwise, the models may perform internal truncation, leading to a loss of detail. We enforce this by mentioning "$\sim$50 words" in the VLM instructions, and by manually truncating to 77 tokens (special and content tokens).

\section{VLM Prompt Templates}
\label{sec:prompts}

We provide the complete VLM prompt templates used by \ourmethod{}.
Section~\ref{sec:structured_prompts} presents the structured templates with step-by-step reasoning scaffolding used in all main experiments.
Section~\ref{sec:grounded_crossover} introduces the prompt used for crossover grounded in parent images (Tab.~\ref{tab:ablation_crossover_grounding}), while Section~\ref{sec:spatial_emphasis} describes the prompts used for VLM inference with spatial emphasis (Tab.~\ref{tab:ablation_spatial}).
Finally, Section~\ref{sec:minimal_prompts} lists the minimal templates without reasoning, used in the ablation study (Tab.~\ref{tab:ablation_template}).
VLM-Baseline experiments used the structured system prompt and the initialization (verbalized sampling) template.
In all templates, text in curly braces
(e.g., \texttt{\{n\}}, \texttt{\{prompt\_1\}})
denotes placeholders filled at runtime.

\subsection{Structured Templates (Default)}
\label{sec:structured_prompts}

\paragraph{System Prompt.}
Used as the system message for all VLM calls.

{\footnotesize
\begin{verbatim}
You are an expert at generating and refining prompts for
text-to-image diffusion models. You will analyze images and
evolve prompts to more accurately describe them.
Always output your final prompt wrapped in
<prompt></prompt> tags.
Be specific, vivid, and accurate to what you observe
in the image.
\end{verbatim}
}

\paragraph{Initialization (Verbalized Sampling).}
Generates the initial population of $N$ diverse prompts
from a single VLM call.

{\footnotesize
\begin{verbatim}
Generate {n} diverse text-to-image prompts that accurately
describe this image.

For EACH prompt, follow these analysis steps:
1. Identify the main subject(s) - what is the primary focus?
2. Describe the setting/environment - where is this taking
   place?
3. Note the composition - how are elements arranged? What
   perspective?
4. Capture the lighting - what is the light source, quality,
   mood?
5. Identify the style - is it photorealistic, artistic,
   rendered, etc.?
6. Add important details - colors, textures, atmosphere, any
   text visible

Combine these observations into a cohesive, detailed prompt.
Each prompt should be approximately 50 words - detailed and
information-dense. Include all important visual elements -
do not omit key subjects, objects, or setting details.

Diversity requirements:
- Each prompt should emphasize DIFFERENT aspects or
  perspectives of the image
- Vary which details you prioritize (e.g., one focuses on
  subjects, another on atmosphere)
- Use different descriptive styles while maintaining accuracy
- Ensure genuine diversity, not just word substitutions

For each prompt, provide a probability (0-1) indicating how
well it captures the image.
The probabilities should sum to approximately 1.0.

Output format (output ONLY the prompts, no other text):
<prompt probability="0.XX">detailed prompt text here</prompt>
<prompt probability="0.XX">detailed prompt text here</prompt>
...
\end{verbatim}
}

\paragraph{Crossover.}
Combines two parent prompts into a child, guided by the target image.

{\footnotesize
\begin{verbatim}
Please follow these steps to generate a better prompt by
combining two existing prompts.

Step 1 - Identify SHARED elements between the prompts (these
are likely accurate, keep them):
Prompt 1: {prompt_1}
Prompt 2: {prompt_2}

Step 2 - Identify DIFFERENT elements between the two prompts.

Step 3 - Looking carefully at the image, determine which of
the different elements more accurately describes what you see.

Step 4 - Combine the shared elements with the most accurate
different elements into a single coherent prompt.

Step 5 - Ensure the final prompt flows naturally and maintains
consistency.

Write a detailed, information-dense prompt of approximately
50 words. Include all important visual elements.
Output only the final combined prompt wrapped in
<prompt></prompt> tags.
\end{verbatim}
}

\paragraph{Mutation.}
Applies targeted edits to a child prompt, guided by the target image.

{\footnotesize
\begin{verbatim}
Please follow these steps to improve this prompt by
introducing beneficial variations.

Current prompt to mutate: {prompt}

Step 1 - Evaluate accuracy: Compare each claim in the prompt
against the image. Note any inaccuracies.

Step 2 - Identify gaps: What important visual elements are
missing from the prompt?

Step 3 - Assess specificity: Which terms are too vague and
could be more precise?

Step 4 - Consider alternatives: For key descriptive words,
think of synonyms or more evocative alternatives that might
better capture the image.

Step 5 - Apply mutations: Make 1-3 targeted changes that
improve accuracy, add missing details, or enhance
descriptions. Do not change elements that are already
accurate.

Write a detailed, information-dense prompt of approximately
50 words. Include all important visual elements.
Output only the mutated prompt wrapped in
<prompt></prompt> tags.
\end{verbatim}
}

\subsection{Crossover with Parent Images (Ablation)}
\label{sec:grounded_crossover}

\paragraph{Crossover.}
Combines two parent prompts into a child, guided by the target image and two selected images generated by the parent prompts.

{\footnotesize
\begin{verbatim}
You are given three images and two prompts.
Your task is to combine the prompts into one
that better reconstructs the reference image.
Images (in order):
- Image 1: REFERENCE — the target image we want to reconstruct
- Image 2: GENERATED from Prompt 1
- Image 3: GENERATED from Prompt 2
Prompt 1: {prompt_1}
Prompt 2: {prompt_2}
Follow these steps:
Step 1 - Visual comparison: Compare Image 2
(generated from Prompt 1) to Image 1 (reference).
What does the generation get RIGHT — composition,
colors, subjects, style, lighting, details?
What does it get WRONG or MISS?
Step 2 - Visual comparison: Do the same
for Image 3 (generated from Prompt 2) vs Image 1 (reference).
Step 3 - Attribute to prompt language: For each
visual match or mismatch you identified, trace it back
to specific words or phrases in the corresponding prompt.
Which prompt phrases led to accurate reconstruction?
Which led to errors?
Step 4 - Combine: Build a new prompt that:
- KEEPS phrases from either prompt that produced
accurate visual elements
- REPLACES phrases that led to visual errors
with the better alternative from the other prompt
- Uses ONLY language from the two existing prompts — do
not introduce new descriptions
Step 5 - Ensure the final prompt is coherent and natural.
Write a detailed, information-dense prompt of approximately
50 words. Include all important visual elements.
Output only the final combined prompt
wrapped in <prompt></prompt> tags.
\end{verbatim}
}

\subsection{VLM with Spatial Emphasis (Ablation)}
\label{sec:spatial_emphasis}

\paragraph{System Prompt.}
Used as the system message for all VLM calls.

{\footnotesize
\begin{verbatim}
You are an expert at generating and refining prompts for
text-to-image diffusion models. You will analyze images and
evolve prompts to more accurately describe them.
Always output your final prompt wrapped in
<prompt></prompt> tags.
Be specific, vivid, and accurate to what you observe
in the image.
Pay attention to the spatial positioning of objects
and subjects — describe where things are located using spatial
terms (left/right/center, top/bottom, foreground/background).
\end{verbatim}
}

\paragraph{Initialization (Verbalized Sampling).}
Generates the initial population of $N$ diverse prompts
from a single VLM call.

{\footnotesize
\begin{verbatim}
Generate {n} diverse text-to-image prompts that accurately
describe this image.

For EACH prompt, follow these analysis steps:
1. Identify the main subject(s) - what is the primary focus?
2. Describe the setting/environment - where is this taking
   place?
3. Note the composition - how are elements arranged? What
   perspective?
4. Describe the spatial layout - where are subjects
   positioned (left/right/center, top/bottom, foreground/background)?
   Include relative positions of objects to each other.
5. Capture the lighting - what is the light source, quality,
   mood?
6. Identify the style - is it photorealistic, artistic,
   rendered, etc.?
7. Add important details - colors, textures, atmosphere, any
   text visible

Combine these observations into a cohesive, detailed prompt.
Integrate spatial positioning naturally.
Each prompt should be approximately 60 words - detailed and
information-dense. Include all important visual elements -
do not omit key subjects, objects, or setting details.

Diversity requirements:
- Each prompt should emphasize DIFFERENT aspects or
  perspectives of the image
- Vary which details you prioritize (e.g., one focuses on
  subjects, another on atmosphere)
- Use different descriptive styles while maintaining accuracy
- Ensure genuine diversity, not just word substitutions

For each prompt, provide a probability (0-1) indicating how
well it captures the image.
The probabilities should sum to approximately 1.0.

Output format (output ONLY the prompts, no other text):
<prompt probability="0.XX">detailed prompt text here</prompt>
<prompt probability="0.XX">detailed prompt text here</prompt>
...
\end{verbatim}
}

\paragraph{Crossover.}
Combines two parent prompts into a child, guided by the target image.

{\footnotesize
\begin{verbatim}
Please follow these steps to generate a better prompt by
combining two existing prompts.

Step 1 - Identify SHARED elements between the prompts (these
are likely accurate, keep them):
Prompt 1: {prompt_1}
Prompt 2: {prompt_2}

Step 2 - Identify DIFFERENT elements between the two prompts.

Step 3 - Looking carefully at the image, determine which of
the different elements more accurately describes what you see.
Pay special attention to spatial descriptions — if one prompt
places a subject "on the left" and another says "in the center",
check the image to determine the correct position.

Step 4 - Combine the shared elements with the most accurate
different elements into a single coherent prompt.
Preserve accurate spatial positioning.

Step 5 - Ensure the final prompt flows naturally and maintains
consistency.

Write a detailed, information-dense prompt of approximately
60 words. Include all important visual elements.
Output only the final combined prompt wrapped in
<prompt></prompt> tags.
\end{verbatim}
}

\paragraph{Mutation.}
Applies targeted edits to a child prompt, guided by the target image.

{\footnotesize
\begin{verbatim}
Please follow these steps to improve this prompt by
introducing beneficial variations.

Current prompt to mutate: {prompt}

Step 1 - Evaluate accuracy: Compare each claim in the prompt
against the image. Note any inaccuracies.

Step 2 - Identify gaps: What important visual elements are
missing from the prompt?

Step 3 - Check spatial accuracy: Are objects described
in their correct positions (left/right/center, top/bottom,
foreground/background)? Fix any spatial errors and
add positioning details where missing.

Step 4 - Assess specificity: Which terms are too vague and
could be more precise?

Step 5 - Consider alternatives: For key descriptive words,
think of synonyms or more evocative alternatives that might
better capture the image.

Step 6 - Apply mutations: Make 1-3 targeted changes that
improve accuracy, add missing details, or enhance
descriptions. Do not change elements that are already
accurate.
Integrate spatial information naturally.

Write a detailed, information-dense prompt of approximately
60 words. Include all important visual elements.
Output only the mutated prompt wrapped in
<prompt></prompt> tags.
\end{verbatim}
}

\subsection{Minimal Templates (Ablation)}
\label{sec:minimal_prompts}

These task-only templates omit the step-by-step reasoning scaffolding
and were used in the template ablation study.

\paragraph{System Prompt.}

{\footnotesize
\begin{verbatim}
Generate text-to-image prompts based on images.
Output each prompt in <prompt></prompt> tags.
\end{verbatim}
}

\paragraph{Initialization (Verbalized Sampling).}

{\footnotesize
\begin{verbatim}
Generate {n} diverse text-to-image prompts that describe
this image.
Each prompt should be approximately 50 words.

For each prompt, provide a probability (0-1) reflecting how
well it captures the image. Probabilities should sum to
approximately 1.0.

Output format:
<prompt probability="0.XX">prompt text</prompt>
\end{verbatim}
}

\paragraph{Crossover.}

{\footnotesize
\begin{verbatim}
Combine these two prompts into a single better prompt that
more accurately describes the image.

Prompt 1: {prompt_1}
Prompt 2: {prompt_2}

Write approximately 50 words.
Output only the result in <prompt></prompt> tags.
\end{verbatim}
}

\paragraph{Mutation.}

{\footnotesize
\begin{verbatim}
Improve this prompt so it more accurately describes the image.

Current prompt: {prompt}

Write approximately 50 words.
Output only the improved prompt in <prompt></prompt> tags.
\end{verbatim}
}

\section{Prompt Evolution Examples}
\label{sec:prompt_examples}

To demonstrate the effectiveness and necessity of the genetic algorithm in \ourmethod{}, we present representative examples of the prompt evolution process. For each example, we show the reference image alongside two reconstructions: one generated from the initial VLM prompt (before evolution) and one from the final evolved prompt (after the full evolutionary process). This side-by-side comparison highlights how the genetic algorithm refines prompts to capture fine-grained visual details that the initial VLM description misses, leading to substantially more faithful reconstructions. See Fig.~\ref{fig:example_242}--\ref{fig:example_428}.

\promptexample
{figs/main/1/original/image_242_original.png}
{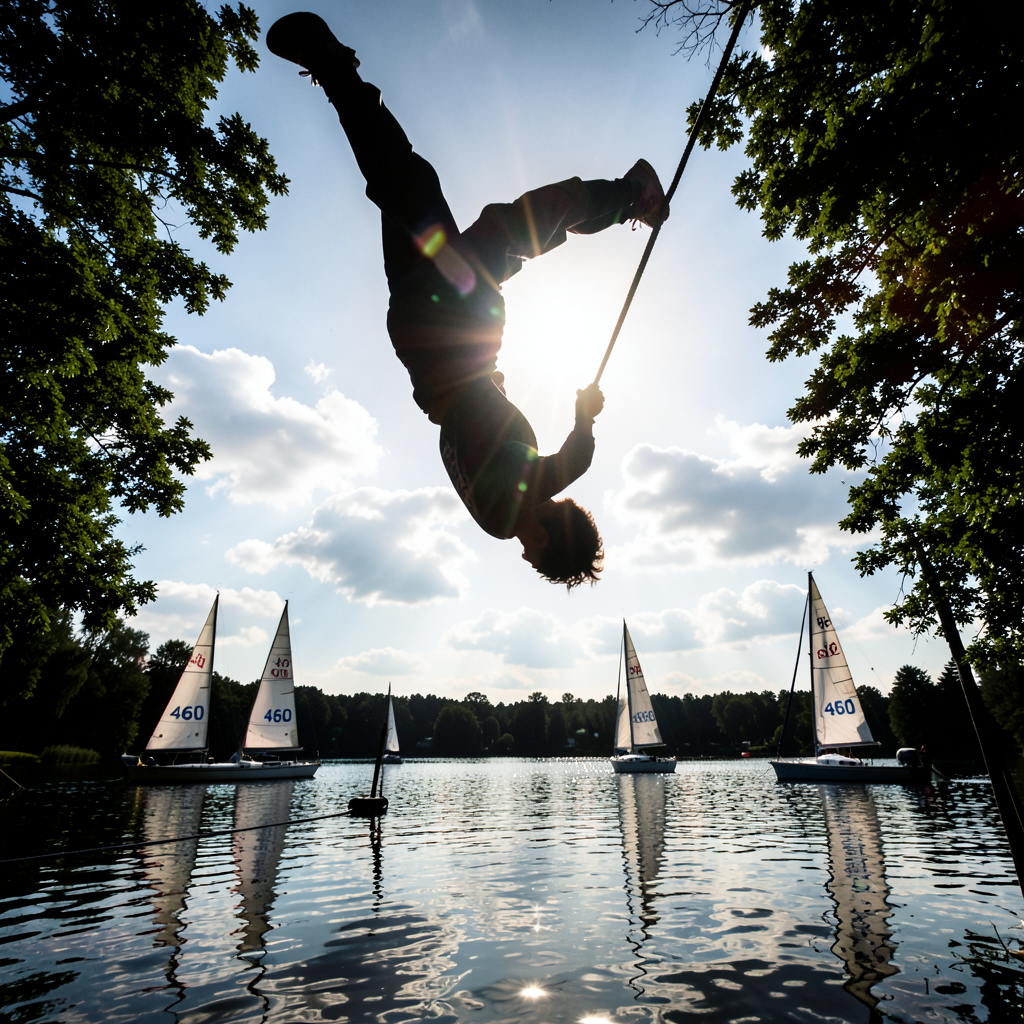}
{figs/main/1/ours/image_242_evolved.png}
{Dramatic low-angle shot capturing a person mid-air during a slackline backflip, their silhouette sharply defined against a sunlit sky. Trees on either side frame the action, while sailboats with ``460'' text glide on a reflective lake. Sun flares and cloud textures add intensity to the adventurous moment.}
{Dramatic low-angle shot of a slackliner mid-backflip, silhouetted against a sun-drenched sky with piercing lens flares and scattered clouds. Framed by dense tree branches, sailboats marked ``460'' with red star decals glide on a reflective lake, capturing adventurous serenity at golden hour.}
{Prompt evolution example (Flickr8K, DreamSim guidance). The evolved prompt adds specific details such as ``red star decals'' and ``golden hour'' that improve reconstruction fidelity.}
{fig:example_242}

\promptexample
{figs/main/1/original/image_402_original.png}
{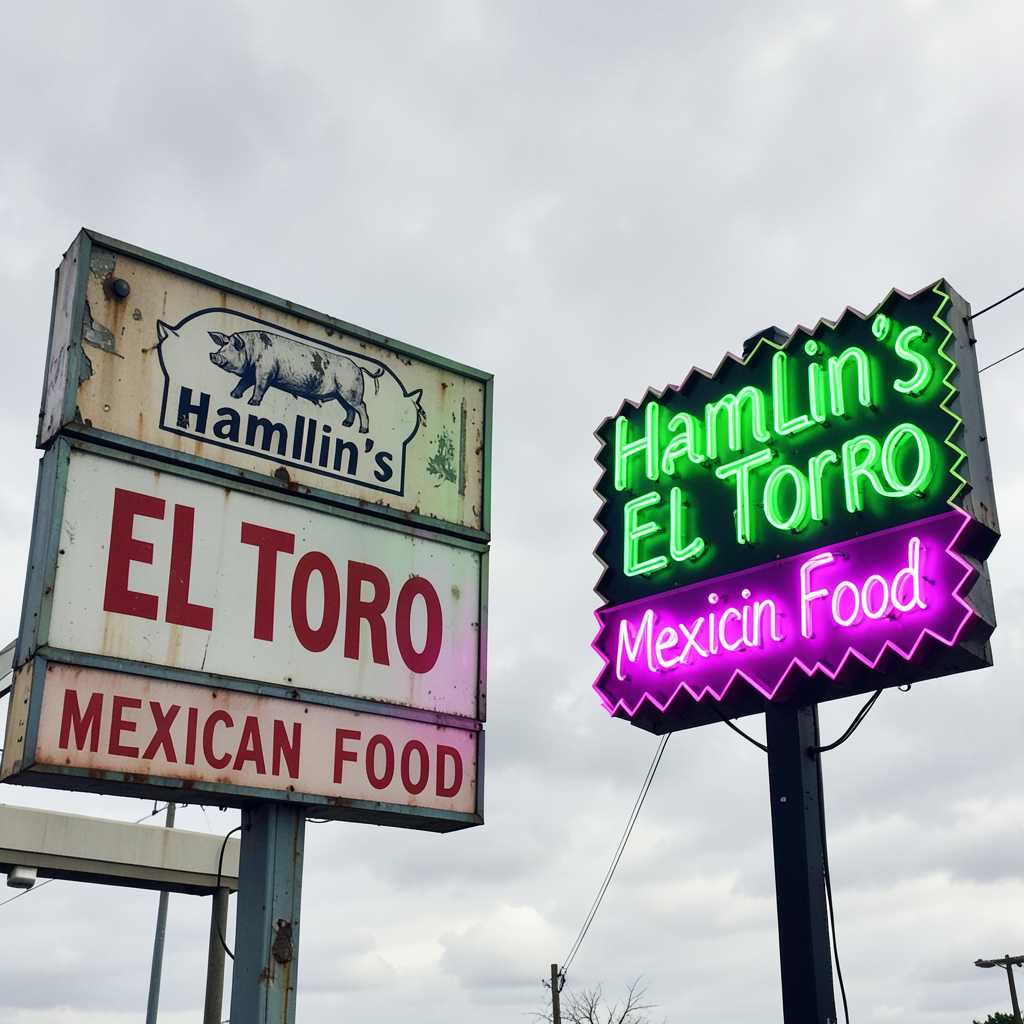}
{figs/main/1/ours/image_402_evolved.png}
{A weathered, multi-layered sign on the left displays ``Hamlin's'' above a pig logo, followed by ``El Toro'' and ``Mexican Food'' in bold red. To its right, a bright neon sign glows with ``Hamlin's El Toro Mexican Food'' in green and magenta, framed by zigzag borders. Both signs stand against a cloudy sky, evoking.}
{Photorealistic split-image under a cloudy, diffused sky: Left, rustic ``Hamlin's'' sign with pig silhouette, ``El Toro'' below, and ``Mexican Food'' plaque; right, glowing neon red sign with green ``El Toro,'' yellow ``Mexican Food,'' and purple zigzag border - both evoke nostalgic roadside vintage charm.}
{Prompt evolution example (MS-COCO, CLIP guidance). The initial prompt was truncated by the CLIP truncation mechanism detailed in \ref{sec:implementation_details}. Evolution produces a complete, structured description that better captures both signs.}
{fig:example_402}

\promptexample
{figs/main/1/original/image_408_original.png}
{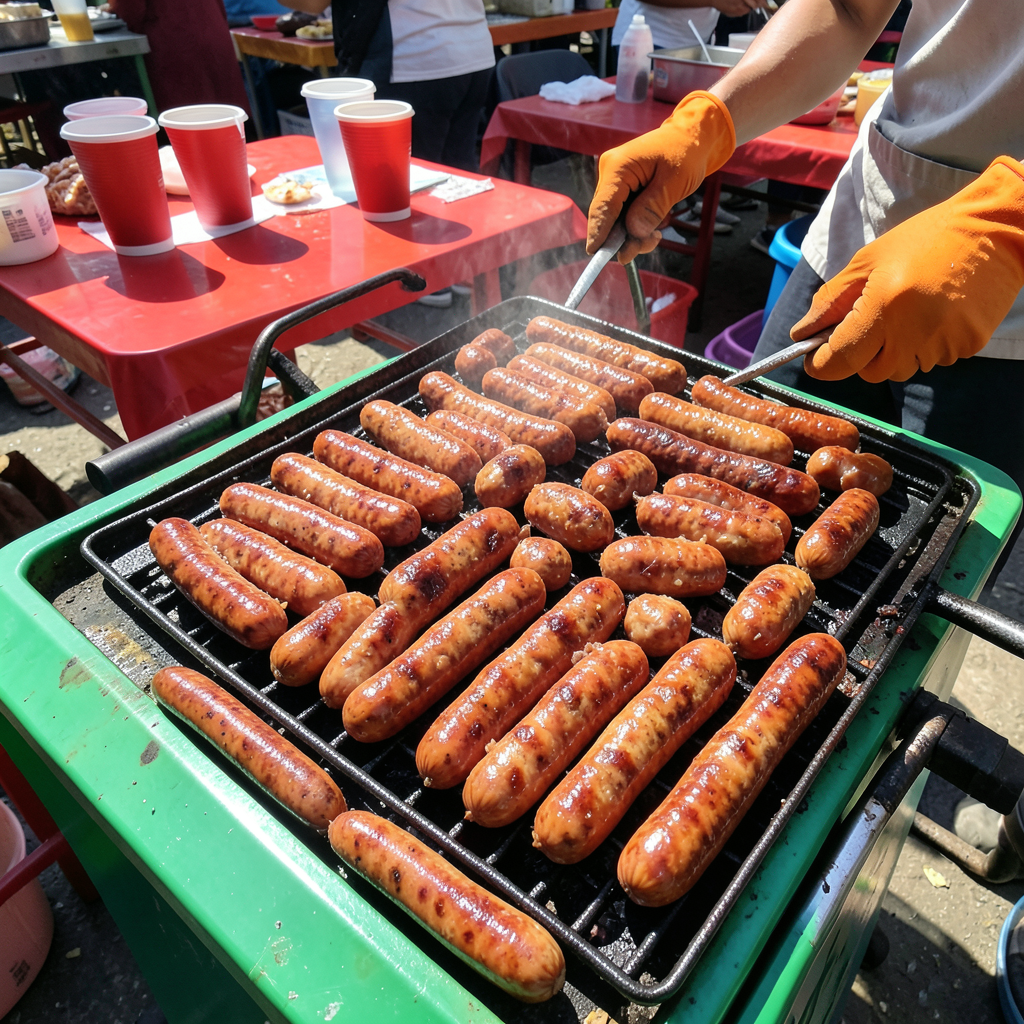}
{figs/main/1/ours/image_408_evolved.png}
{Vivid, photorealistic depiction of a food vendor's grill in an outdoor market. Rows of plump, reddish sausages sizzle under direct sunlight, their surfaces glistening with fat. The green grill contrasts with the red tables and plastic cups in the background, while a person's arm in orange gloves adds motion and context to the scene.}
{A low-angle, slightly elevated close-up of dozens of glistening, reddish sausages sizzling in neat rows on a weathered green grill under bright sunlight. Sharp shadows accentuate their curves. Red plastic tables and a gloved hand frame the bustling street food scene, photorealistic and hyper-detailed.}
{Prompt evolution example (MS-COCO, SigLIP guidance). The evolved prompt corrects the camera angle to ``low-angle, slightly elevated'' and adds texture details like ``weathered green grill'' and ``sharp shadows.''}
{fig:example_408}

\promptexample
{figs/main/1/original/image_428_original.png}
{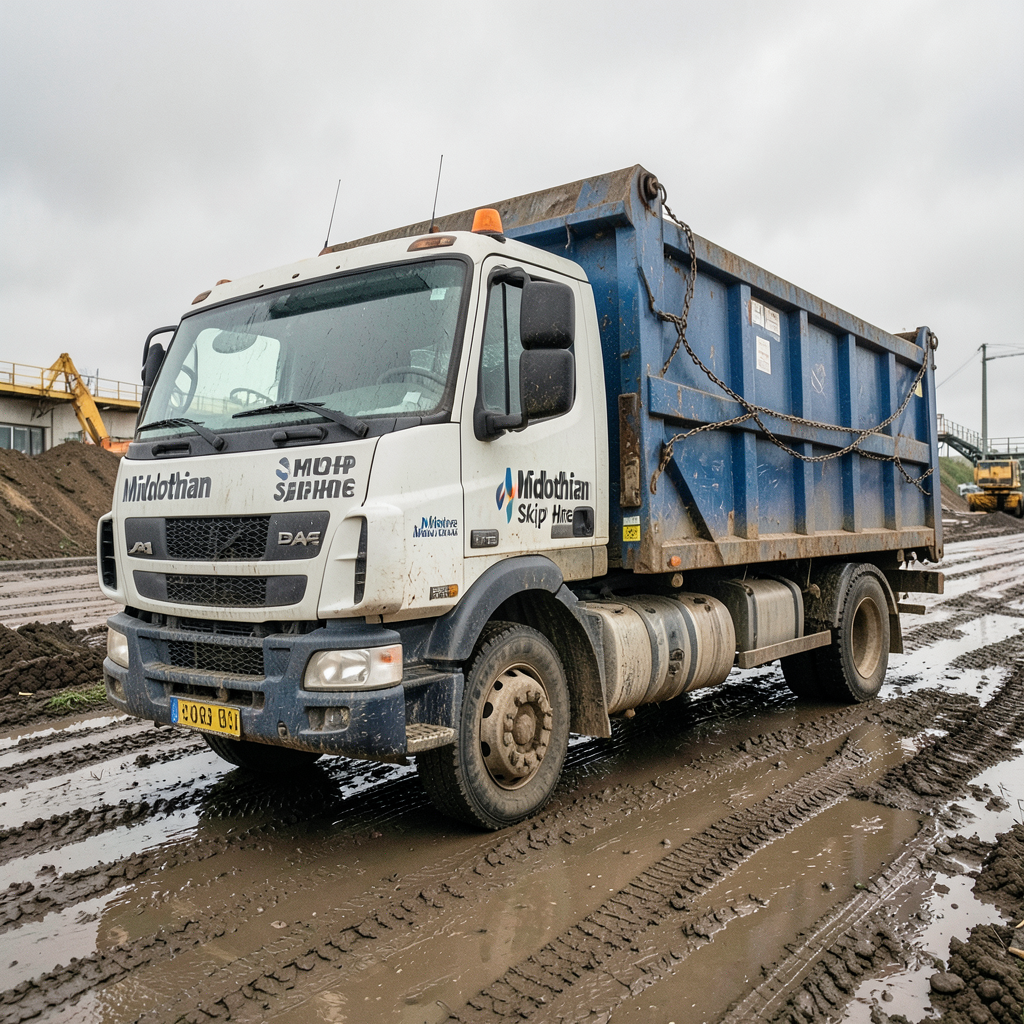}
{figs/main/1/ours/image_428_evolved.png}
{Texture and environment focus: A gritty, realistic depiction of a skip lorry on a muddy construction site. The white DAF truck, branded ``Midlothian Skip Hire,'' features a blue skip secured by chains. Overcast skies cast flat light, emphasizing worn tires, grime, and the industrial atmosphere of the setting.}
{A weathered white DAF skip hire truck, branded ``Midlothian Skip Hire LTD,'' dominates a muddy construction site under overcast skies. Its blue skip, secured with chains, shows wear. Green recycling logos, contact numbers, and license plate ``EX52 XRT'' are visible. Low-angle shot emphasizes rugged industrial grit and commercial identity.}
{Prompt evolution example (MS-COCO, SigLIP guidance). Evolution recovers fine-grained text details (``LTD,'' license plate ``EX52 XRT,'' ``green recycling logos'') absent from the initial prompt.}
{fig:example_428}

\section{Failure Cases}
\label{sec:failure_cases}

While \ourmethod{} achieves strong results overall, we observe characteristic failure modes that highlight limitations of the prompt inversion paradigm. An example is shown in Fig.~\ref{fig:failure_attrib}, illustrating cases where VLM-driven reasoning captures plausible visual attributes yet fails to reflect less common but semantically important details.

\failurecase
{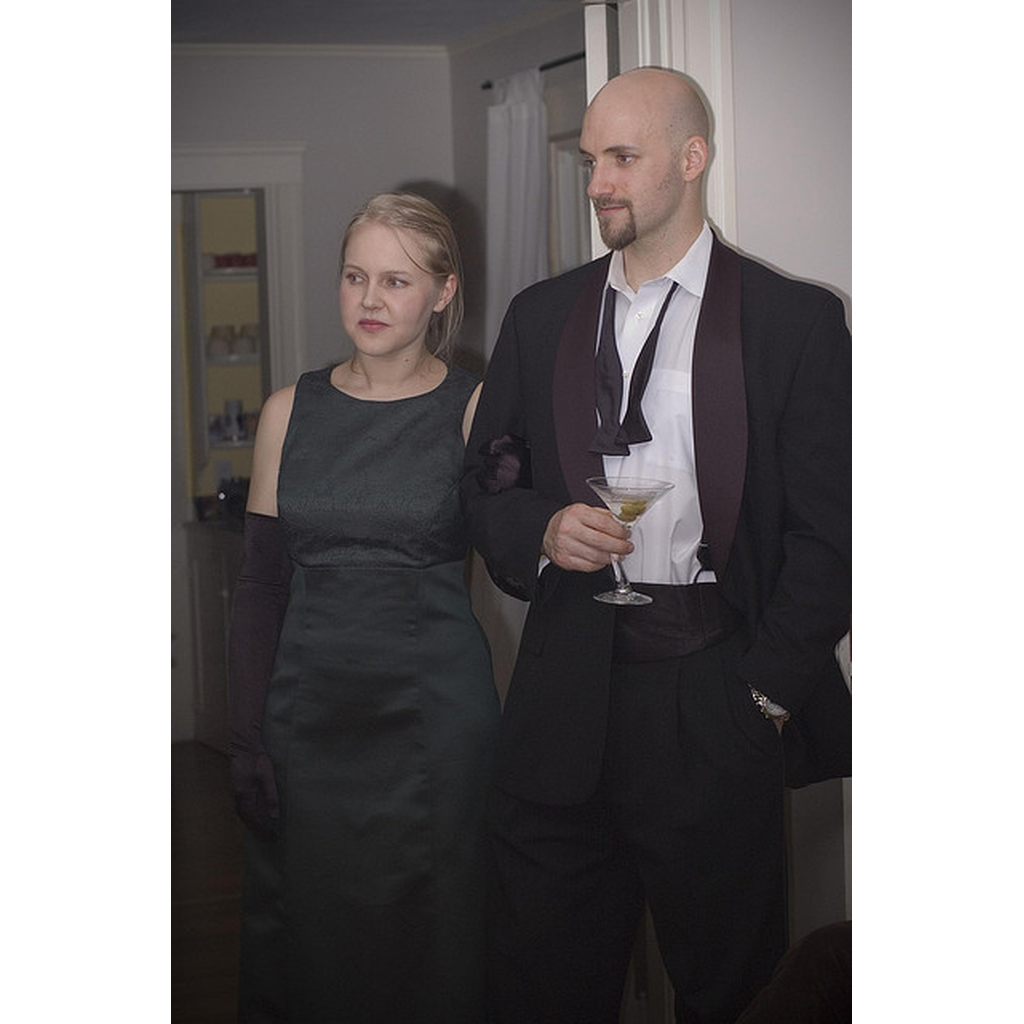}
{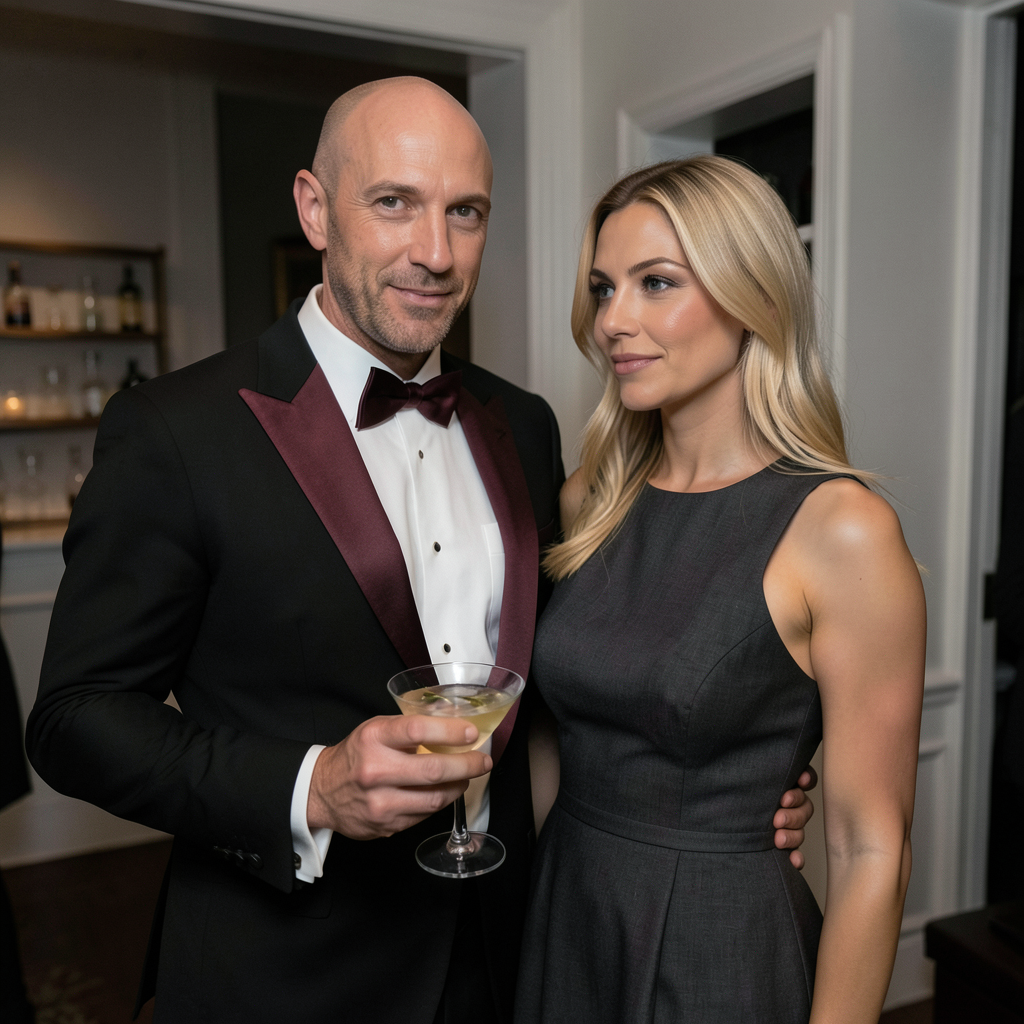}
{Photorealistic, cinematic close-up of a stylish couple at an elegant indoor soir\'{e}e: bald man in black tuxedo with burgundy satin lapels, holding a martini, beside a blonde woman in a sleeveless charcoal gown. Soft, dim lighting highlights formal attire. Background: white walls, doorway, shelves. Subdued elegance.}
{Failure case: The T2I model inferred the bowtie correctly without specific mention in the prompt (presumably because of the tuxedo), however, the fact that it is open was not mentioned in any of the prompts throughout the evolution process. We hypothesize that this is due to the fact that the unnatural state of the bowtie receives a lot of human attention, however modern foundational models have a bias towards the more common state.}
{fig:failure_attrib}

\section{Human Preference Study Details}
\label{sec:human_eval_details}

This section provides the full experimental protocol for the human preference study summarized in Sec.~\ref{sec:human_eval}.

\paragraph{Objective.}
We conducted a two-alternative forced choice (2AFC) study to compare the reconstruction quality of \ourmethod{} against the VLM-Baseline. For each comparison, an evaluator was shown a reference image alongside two reconstructions (one from each method) and asked to select the one that more faithfully matches the reference.

\paragraph{Stimuli.}
We used all 500 images from our evaluation set (100 per dataset: Lexica, MS-COCO, CelebA, Flickr8K, and LAION-400M). For each image, the \ourmethod{} reconstruction was drawn from one of the four guidance variants (CLIP, BLIP, OpenCLIP-SigLIP, DreamSim), assigned via balanced allocation (125 images per variant, seed 42). For both methods, one of the $K{=}3$ generated images was randomly selected per image.

\paragraph{Platform.}
The study was implemented as a custom web application using Google Sheets and Google Apps Script. Each evaluator accessed the study via a web link and was identified by their Google account email. The application implemented balanced item assignment: each evaluator was shown 10 comparisons, selected from the items with the fewest existing annotations to ensure even coverage across all 500 image pairs.

\begin{figure}[H]
\centering
\includegraphics[width=0.65\textwidth]{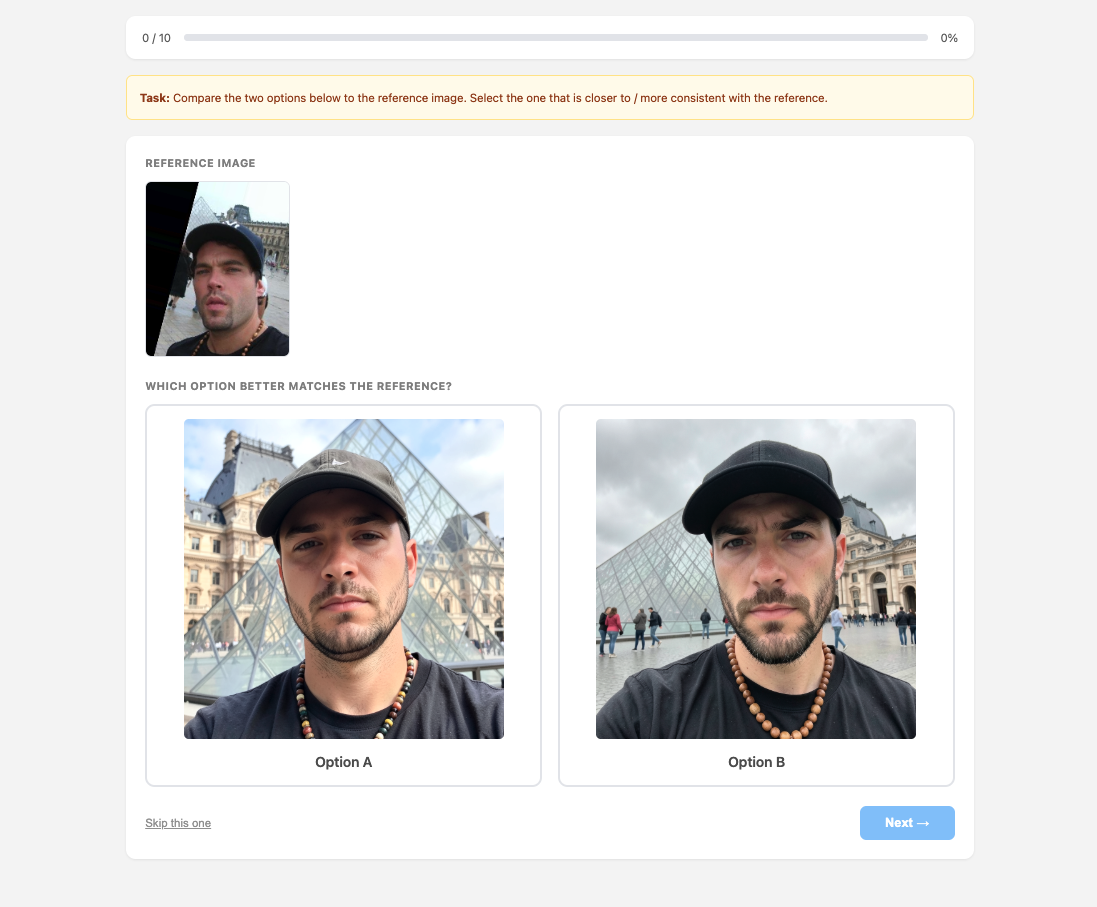}
\caption{Screenshot of the human evaluation interface for a single comparison. The reference image is shown at the top, with two reconstruction options displayed below. Evaluators select the option that more faithfully matches the reference.}
\label{fig:human_eval_ui}
\end{figure}

\paragraph{Interface.}
The evaluation interface displayed the reference image at the top, followed by two reconstructions labeled ``Option A'' and ``Option B'' placed side by side below (see Fig.~\ref{fig:human_eval_ui}). The assignment of \ourmethod{} and VLM-Baseline to left/right positions was randomized independently for each comparison. The presentation order of the 10 assigned items was also shuffled per evaluator. Both performed in order to ensure fairness and to ensure no bias is possible. The instruction read: \emph{``Task: Compare the two options below to the reference image. Select the one that is closer to / more consistent with the reference.''}

\paragraph{Coverage.}
We recruited 94 unique evaluators. Participants were volunteers and were not compensated.
In total, we collected 956 pairwise annotations across 500 image pairs, yielding an average of 1.91 annotations per pair.

\paragraph{Results.}
Of the 956 comparisons, 521 (54.5\%) favored \ourmethod{} and 435 (45.5\%) favored the VLM-Baseline. A one-sided binomial test yields $p = 0.003$, confirming a statistically significant preference for our method. Tab.~\ref{tab:human_eval_per_dataset} reports the per-dataset breakdown.

\begin{table}[H]
\centering
\caption{Per-dataset breakdown of the human preference study. We report the number of comparisons favoring \ourmethod{} out of the total for each dataset.}
\label{tab:human_eval_per_dataset}
\begin{tabular}{lcccc}
\toprule
\textbf{Dataset} & \textbf{Ours} & \textbf{VLM-Baseline} & \textbf{Total} & \textbf{Pref.\ (\%)} \\
\midrule
Lexica     & 124 &  73 & 197 & 62.9 \\
CelebA     &  84 & 108 & 192 & 43.8 \\
Flickr8K   & 111 &  85 & 196 & 56.6 \\
LAION-400M & 106 &  91 & 197 & 53.8 \\
MS-COCO    &  96 &  78 & 174 & 55.2 \\
\midrule
\textbf{Total} & \textbf{521} & \textbf{435} & \textbf{956} & \textbf{54.5} \\
\bottomrule
\end{tabular}
\end{table}

Notably, \ourmethod{} is preferred on four out of five datasets. The exception is CelebA, where the VLM-Baseline is preferred (43.8\% vs.\ 56.2\%). We hypothesize that portrait images, which feature a single dominant subject with relatively uniform composition, are already well-described by a single VLM pass, leaving less room for evolutionary refinement.


\section{Extended Quantitative Results}
\label{sec:extended_results}

In this section we provide additional experiments that complement the main paper due to space constraints.

\subsection{Extended Evaluation with OpenCLIP-SigLIP}
\label{sec:openclip_eval}

Tab.~\ref{tab:main_results} reports CLIP, BLIP, and DreamSim evaluation metrics. Here we extend the comparison by adding OpenCLIP-SigLIP~\cite{ilharco_gabriel_2021_5143773} as a fourth evaluation metric. Tab.~\ref{tab:extended_openclip} reproduces the full table with the additional OpenCLIP column for each dataset. The results confirm that \ourmethod{} consistently outperforms all baselines on this metric as well, with the OpenCLIP-guided variant achieving the highest scores. Notably, the strong correlation across all four metrics supports the claim that our method finds prompts that generalize well across different similarity measures, rather than overfitting to a single score function.

\begin{table}[H]
\centering
\caption{Extended quantitative comparison including OpenCLIP-SigLIP evaluation. We report CLIP, BLIP, DreamSim, and OpenCLIP image-to-image similarities between the original and reconstructed images across all five datasets.}
\label{tab:extended_openclip}
\adjustbox{max width=\textwidth}{%
\begin{tabular}{l cccc cccc cccc cccc cccc}
\toprule
& \multicolumn{4}{c}{\textbf{Lexica}} & \multicolumn{4}{c}{\textbf{COCO}} & \multicolumn{4}{c}{\textbf{CelebA}} & \multicolumn{4}{c}{\textbf{Flickr8K}} & \multicolumn{4}{c}{\textbf{LAION-400M}} \\
\cmidrule(lr){2-5} \cmidrule(lr){6-9} \cmidrule(lr){10-13} \cmidrule(lr){14-17} \cmidrule(lr){18-21}
 & CLIP & BLIP & DSim & OCL & CLIP & BLIP & DSim & OCL & CLIP & BLIP & DSim & OCL & CLIP & BLIP & DSim & OCL & CLIP & BLIP & DSim & OCL \\
\midrule
PEZ                   & 0.75 & 0.77 & 0.70 & 0.75 & 0.73 & 0.67 & 0.66 & 0.71 & 0.64 & 0.66 & 0.68 & 0.67 & 0.74 & 0.69 & 0.66 & 0.73 & 0.63 & 0.63 & 0.63 & 0.61 \\
VGD                   & 0.77 & 0.78 & 0.70 & 0.76 & 0.73 & 0.68 & 0.66 & 0.70 & 0.65 & 0.68 & 0.68 & 0.67 & 0.75 & 0.72 & 0.67 & 0.72 & 0.61 & 0.60 & 0.62 & 0.59 \\
CLIP Interrogator     & 0.79 & 0.80 & 0.72 & 0.77 & 0.74 & 0.72 & 0.67 & 0.72 & 0.68 & 0.72 & 0.71 & 0.70 & 0.78 & 0.74 & 0.67 & 0.76 & 0.66 & 0.68 & 0.65 & 0.62 \\
STEPS                 & 0.75 & 0.78 & 0.71 & 0.75 & 0.76 & 0.70 & 0.68 & 0.73 & 0.65 & 0.68 & 0.69 & 0.69 & 0.77 & 0.72 & 0.68 & 0.75 & 0.65 & 0.66 & 0.65 & 0.63 \\
\midrule
VLM-Baseline          & 0.77 & 0.85 & 0.78 & 0.78 & 0.79 & 0.85 & 0.76 & 0.79 & 0.64 & 0.79 & 0.76 & 0.72 & 0.77 & 0.85 & 0.76 & 0.78 & 0.79 & 0.84 & 0.78 & 0.78 \\
\midrule
\ourmethod{}-CLIP     & \textbf{0.82} & 0.86 & 0.77 & 0.80 & \textbf{0.83} & 0.86 & 0.75 & 0.81 & \textbf{0.69} & 0.80 & 0.75 & 0.73 & \textbf{0.82} & 0.85 & 0.75 & 0.79 & \textbf{0.84} & 0.85 & 0.77 & 0.80 \\
\ourmethod{}-BLIP     & 0.78 & \textbf{0.89} & 0.77 & 0.79 & 0.79 & \textbf{0.89} & 0.76 & 0.80 & 0.64 & \textbf{0.83} & 0.75 & 0.73 & 0.78 & \textbf{0.88} & 0.76 & 0.79 & 0.80 & \textbf{0.87} & 0.78 & 0.79 \\
\ourmethod{}-DreamSim & 0.78 & 0.87 & \textbf{0.79} & 0.80 & 0.79 & 0.87 & \textbf{0.78} & 0.80 & 0.65 & 0.81 & \textbf{0.77} & 0.73 & 0.78 & 0.86 & \textbf{0.77} & 0.79 & 0.80 & 0.85 & \textbf{0.80} & 0.79 \\
\ourmethod{}-OpenCLIP & 0.79 & 0.86 & 0.77 & \textbf{0.82} & 0.80 & 0.86 & 0.75 & \textbf{0.83} & 0.65 & 0.80 & 0.75 & \textbf{0.76} & 0.79 & 0.85 & 0.75 & \textbf{0.82} & 0.81 & 0.85 & 0.77 & \textbf{0.82} \\
\bottomrule
\end{tabular}
}%
\end{table}

\subsection{Alternative T2I Model: SDXL-Turbo}
\label{sec:sdxl_turbo}

To demonstrate that \ourmethod{} generalizes beyond the FLUX-2-Klein model used in the main experiments, we repeat the full evaluation using SDXL-Turbo~\cite{sauer2023adversarial} as the T2I backbone. SDXL-Turbo is a distilled variant of Stable Diffusion XL that generates images in a single forward pass (1--4 steps), offering a substantially different architecture and generation paradigm compared to FLUX. All other settings (VLM, datasets, population size, generations) remain identical to the main experiments.

The results are presented in Tab.~\ref{tab:sdxl_turbo}. Despite the architectural differences, \ourmethod{} consistently outperforms the VLM-Baseline across all datasets and metrics, confirming that our evolutionary approach is T2I-model agnostic.

\begin{table}[H]
\centering
\caption{Quantitative comparison using SDXL-Turbo as the T2I model. We report CLIP, BLIP, DreamSim, and OpenCLIP image-to-image similarities.}
\label{tab:sdxl_turbo}
\adjustbox{max width=\textwidth}{%
\begin{tabular}{l cccc cccc cccc cccc cccc}
\toprule
& \multicolumn{4}{c}{\textbf{Lexica}} & \multicolumn{4}{c}{\textbf{COCO}} & \multicolumn{4}{c}{\textbf{CelebA}} & \multicolumn{4}{c}{\textbf{Flickr8K}} & \multicolumn{4}{c}{\textbf{LAION-400M}} \\
\cmidrule(lr){2-5} \cmidrule(lr){6-9} \cmidrule(lr){10-13} \cmidrule(lr){14-17} \cmidrule(lr){18-21}
 & CLIP & BLIP & DSim & OCL & CLIP & BLIP & DSim & OCL & CLIP & BLIP & DSim & OCL & CLIP & BLIP & DSim & OCL & CLIP & BLIP & DSim & OCL \\
\midrule
VLM-Baseline          & 0.81 & 0.84 & 0.75 & 0.80 & 0.79 & 0.79 & 0.71 & 0.76 & 0.68 & 0.77 & 0.73 & 0.70 & 0.80 & 0.80 & 0.71 & 0.76 & 0.75 & 0.77 & 0.71 & 0.70 \\
\midrule
\ourmethod{}-CLIP     & \textbf{0.83} & 0.84 & 0.76 & 0.81 & \textbf{0.81} & 0.81 & 0.72 & 0.77 & \textbf{0.70} & 0.77 & 0.73 & 0.70 & \textbf{0.82} & 0.81 & 0.71 & 0.78 & \textbf{0.76} & 0.77 & 0.71 & 0.70 \\
\ourmethod{}-BLIP     & 0.80 & \textbf{0.87} & 0.76 & 0.81 & 0.77 & \textbf{0.83} & 0.72 & 0.76 & 0.66 & \textbf{0.81} & 0.74 & 0.71 & 0.78 & \textbf{0.84} & 0.72 & 0.76 & 0.73 & \textbf{0.81} & 0.72 & 0.70 \\
\ourmethod{}-DreamSim & 0.80 & 0.85 & \textbf{0.78} & 0.81 & 0.77 & 0.82 & \textbf{0.74} & 0.76 & 0.67 & 0.78 & \textbf{0.75} & 0.71 & 0.78 & 0.81 & \textbf{0.73} & 0.76 & 0.73 & 0.78 & \textbf{0.73} & 0.71 \\
\ourmethod{}-OpenCLIP & 0.80 & 0.85 & 0.76 & \textbf{0.83} & 0.78 & 0.81 & 0.72 & \textbf{0.79} & 0.67 & 0.78 & 0.74 & \textbf{0.74} & 0.79 & 0.80 & 0.71 & \textbf{0.80} & 0.74 & 0.78 & 0.71 & \textbf{0.74} \\
\bottomrule
\end{tabular}
}%
\end{table}

\subsection{Alternative VLM: Qwen-2.5-VL-Instruct}
\label{sec:qwen25vl}

We further evaluate the robustness of \ourmethod{} to the choice of VLM by replacing Qwen3-VL-Instruct with the earlier Qwen-2.5-VL-Instruct~\cite{bai2023qwen} model. This tests whether our evolutionary framework relies on a specific VLM's capabilities or generalizes across different VLM backbones. The T2I model (FLUX-2-Klein) and all other settings remain unchanged.

Tab.~\ref{tab:qwen25vl} reports the results. \ourmethod{} still improves over the VLM-Baseline across all datasets and guidance metrics, demonstrating that the evolutionary optimization provides consistent gains regardless of the underlying VLM. Interestingly, comparing with Tab.~\ref{tab:main_results}, we observe that the Qwen-2.5-VL baseline already achieves competitive scores on several datasets (e.g., LAION-400M CLIP: 0.82), suggesting that this VLM produces strong initial descriptions. Nevertheless, \ourmethod{} consistently pushes these scores higher.

\begin{table}[H]
\centering
\caption{Quantitative comparison using Qwen-2.5-VL-Instruct as the VLM backbone (with FLUX-2-Klein as T2I model). We report CLIP, BLIP, DreamSim, and OpenCLIP image-to-image similarities.}
\label{tab:qwen25vl}
\adjustbox{max width=\textwidth}{%
\begin{tabular}{l cccc cccc cccc cccc cccc}
\toprule
& \multicolumn{4}{c}{\textbf{Lexica}} & \multicolumn{4}{c}{\textbf{COCO}} & \multicolumn{4}{c}{\textbf{CelebA}} & \multicolumn{4}{c}{\textbf{Flickr8K}} & \multicolumn{4}{c}{\textbf{LAION-400M}} \\
\cmidrule(lr){2-5} \cmidrule(lr){6-9} \cmidrule(lr){10-13} \cmidrule(lr){14-17} \cmidrule(lr){18-21}
 & CLIP & BLIP & DSim & OCL & CLIP & BLIP & DSim & OCL & CLIP & BLIP & DSim & OCL & CLIP & BLIP & DSim & OCL & CLIP & BLIP & DSim & OCL \\
\midrule
VLM-Baseline          & \textbf{0.80} & 0.85 & 0.76 & 0.79 & 0.81 & 0.85 & 0.74 & 0.79 & 0.67 & 0.79 & 0.74 & 0.72 & 0.80 & 0.85 & 0.74 & 0.79 & 0.82 & 0.84 & 0.76 & 0.78 \\
\midrule
\ourmethod{}-CLIP     & 0.79 & 0.84 & 0.75 & 0.77 & \textbf{0.82} & 0.84 & 0.75 & 0.80 & 0.67 & 0.76 & 0.72 & 0.70 & \textbf{0.82} & 0.84 & 0.74 & 0.80 & 0.82 & 0.82 & 0.75 & 0.77 \\
\ourmethod{}-BLIP     & 0.76 & \textbf{0.87} & 0.76 & 0.78 & 0.79 & \textbf{0.88} & 0.76 & 0.79 & 0.61 & 0.79 & 0.72 & 0.70 & 0.77 & \textbf{0.87} & 0.75 & 0.79 & 0.78 & \textbf{0.85} & 0.76 & 0.76 \\
\ourmethod{}-DreamSim & 0.76 & 0.85 & \textbf{0.77} & 0.77 & 0.78 & 0.86 & \textbf{0.77} & 0.79 & 0.63 & 0.77 & 0.74 & 0.70 & 0.77 & 0.85 & \textbf{0.77} & 0.79 & 0.78 & 0.83 & \textbf{0.78} & 0.76 \\
\ourmethod{}-OpenCLIP & 0.77 & 0.85 & 0.75 & \textbf{0.80} & 0.80 & 0.85 & 0.75 & \textbf{0.83} & 0.63 & 0.76 & 0.72 & 0.72 & 0.78 & 0.85 & 0.75 & \textbf{0.83} & 0.79 & 0.82 & 0.75 & \textbf{0.80} \\
\bottomrule
\end{tabular}
}%
\end{table}

\subsection{Cross-Model Prompt Transfer}
\label{sec:cross_model_transfer}

A key advantage of \ourmethod{} is that it produces natural-language prompts, which are inherently portable across different T2I models. To evaluate this transferability, we take the prompts evolved using FLUX-2-Klein (our default T2I model) and generate images using two alternative models: SD3.5-Turbo and SDXL-Turbo, \emph{without any re-optimization}. This tests whether prompts optimized for one model generalize to others.

Tab.~\ref{tab:cross_model} reports the results. We compare the VLM-Baseline prompts and each \ourmethod{} variant when transferred to each target model. The evolved prompts consistently maintain their advantage over the VLM-Baseline across both target models, confirming that the quality improvements from evolution are not specific to the T2I model used during optimization.

\begin{table}[H]
\centering
\caption{Cross-model prompt transfer: prompts evolved with FLUX-2-Klein are used to generate images with SD3.5-Turbo and SDXL-Turbo without re-optimization. We report CLIP, BLIP, DreamSim, and OpenCLIP image-to-image similarities.}
\label{tab:cross_model}
\adjustbox{max width=\textwidth}{%
\begin{tabular}{ll cccc cccc cccc cccc cccc}
\toprule
& & \multicolumn{4}{c}{\textbf{Lexica}} & \multicolumn{4}{c}{\textbf{COCO}} & \multicolumn{4}{c}{\textbf{CelebA}} & \multicolumn{4}{c}{\textbf{Flickr8K}} & \multicolumn{4}{c}{\textbf{LAION-400M}} \\
\cmidrule(lr){3-6} \cmidrule(lr){7-10} \cmidrule(lr){11-14} \cmidrule(lr){15-18} \cmidrule(lr){19-22}
\textbf{Target Model} & \textbf{Source} & CLIP & BLIP & DSim & OCL & CLIP & BLIP & DSim & OCL & CLIP & BLIP & DSim & OCL & CLIP & BLIP & DSim & OCL & CLIP & BLIP & DSim & OCL \\
\midrule
\multirow{5}{*}{SD3.5-Turbo}
 & VLM-Baseline          & 0.77 & 0.82 & 0.74 & 0.77 & 0.77 & 0.83 & 0.73 & 0.75 & 0.62 & 0.77 & 0.72 & 0.66 & 0.77 & 0.82 & 0.72 & 0.74 & 0.75 & 0.79 & 0.73 & 0.71 \\
 & \ourmethod{}-CLIP     & \textbf{0.79} & 0.83 & 0.74 & 0.78 & 0.77 & 0.82 & 0.72 & 0.75 & \textbf{0.63} & 0.77 & 0.72 & 0.67 & \textbf{0.78} & 0.81 & 0.72 & 0.74 & 0.75 & 0.79 & 0.73 & 0.72 \\
 & \ourmethod{}-BLIP     & 0.78 & \textbf{0.84} & 0.74 & 0.78 & 0.77 & 0.83 & 0.73 & 0.75 & 0.62 & \textbf{0.78} & 0.72 & 0.67 & 0.77 & \textbf{0.83} & 0.72 & 0.74 & 0.75 & \textbf{0.80} & 0.73 & 0.72 \\
 & \ourmethod{}-DreamSim & 0.79 & 0.84 & \textbf{0.75} & 0.79 & 0.77 & 0.83 & 0.73 & 0.75 & 0.62 & 0.77 & 0.72 & 0.66 & 0.77 & 0.82 & \textbf{0.73} & 0.74 & 0.75 & 0.79 & 0.73 & 0.72 \\
 & \ourmethod{}-OpenCLIP & 0.78 & 0.83 & 0.74 & \textbf{0.79} & 0.77 & 0.83 & 0.72 & \textbf{0.76} & 0.63 & 0.78 & 0.72 & \textbf{0.68} & 0.77 & 0.82 & 0.72 & \textbf{0.75} & 0.75 & 0.79 & 0.72 & \textbf{0.72} \\
\midrule
\multirow{5}{*}{SDXL-Turbo}
 & VLM-Baseline          & 0.77 & 0.81 & 0.74 & 0.78 & 0.75 & 0.78 & 0.71 & 0.74 & 0.63 & 0.76 & 0.72 & 0.68 & 0.76 & 0.78 & 0.70 & 0.73 & 0.69 & 0.74 & 0.69 & 0.66 \\
 & \ourmethod{}-CLIP     & \textbf{0.79} & 0.83 & 0.74 & 0.79 & \textbf{0.76} & 0.78 & 0.70 & 0.74 & \textbf{0.64} & 0.76 & 0.72 & 0.68 & 0.76 & 0.79 & 0.70 & 0.74 & \textbf{0.70} & 0.75 & 0.69 & \textbf{0.68} \\
 & \ourmethod{}-BLIP     & 0.78 & \textbf{0.83} & 0.74 & 0.79 & 0.75 & \textbf{0.79} & 0.71 & 0.74 & 0.64 & \textbf{0.77} & 0.72 & 0.67 & 0.76 & \textbf{0.80} & 0.70 & 0.75 & 0.69 & \textbf{0.75} & 0.69 & 0.67 \\
 & \ourmethod{}-DreamSim & 0.78 & 0.82 & \textbf{0.75} & 0.79 & 0.75 & 0.79 & 0.71 & 0.74 & 0.64 & 0.76 & 0.72 & 0.68 & 0.76 & 0.79 & \textbf{0.71} & 0.74 & 0.69 & 0.74 & 0.69 & 0.67 \\
 & \ourmethod{}-OpenCLIP & 0.78 & 0.82 & 0.74 & \textbf{0.79} & 0.76 & 0.79 & 0.71 & 0.74 & 0.64 & 0.77 & \textbf{0.73} & \textbf{0.69} & 0.76 & 0.79 & 0.70 & \textbf{0.75} & 0.69 & 0.74 & 0.69 & 0.67 \\
\bottomrule
\end{tabular}
}%
\end{table}

\subsection{Prompt Naturalness}
\label{sec:nll}

To assess the linguistic quality of the evolved prompts, we measure their naturalness using the negative log-likelihood (NLL) under a pretrained Mistral-7B-v0.3 language model. We chose this model, which is not from the Qwen family, to make sure to not have bias with the VLM language model. Lower NLL indicates more fluent, natural-sounding text. Tab.~\ref{tab:nll} reports the mean NLL ($\pm$ std) over all 500 evaluation images for each method. Both VLM-Baseline and \ourmethod{} produce dramatically more natural prompts (NLL ${\sim}2.8{-}2.9$) compared to all prior methods ($4.0{-}7.1$). Gradient-based approaches (PEZ: $6.88$, STEPS: $7.10$) yield the least natural text, as they optimize in a continuous embedding space and decode to tokens post-hoc. Gradient-free methods (CLIP Interrogator: $4.05$, VGD: $4.52$) fare better but still lag far behind. In contrast, \ourmethod{} variants show only a modest NLL increase over the VLM-Baseline ($+0.14$), confirming that the evolutionary process preserves prompt fluency while improving reconstruction fidelity.

\begin{table}[H]
\centering
\caption{Prompt naturalness measured by negative log-likelihood (NLL$\downarrow$) under Mistral-7B-v0.3. Lower values indicate more natural, fluent prompts. We report mean $\pm$ std over all 500 evaluation images.}
\label{tab:nll}
\begin{tabular}{lc}
\toprule
\textbf{Method} & \textbf{NLL}$\downarrow$ \\
\midrule
PEZ                   & $6.88 \pm 0.70$ \\
VGD                   & $4.52 \pm 0.51$ \\
CLIP Interrogator     & $4.05 \pm 0.36$ \\
STEPS                 & $7.10 \pm 0.73$ \\
\midrule
VLM-Baseline          & $\mathbf{2.80 \pm 0.32}$ \\
\midrule
\ourmethod{}-CLIP     & $2.93 \pm 0.30$ \\
\ourmethod{}-BLIP     & $2.94 \pm 0.31$ \\
\ourmethod{}-DreamSim & $2.94 \pm 0.31$ \\
\ourmethod{}-OpenCLIP & $2.94 \pm 0.31$ \\
\bottomrule
\end{tabular}
\end{table}

\end{document}